\documentclass[onecolumn]{article}

\usepackage{PRIMEarxiv}

\usepackage[T1]{fontenc}
\usepackage[table, svgnames, dvipsnames]{xcolor}

\usepackage{makeidx}  
\makeindex

\usepackage{url}
\usepackage{graphicx}
\usepackage{amsmath}
\usepackage{amssymb}
\usepackage{algorithm}      
\usepackage{algpseudocode}   
\usepackage[numbers]{natbib}
\usepackage{abstract}
\usepackage{booktabs}
\usepackage{array}
\usepackage{fancyhdr}
\usepackage{comment}
\usepackage{makecell}
\usepackage{placeins}
\usepackage{wrapfig}
\usepackage[many]{tcolorbox}  
\usepackage{parskip}
\usepackage{hyperref}

\hypersetup{
    colorlinks=true,
    linkcolor=Red,
    filecolor=magenta,
    urlcolor=Red,
}

\definecolor{light-gray}{gray}{0.95}

\DeclareMathOperator{\EX}{\mathbb{E}}

\newcommand*{\belowrulesepcolor}[1]{%
  \noalign{%
    \kern-\belowrulesep 
    \begingroup 
      \color{#1}%
      \hrule height\belowrulesep 
    \endgroup 
  }%
} 
\newcommand*{\aboverulesepcolor}[1]{%
  \noalign{%
    \begingroup 
      \color{#1}%
      \hrule height\aboverulesep 
    \endgroup 
    \kern-\aboverulesep 
  }%
} 

\newcommand{\appropto}{\mathrel{\vcenter{
  \offinterlineskip\halign{\hfil$##$\cr
    \propto\cr\noalign{\kern2pt}\sim\cr\noalign{\kern-2pt}}}}}

\newcommand{\textBF}[1]{%
    \pdfliteral direct {2 Tr 0.3 w} 
     #1%
    \pdfliteral direct {0 Tr 0 w}%
}

\newtcolorbox{boxC}{
    fontupper = \bf\color{black}, 
    boxrule = 1.5pt,
    colframe = black,
    colback   = white,
    rounded corners,
    arc = 5pt   
}

\newcolumntype{P}[1]{>{\centering\arraybackslash}p{#1}}

\begin{document}

\title{Variational autoencoders with latent high-dimensional steady geometric flows for dynamics}

\author{Andrew Gracyk \\  University of Illinois Urbana-Champaign
\\
605 E. Springfield Ave, Champaign, IL 61820, United States}

\date{}

\index{Gracyk, A.}

\maketitle

\begin{abstract}
We develop Riemannian approaches to variational autoencoders (VAEs) for PDE-type ambient data with regularizing geometric latent dynamics, which we refer to as VAE-DLM, or VAEs with dynamical latent manifolds. We redevelop the VAE framework such that manifold geometries, subject to our geometric flow, embedded in Euclidean space are learned in the intermediary latent space developed by encoders and decoders. By tailoring the geometric flow in which the latent space evolves, we induce latent geometric properties of our choosing, which are reflected in empirical performance. We reformulate the traditional evidence lower bound (ELBO) loss with a considerate choice of prior. We develop a linear geometric flow with a steady-state regularizing term. This flow requires only automatic differentiation of one time derivative, and can be solved in moderately high dimensions in a physics-informed approach, allowing more expressive latent representations. We discuss how this flow can be formulated as a gradient flow, and maintains entropy away from metric singularity. This, along with an eigenvalue penalization condition, helps ensure the manifold is sufficiently large in measure, nondegenerate, and a canonical geometry, which contribute to a robust representation. Our methods focus on the modified multi-layer perceptron architecture with tanh activations for the manifold encoder-decoder. We demonstrate, on our datasets of interest, our methods perform at least as well as the traditional VAE, and oftentimes better. Our methods can outperform this and a VAE endowed with our proposed architecture, frequently reducing out-of-distribution (OOD) error between 15\% to 35\% on select datasets. We highlight our method on ambient PDEs whose solutions maintain minimal variation in late times. We provide empirical justification towards how we can improve robust learning for external dynamics with VAEs.
\end{abstract}

\vspace{2mm}


\tableofcontents

\section{Introduction}

We operate through the lens of Riemannian geometry to study variational autoencoders (VAEs) \cite{kingma2022autoencodingvariationalbayes} \cite{Kingma_2019} \cite{dai2019hiddentalentsvariationalautoencoder} \cite{chadebec2022geometricperspectivevariationalautoencoders} by incorporating custom geometric flows as part of the latent dynamics. We study these variational approaches in the context of learning partial differential equation (PDE) dynamics, a topic which has garnered machine learning flavor from the research community \cite{ovadia2023realtimeinferenceextrapolationdiffusioninspired} \cite{li2021fourierneuraloperatorparametric} \cite{oommen2024integratingneuraloperatorsdiffusion} \cite{lopez2022gdvaesgeometricdynamicvariational} \cite{lopez2021variationalautoencoderslearningnonlinear} \cite{wang2021learningsolutionoperatorparametric} \cite{Lu_2021} \cite{Peyvan_2024} \cite{wang2024bridgingoperatorlearningconditioned}. The interconnections between Riemannian geometry with encoder-based machine learning frameworks act as sources to compose structure through data, and doing so accelerates desirable properties \cite{lopez2022gdvaesgeometricdynamicvariational} \cite{gong2012robustmultiplemanifoldsstructure} \cite{lassance2020representingdeepneuralnetworks} \cite{law2021ultrahyperbolicrepresentationlearning} \cite{jang2023geometrically} \cite{chadebec2022geometricperspectivevariationalautoencoders}
\cite{zhang2019manifoldadversariallearning}
\cite{jang2023geometrically}
\cite{moosavidezfooli2018robustnesscurvatureregularizationvice}, such as robust learning, identifiability, and extrapolation capacity.  We extend progress in VAEs and geometric VAEs by applying our choice of dynamical manifold latent space, in which a geometric flow is the accord to conduct such learning. In particular, we will enforce our own geometric flows in our latent spaces over latent spaces that develop in training without any additional constraints. Previous frameworks have been explored with latent manifolds, but primarily in the static setting \cite{connor2021variationalautoencoderlearnedlatent} \cite{Grattarola_2019} \cite{lopez2022gdvaesgeometricdynamicvariational} A latent geometric flow permits the entire ambient interval of the data to be learned and represented in a regularizing, geometric way. By regularization, we refer to improvements in performance in regard to generalization, typically for out-of-distribution data, which is done by constraining or penalizing the model in some way. We will study the steady dynamical manifold perspective to VAEs along with the empirical outcomes of the aforementioned qualities.

\vspace{2mm}

We perform computational approaches to address the extent to which geometric latent spaces add value to the relevant dynamics in the original space. We do so by enforcing manifold geometries that we organize beforehand through our flow construction. We use learning strategies to construct geometries and the corresponding flows, and so we develop empirical applications with comparisons to vanilla VAEs and extended VAEs with latent representations permitted to lie in unstructured Euclidean space with no enforced properties, constraints, or geometries aside from those that naturally develop under training. In particular, we attempt to quantify  sensitivity to noise, extrapolation with scaling, insertion of triviality in initial data, and learning for similar yet disparate conditions, as well as provide interpretability in relation to how data formulates structure for governing dynamics with comparisons to traditional VAEs. It is our goal to understand the role of geometry and structure in the latent representations.

\vspace{2mm}

Oftentimes, encoder-type frameworks, such as VAEs, have potential to naturally develop some notion of Riemannian structure, as discussed in \cite{chadebec2022geometricperspectivevariationalautoencoders}; however, the potential development of structure can be hindered. As discussed in \cite{connor2021variationalautoencoderlearnedlatent}, regularization with a prior can prevent latent structure. In \cite{davidson2022hypersphericalvariationalautoencoders}, they note in an autoencoder framework autoencoders can naturally discover latent circles, but not in the variational setting because of the regularization. We observe in our extended VAE framework, or VAE with both a parameterization and non-regularized latent stage, manifolds can develop on their own accord (see Appendix \ref{app:additional_figs}): we do not enforce any specific constraints or properties in this latent stage, yet this data can cohere to an underlying structure. Based on our observations, notably in the low-dimensional case, we remark this structure, if it exists, is somewhat irregular, and is inconsistent with more recognizable geometries. Also, as this latent space evolves in time, it follows more disorderly patterns. To counter these characteristics, our methods take this notion of geometric latent spaces further: we enforce our own manifold geometries in the latent space with properties we choose. In particular, we develop steady-state flows of manifolds that are nondegenerate, of sufficient measure, and canonical. Our methods follow a more smooth, regular flow. We find, by enforcing our own geometric properties, we can extract qualities out of the data that are empirically beneficial. As a consequence, we investigate the role of latent geometry and how the organization of different latent spaces has different empirical outcomes.

\vspace{2mm}

We focus our attention to intrinsic geometric flows in which the Riemannian metric corresponding to the latent manifold evolves according to own its PDE. This flow, also being a PDE, can be solved in a physics-informed approach in conjunction with the modified variational ELBO loss to solve the ambient dynamics. Physics-informed learning is another concentration of learning dynamics as of recent, and so we emphasize general techniques and strategies from this area \cite{wang2023expertsguidetrainingphysicsinformed} \cite{wang2021learningsolutionoperatorparametric} \cite{raissi2017physicsinformeddeeplearning} \cite{anandh2024fastvpinnstensordrivenaccelerationvpinns} \cite{fang2024ensemblelearningphysicsinformed}
\cite{raissi2024physicsinformedneuralnetworksextensions} to bolster our variational approaches to conduct the learning.

\vspace{2mm}

We consider choices of the prior. The first is by injecting noise into the manifold in its embedding. With this, structure is lost, as latent representations are now in regions about and not on the manifold. The question arises how much beneficial quality is introduced. The prior in the manifold also poses issues because it becomes difficult in how to formulate the manifold: creation of tangent vectors is now a hard task, and open regions of space are done via a mapping of a single point of parameterization. A potential remedy to the tangent issue is to construct the manifold with the means of the encoder creating the manifold, assuming the encoder returns means and variances as is done in the VAE framework. Furthermore, using KL divergence with a Gaussian prior to regularize the latent space is no longer a feasible task, and so maintaining suitable variance with typical means of regularization is a challenge. Lastly, with noise injected around the manifold, we cannot properly decode the output, as intersecting regions in space are meant to correspond to unique ambient times.

\vspace{2mm}

We propose parameterizing the manifold, and making the parameterization of the manifold the primary setting of the regularized prior. This resolves all previous issues. Now, the encoder is variational with means and variances in unrestricted Euclidean space with dimension being that of the intrinsic of the manifold, and we can impose regularization in this parameterization. By mapping this space to the manifold, noise now lies in the manifold space too but exactly on the manifold—not around the manifold. The geometric flow can now be formulated without issues, and the manifold can be properly decoded for the dynamics. The manifold, acting as a transformation of the data regularized with the prior, still captures this randomness, lying within the manifold to be decoded.

\vspace{2mm}

\section{Setup}

\begin{figure}
  \vspace{0mm}
  \centering
  \includegraphics[scale=0.8]{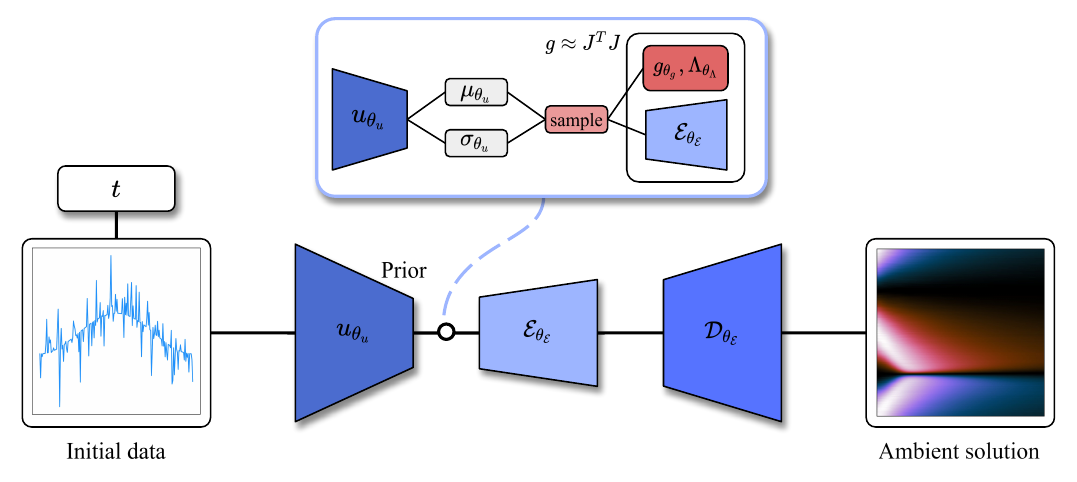}
  \caption{We illustrate our method. Initial data, possibly noised or out-of-distribution, is mapped through a series of encoders and decoders to predict the ambient PDE dynamics. Variational stage $u_{\theta_u}$ parameterizes the proceeding manifold stage and acts as a prior. Stage $\mathcal{E}_{\theta_{\mathcal{E}}}$ is a latent manifold regularizer that is subject to a geometric flow, which helps robust learning. A point on the manifold at a time $t$ can be decoded with stage $\mathcal{D}_{\theta_{\mathcal{D}}}$ to infer the dynamics also at time $t$. We refer to \cite{lopez2022gdvaesgeometricdynamicvariational} for a similar diagram for generalized VAEs for dynamics.}
  \label{fig:method}
\end{figure}

We seek a composition of neural networks
\begin{equation}
x_t \approx ( \mathcal{D} \circ \underbrace{\mathcal{E}}_{\text{manifold subject to geometric flow}} \circ \underbrace{u}_{\text{setting of the initial prior}})(x_0, t) = \mathcal{D}(\mathcal{E}(u(x_0),t))
\end{equation}
in a variational setting, where a prior $p$ is imposed in the parameterization space created by $u$. Because this region is such that $u \in \mathcal{U}$ parameterizes the manifold, governed by $\mathcal{E}$, we also have a probabilistic setting exactly along the manifold before it is decoded with $\mathcal{D}$. The geometric flow of interest is placed upon $\mathcal{E}$, which allows geometric latent representations of the entire continuum of the governing dynamics.

\vspace{2mm}

We restrict our attention to datasets, typically but not exclusively, of the form
\begin{align}
& f \in C(\mathbb{R}^D, [0,T] ; \mathbb{R}) \bigcap W^{n,1}(  \mathcal{X}, [0,T]; \mathbb{R}) \\
& = \Big\{ f : \int_{[0,T]} \int_{\mathcal{X}} | D^{\alpha_n} f(x,t) | dx dt < \infty,  f \in C(\mathbb{R}^D, [0,T]; \mathbb{R} ), | \alpha_n| \leq n \Big\} ,
\end{align}
for continuous functions with finite derivative integrals over $\mathcal{X} \times [0,T]$. We take $x_0 = f(x,0), x_t = f(x,t)$. In general, we work with continuous, bounded in $L^p$ norms, time-dependent data.

\subsection{Change of variables}

We establish our probabilistic framework by constructing a density in the parameterization domain. Then the encoder $\mathcal{E} : \mathcal{U} \times [0,T] \rightarrow \mathcal{M} \subseteq \mathbb{R}^d$ maps the probabilistic data in the parameterization to the manifold. Assume the inverse map $\mathcal{E}^{-1} : y_t \rightarrow u, y_t \in \mathcal{M}, u \in \mathcal{U}$ is well-defined. When considering a change of variable in regard to densities for square matrices, we have the result
\cite{zhang2021diffusionnormalizingflow}
\begin{equation}
p(y_t) = p(u) | \text{det} (\frac{\partial \mathcal{E}^{-1}}{\partial y_t} ) | = p(u) | \text{det} ( (\frac{\partial \mathcal{E}}{\partial u})^{-1} ) | = p(u) | \text{det} ( \frac{\partial \mathcal{E}}{\partial u} ) |^{-1},
\end{equation}
where $\mathcal{E}^{-1}$ is the inverse map $y_t \rightarrow u$, the second equality is by the inverse function theorem, and the third equality is by determinant properties. In our case, our Jacobian is non-square. Hence, the corresponding density over the manifold is done via the change of variables \cite{docarmo_riemannian}
\begin{align}
\label{eqn:change_of_variables}
p(y_t) & = p(u) \sqrt{ \text{det} \Big( ( \frac{ \partial \mathcal{E}^{-1}}{\partial y_t}  )^T ( \frac{ \partial \mathcal{E}^{-1}}{\partial y_t}  ) \Big)  } ,
\end{align}
where $u \in \mathcal{U} \subseteq \mathbb{R}^{d-1}, y_t \in \mathcal{M}$; however, the inverse function theorem does not apply because the Jacobian of the inverse is non-square, and we cannot simplify this further. The extrinsic dimension is one dimension higher than the intrinsic, i.e. $\mathcal{U} \subseteq \mathbb{R}^{d-1}$, and $y_t \in \mathbb{R}^d$. We will denote the above square root of the determinant as 
\begin{equation}
p(y_t) = p(u) \sqrt{\text{det}(J_{-1}^T J_{-1})}
\end{equation} 
for short. As we will see, our derived loss is independent of this change of variables, and whether or not the assumption of invertibility is valid is not of particular importance. Observe there is some relation to the calculation $J^T J$, which is actually the Riemannian metric. There is a known equivalence
\begin{equation}
g = J^T J ,
\end{equation}
which is explored in \cite{gemici2016normalizingflowsriemannianmanifolds}, \cite{doi:10.1137/S0895479895296896} \cite{elemofdiffgeo}.

\subsection{Modified evidence lower bound (ELBO) loss with a prior in the parameterization}

Denote $u_{\theta_u}$ the network mapping $x_0$ to the parameterization domain, $\mathcal{E}_{\theta_{\mathcal{E}}}$ the encoder function, and $\mathcal{D}_{\theta_{\mathcal{D}}}$ the decoder. Denote $\tilde{q}_{\theta_u}$ the encoding distribution in the parameterization domain created by $u_{\theta_u}$, $q_{\theta_{\mathcal{E}}}$ the encoding distribution onto the manifold. $\tilde{q}_{\theta_u}(u|x_0)$ is sampled with the reparameterization trick, $\tilde{q}_{\theta_u}(u|x_0) = \mu_q + \sigma_q \odot \epsilon$. $\mu_q$ and $\sigma_q$ are parameters computed by the parameterization neural network, and $\epsilon \sim N(0,I).$

\vspace{2mm}

To maximize the ELBO \cite{ormVI} \cite{Kingma_2019} \cite{fang_vae_elbo}, we are interested in maximizing the log-likelihood
\begin{equation}
\log p(x_t) .
\end{equation}
Observe in our time-dependent framework, this is different than input $x_0$. We denote the transformation $u \rightarrow z$ with $z_t = \mathcal{E}_{\theta_{\mathcal{E}}}(u,t)$. Now, observe
\begin{align}
\EX_{t \sim U[0,T]} [ \log p(x_t) ] =  \frac{1}{T} \int_{[0,T]} ( \log \int_{\mathcal{M}_t} p(x_t,z_t) \frac{q(z_t,t|x_0)}{q(z_t,t|x_0)} dV_{z_t} ) dt \geq  \EX_{t \sim U[0,T]} \EX_{z_t \sim q(\cdot,t | x_0)} [ \log( \frac{ p(x_t,z_t)}{q(z_t,t|x_0)} ) ]
\end{align}
where the inequality is by the analog of Jensen's inequality for concave functions. $q$ is the distribution along the manifold. Hence, using the change of variables, we have
\allowdisplaybreaks
\begin{align}
& \EX_{t \sim U[0,T]} \EX_{z_t \sim q(\cdot,t | x_0)} [  \log( \frac{ p(x_t,z_t)}{q(z_t,t|x_0)} ) ]  \stackrel{(1)}{=}  \EX_{t \sim U[0,T]} \EX_{u \sim \tilde{q}(\cdot|x_0)} [ \log(  \frac{p(x_t,z_t) }{ \tilde{q}(u|x_0) \sqrt{ \text{det} ( J_{-1}^T J_{-1}) }}   )  ]
\\
& = \EX_{t \sim U[0,T]}  \EX_{u \sim \tilde{q}(\cdot|x_0)} [   \log(  \frac{p(x_t|z_t) p(z_t) }{ \tilde{q}(u|x_0) \sqrt{ \text{det} ( J_{-1}^T J_{-1}) } }   )  ] 
\\ 
& =  \EX_{t \sim U[0,T]} \EX_{u \sim \tilde{q}(\cdot|x_0)} [   \log p(x_t|z_t) +  \log ( \frac{ p(z_t) }{ \tilde{q}(u|x_0) \sqrt{ \text{det} ( J_{-1}^T J_{-1}) }  } )  ]   
\\ & = \EX_{t \sim U[0,T]} \EX_{u \sim \tilde{q}(\cdot|x_0)} [  \log p(x_t|z_t)  +  \log ( \frac{ p(u) \sqrt{ \text{det} ( J_{-1}^T J_{-1}) } }{ \tilde{q}(u|x_0) \sqrt{ \text{det} ( J_{-1}^T J_{-1}) }  } )  ]   
\\
& =  \EX_{t \sim U[0,T]} \EX_{u \sim \tilde{q}(\cdot|x_0)} [   \log p(x_t|z_t) +  \log ( \frac{ p(u) }{ \tilde{q}(u|x_0)} )  ]
\end{align}
which gives us our modified ELBO loss function. The change in what the expectation is with respect to in the equality in (1) is in part due to equivalence of expectations under a change of variables. We remark this result does not rely on the use of the Jacobian. We enforce the prior distribution $p$ upon the parameterization and not the manifold. Hence, the loss to minimize is
\begin{align}
- \EX_{u \sim \tilde{q}(\cdot|x_0), t\sim U[0,T]} [ \log p(x_t|z_t)] +  D_{KL} ( \tilde{q}(u|x_0) || p(u) ) .
\end{align}
The expectation running over training data $x_0 \sim p(x_0)$ is implicit and omitted. As a further consideration, we consider a weighted version of the loss of the form
\begin{align}
-  \EX_{u \sim \tilde{q}(\cdot|x_0), t\sim U[0,T]} [ \log p(x_t|z_t)] +  \EX_{t \sim U[0,T]} [ D_{KL}^{\text{weighted}} ( \tilde{q}(u|x_0) || p(u) ) ]  & 
\\
D_{KL}^{\text{weighted}}  =  \EX_{u \sim \tilde{q}( \cdot | x_0) } [ \mathcal{J}(u,t)  \log( \frac{ \tilde{q}(u | x_0) }{ p(u) } ) ]  &
\end{align}
Here $\mathcal{J}$ is a weighting function. We found empirical success in taking $\mathcal{J} = \sqrt{ \text{det} (J^T J) }$. This leads to greater penalization of large geometric distortions in the manifold, which acts as a regularizer. The primary reason we consider this as is it yields potentially favorable empirical results.

\subsection{Generalized linear flows}

\begin{figure}
  \vspace{0mm}
  \centering
  \includegraphics[scale=0.7]{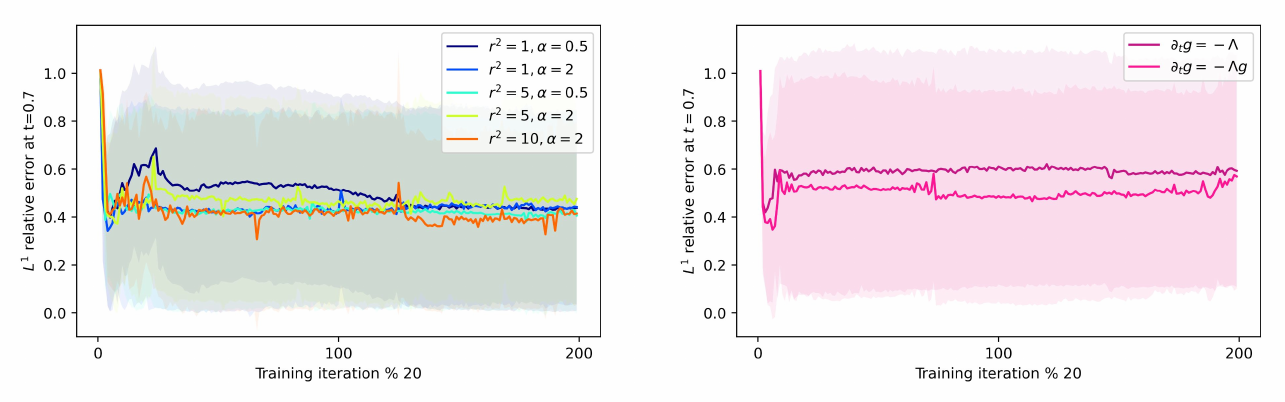}
  \caption{We examine $L^1$ relative error on an out-of-distribution (OOD) robustness test dataset of the Allen-Cahn equation with initial data $\phi = \alpha_1 \cos(\pi x) + \alpha_2 \cos^3(\pi x), \alpha_i \sim U([-1,1])$. All hyperparameters and architectures are held constant except choice of flow. The seeds are set for identical training and OOD data. For the OOD data, we scale in-distribution samples by $1.15$ and inject noise with $\sigma = 0.55$ in 150 locations over the 201-sized initial condition. Batch size is 400.  \textit{(left)} We examine linear geometric flows with a steady state term with various parameters, being $r^2$ and $\alpha$. Generally, there is dissimilarity among each flow upon retraining. For fair comparison, we observe the first handful of iterations, which tends to characterize overall training, and if loss is sufficiently desirable we continue training. We find the case $r^2 = 10$, $\alpha=2$ loosely performs best, but as we see in Figure \ref{fig:vanilla_and_best_linear}, it has variation among new instances of training. \textit{(right)} We examine the linear flows of $\partial_t g= -\Lambda$ and $\partial_t g = - \Lambda g$. As we see, this type of flow underperforms a steady state flow. We use eigenvalue regularization in all settings.}
  \label{fig:lin_geoflows}
\end{figure}

The primary geometric flow we consider is the linear partial differential equation on the Riemannian metric of the latent space of the form
\begin{equation}
\label{eqn:lin_flow}
\partial_t g(u,t) = -  A(u,t)^T A(u,t) g(u,t)   + \alpha(- g(u,t) + \Sigma(u) )  =  -  \Lambda(u,t) g(u,t)  + \alpha( -g(u,t) + \Sigma(u) ) .
\end{equation}
Our terms are organized as:
\begin{enumerate}
  \item $A^T A g = \Lambda g$ as a learned linear term
  \item $\alpha(g - \Sigma)$ as a steady state regularizing term.
\end{enumerate}
We denote $\Lambda$ as such as it inspired with the Einstein constant of differential geometry, but in our context, $\Lambda$ is a scaling matrix and not a constant. We denote $\Sigma$ the metric of the $d-1$-sphere,
\begin{align}
\Sigma(u,t) & = r^2(t) du^1 \otimes du^1 + r^2(t) \sum_{i=2}^{d-1} \prod_{j=1}^{i-1} \sin^2(u^j) du^i \otimes du^i , \ \ \ \ \ \text{or equivalently,}
\\
& = r^2(t) \begin{pmatrix}
& 1 & 0  &  \hdots & &  \\
& 0 & \sin^2(u^1) & 0 & \vdots
\\
& \vdots  & 0 & \ddots &
 & \\
& & \hdots & & \prod_{i=1}^{d-2} \sin^2 (u^i)
\end{pmatrix}
\in \mathbb{M}^{(d-1) \times (d-1)} ;
\end{align}
however, we remark this choice of the sphere can be granted flexibility for other geometries, although diagonal matrices work well. This steady state term is primarily a regularizer and helps robustness. $\alpha$ is a parameter to control the strength of the steady state across the flow, but the other term is allowed freedom and can be arbitrarily large despite the steady state tendency so that the data is learned appropriately. Indeed, it can be shown that the geometric flow without the steady state regularizer is a gradient flow that reaches a steady energy functional dissipation only at singularity. The steady state altogether produces a more robust representation that leads to lower out-of-distribution error, as we see in Figure \ref{fig:lin_geoflows}. We show our proposed geometric flow is also a gradient flow, which is important as we achieve stability far from singularity/triviality. The solution to the above geometric flow is of the form
\begin{equation}
g(u,t) = \Sigma(u) + e^{- \alpha t} ( g(u,0) - \Sigma(u) ) + \text{other terms} ,
\end{equation}
which is observable since, by taking the derivative,
\begin{align}
\partial_t g(u,t) &  =  \alpha e^{- \alpha t} ( \Sigma(u) - g(u,0) ) + \text{other terms}
\\
& = - \alpha g(u,t) + \alpha \Sigma(u) + \text{other terms} .
\end{align}
When the PDE is linear, the ability to split the solution as a sum corresponding to the sum in the PDE is known as superposition \cite{trinity_lecture_slides}.

\vspace{2mm}

We emphasize the highlighting feature of this geometric flow is that it can be computed with very low computational expensive. It requires only automatic differentiation of one time derivative. Other flows more established in literature, such as Ricci flow, require a more rigorous computational procedure, and automatic differentiation to a larger scale. This yields nontriviality in solving the flow in a physics-informed setting, and a high training cost. Our flow, by greatly reducing this cost, can be solved in moderately high-dimensions (up to intrinsic dimension 15 or even higher depending on the cost one is willing to incur). This allows more expressive latent representations, as we can represent an initial condition is a larger quantity of numbers over that in a more computationally sophisticated flow. We find an intrinsic dimension of around 6-10 generally performs well, and is sufficient to represent the initial data. This dimension is fairly low as far as our geometric flow is concerned, and does not add an extreme amount of training time by including a latent geometric flow.

\vspace{2mm}

We can assert that the Riemannian metric can approach a steady state from a manifold of larger or lower measure by penalizing eigenvalues. 
The flow approaching the steady state is dependent on external factors such as weight initialization to determine if the manifold expands or contracts in volume. This can be addressed by ensuring positive semi-definiteness of the difference matrix between the two, being $g - \Sigma$, as this ensures a metric at $t=0$ of appropriate measure. We can add to the objective function
\begin{equation}
\label{eqn:eigen_penalty}
\EX_{t \sim U[0,T], u \sim \tilde{q}(\cdot|x_0)} [ \sum_i \text{max}(0, \hat{\beta} \lambda_i) ]  = \EX_{t \sim U[0,T], u \sim \tilde{q}(\cdot|x_0)} [ \sum_i \text{relu}( \hat{\beta} \lambda_i( g - \Sigma ) ) ],
\end{equation}
where $\lambda_i$ is an eigenvalue of the matrix difference $g - \Sigma$. $\hat{\beta} = \pm 1$, and the sign can be chosen whether the metric expands or contracts. Without loss of generality, we will take $\hat{\beta}=-1$, which ensures the metric contracts, although we find this choice not particularly important as long as the manifold does not approach degeneracy.

\begin{figure}
  \vspace{0mm}
  \centering
  \includegraphics[scale=0.7]{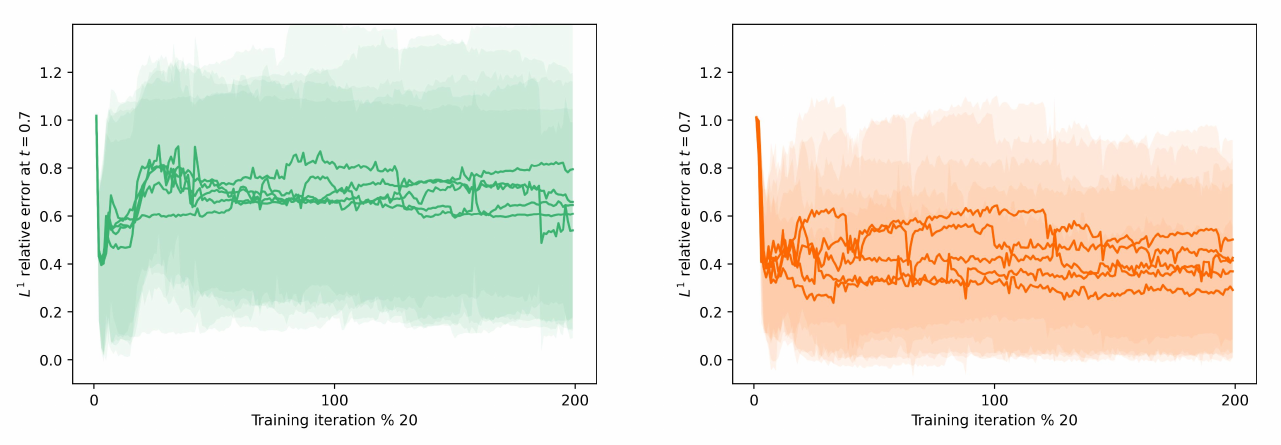}
  \caption{We examine $L^1$ relative error on 5 instances of training on the same OOD date of Figure \ref{fig:lin_geoflows} for \textit{(left)} a vanilla VAE, and \textit{(right)} the loosely best performing method of Figure \ref{fig:lin_geoflows}, which was $r^2 = 10, \alpha=2$. With this flow, error gets as low as $< 0.3$, which we did not observe with other steady state flows. The vanilla VAE error centralizes around $0.7$, while our method is about $0.4$.}
  \label{fig:vanilla_and_best_linear}
\end{figure}

\begin{figure}
  \vspace{0mm}
  \centering
  \includegraphics[scale=0.7]{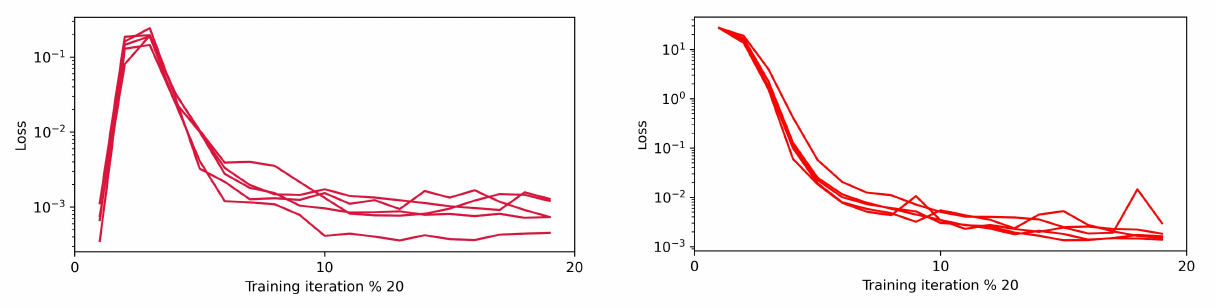}
  \caption{We examine training loss for the \textit{(left)} loss in learning the Riemannian metric, and \textit{(right)} physics-informed loss for the steady state geometric flow use a log scale on the y-axis. We see loss converges quite quickly, meaning a manifold is learned very rapidly. This conciliates the notion that the manifold takes effect immediately in Figures \ref{fig:lin_geoflows} and \ref{fig:metric_and_flow_loss}. We also remark in the two plots of Figure \ref{fig:vanilla_and_best_linear}, the error converges to around $0.4$ after several iterations in both cases, but the loss remains around here for our method (right), further conveying the manifold takes effect in early training.}
  \label{fig:metric_and_flow_loss}
\end{figure}

\begin{figure}
  \vspace{0mm}
  \centering
  \includegraphics[scale=0.7]{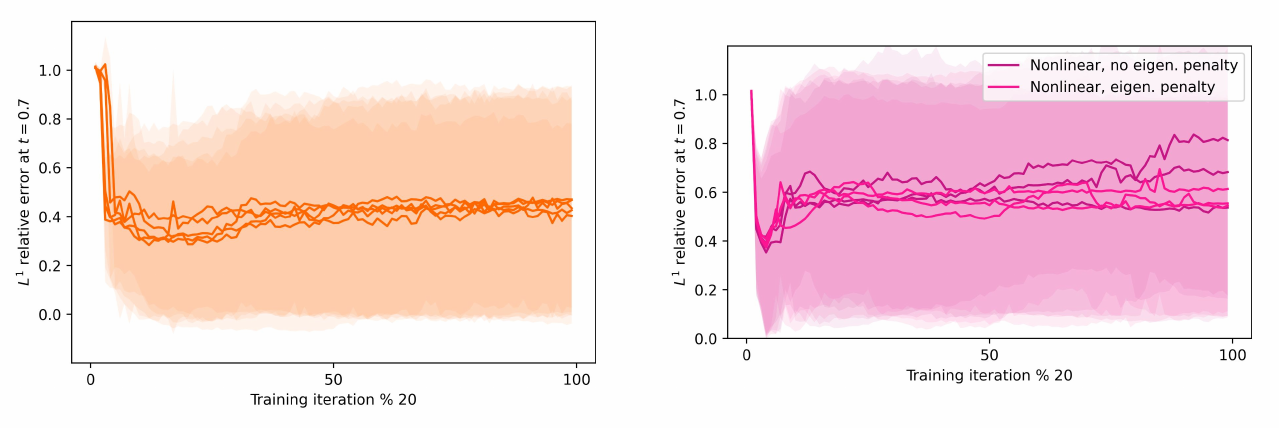}
  \caption{\textit{(left)} We can tune the training procedure on our custom linear geometric flow to make results more consistent. We added a slightly larger batch size of $1,000$; modifying the $\beta$ coefficient enforcing the KL divergence upon the prior, raising this from $1\mathrm{e}{-5}$ to $1\mathrm{e}{-2}$. Note that results are consistent but not necessarily quite as low in error as in Figure \ref{fig:vanilla_and_best_linear}; we remark these results are on the same data as in Figure \ref{fig:vanilla_and_best_linear}. \textit{(right)} We examine $L^1$ OOD results with nonlinear geometric flows, where $\Lambda = \Lambda(u,t,g(u,t))$ now depends on the learned Riemannian metric $g$. We plot 3 different instances of training for each category, where the the category is an inclusion of the eigenvalue penalty as discussed in equation \ref{eqn:eigen_penalty}. As we can see, the linear flow outperforms.}
  \label{fig:consistent_training}
\end{figure}

\vspace{2mm}

We establish some basic theory that our geometric flow of interest is a gradient flow. In our preliminary experiments without the steady state term, we found that a small latent space is learned in the sense that latent representations consisted of small numerical values. This in part likely develops since smaller values solve the objective function more easily. We also found the metric develops more randomness among its values, and overall less coherence to a more recognizable manifold. As we mentioned, the basic geometric flow without the steady state term gains stable entropy with respect to a functional only at singularity. This is undesirable because the latent space will tend to arbitrarily small values regardless. The steady state term, along with the optional eigenvalue condition, rectifies these issues. Also, the eigenvalue condition helps ensure sufficient measure and nondegeneracy, as well by a latent representation of a nonrandom geometry.

\vspace{2mm}

\textbf{Theorem 1.} The geometric flow of equation \ref{eqn:lin_flow} is a gradient flow with respect to the Euclidean metric and the functional
\begin{equation}
\label{eqn:functional}
\mathcal{F}(g) =  \int_{\mathcal{U}}  \frac{1}{2} \langle (\Lambda + \alpha I) g, g \rangle_F - \alpha \langle \Sigma, g \rangle_F du .
\end{equation}

\vspace{2mm}

\textbf{Lemma 1.} Suppose $\Lambda$ is positive semi-definite, $g$ a valid Riemannian metric, and the loss function of equation \ref{eqn:eigen_penalty} is exactly satisfied, i.e. $g - \Sigma \succeq 0$. Then the functional of equation \ref{eqn:functional} is nonnegative.

\vspace{2mm}

\textit{Proof.} Observe $\langle \Lambda g, g \rangle \geq 0$ since $\Lambda$ is positive semi-definite. Now, we have
$\langle \alpha I g, g \rangle - \alpha \langle \Sigma, g \rangle = \alpha ( || g ||_F^2 - \langle \Sigma, g \rangle )$. Now, observe $\Sigma$ is a diagonal matrix. Also, the elements of the diagonal of $g$ are greater than those of $\Sigma$, because if not, the diagonal of $g - \Sigma$ has negative elements, a contradiction that this is positive semi-definite. Hence, we have $\sum_{ij} g_{ij}^2 - \sum_{i} \Sigma_{ii} g_{ii} \geq \sum_i g_{ii}^2 - \sum_i \Sigma_{ii} g_{ii} \geq 0$, and we are done.

\vspace{2mm}

\textbf{Theorem 2.} The functional defined by \ref{eqn:functional} is nonincreasing on $t > 0$ if $\partial_t \Lambda$ is sufficiently small in the sense that
\begin{equation}
\label{eqn:func_deriv}
\frac{1}{2} \langle \Lambda_t g, g \rangle_F \leq || ( \Lambda + \alpha I)g - \alpha \Sigma ||_F^2 .
\end{equation}

\vspace{2mm}

\textbf{Remark.} We define a singularity as the case that $g$ is the zero matrix. This is notable as this definition is not used for all definitions of singularity. Our definition is consistent with the notion that as the manifold becomes arbitrarily small, its metric becomes arbitrarily small in all elements. Observe the functional is 0 at singularity, but the time derivative is nonzero. Also, the condition defined in equation \ref{eqn:func_deriv} holds at singularity. Additionally, the eigenvalue condition does not hold at singularity, so Lemma 1 does not hold.

\vspace{2mm}

\begin{figure}
  \vspace{0mm}
  \centering
  \includegraphics[scale=0.7]{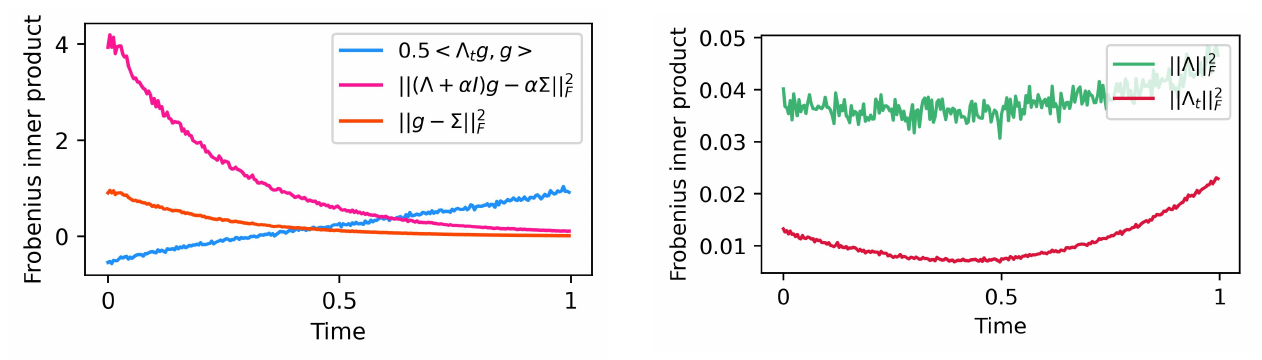}
  \caption{We illustrate the terms of equation \ref{eqn:func_deriv} plus additional norms on the Allen-Cahn equation experiment over a fixed normalized batch. As we can see, the entropy condition is not entirely satisfied in late training. Yet, the metric still stabilizes and approaches that of $\Sigma$, the sphere. From the right, we determine the contribution of $\Lambda$ to the flow is not particularly significant. As we can see, its derivative increases in late training, which contribute to the positive entropy derivative in late training. }
  \label{fig:functional_deriv}
\end{figure}

\textbf{Remark.} This condition is usually satisfied over much of the time domain in early training (see Figure \ref{fig:functional_deriv}); as we see, the magnitude of the term on the right-hand side of equation \ref{eqn:func_deriv} outscales the term on the left in early time, which helps push the functional towards somewhat stable entropy in early stages of the flow. What is important is that nondegeneracy of the manifold is achieved, which we have. We see in Figure \ref{fig:functional_deriv} the norm of $\Lambda$ is close to $0$, and the metric still tends to that of $\Sigma$, despite the entropy condition not satisfied in this late stage. Since $\Lambda$ remains small, this helps ensure relative stabilization of entropy. For example, take $g \approx \Sigma$ and $\Lambda$ small. Then, the functional value of equation \ref{eqn:functional} is quite small, and it is guaranteed nonnegative by Lemma 1.

\vspace{2mm}

The physics-informed geometric flow to be computed is
\begin{equation}
\label{eqn:physics_informed_loss}
\Big| \Big| \ \Big| \Big| (\partial_t +  \Lambda_{\theta_{\Lambda}}(u,t) ) g_{\theta_g}(u,t) + \alpha ( g_{\theta_g}(u,t) - \Sigma(u) ) \Big| \Big|_F^2 \ \Big| \Big|_{L^1(\mathcal{U} \times [0,T])} ,
\end{equation}
where the $L^1$ norm is with respect to a measure $\mu$ on $\mathcal{U}$ and Lebesgue measure on the time interval. Observe the physics-informed loss may be rewritten, and we have
\begin{align}
\label{eqn:integral_physics_loss}
\ref{eqn:physics_informed_loss} & = \frac{1}{T} \sum_{i} \int_{[0,T]} \int_{\mathcal{U}} ||  (\partial_t +  \Lambda_{\theta_{\Lambda}}(u,t) ) g_{\theta_g}(u,t) + \alpha ( g_{\theta_g}(u,t) - \Sigma(u) )  ||_F^2 \rho_i(u) du dt ,
\end{align}
where $i$ runs over training data. In practice, we take $\rho_i \propto \tilde{q}(\cdot|x_{0,i})$. This loss can be minimized as
\begin{align}
\label{eqn:expectation_physics_loss}
& \EX_{u \sim \tilde{q}(\cdot|x_0), t \sim U[0,T]} [ || (\partial_t +  \Lambda_{\theta_{\Lambda}}(u,t) ) g_{\theta_g}(u,t) + \alpha ( g_{\theta_g}(u,t) - \Sigma(u) )  ||_F^2 ]
\\
& \approx \frac{1}{N} \sum_{(u_i,t_i) \sim \tilde{q}(\cdot | x_{0,i}) \otimes U[0,T], x_{0,i} \sim p(x_0), i \in [N]}  || (\partial_t +  \Lambda_{\theta_{\Lambda}}(u_i,t_i) ) g_{\theta_g}(u_i,t_i) + \alpha ( g_{\theta_g}(u_i,t_i) - \Sigma(u_i) )  ||_F^2  .
\end{align}

Again, the expectation running over $x_0 \sim p(x_0)$ is omitted for brevity, as is typical in VAE literature. $\rho_i(u)$ is the Gaussian Radon-Nikodym derivative of a measure $\mu_i$ on $u$ such that the sum of $\rho_i$ is intended to match the Radon-Nikodym derivative of $\mu$ ($\mu_i, \mu$ are absolutely continuous with respect to Lebesgue measure $\lambda^*$). The scaling $1/T$ on the integral formulation is because time is sampled uniformly on $U([0,T])$. The summation runs over the batch. An alternative plausible geometric flow is by constructing the term $\Lambda$ to depend on $g$, making the flow nonlinear. We experimented with this, and altogether did not find it preferable (see Figure \ref{fig:consistent_training}). 

\vspace{2mm}

\textbf{Theorem 3.} A generalized nonlinear geometric flow of the form
\begin{equation}
\label{eqn:nonlinear_geo_flow}
\partial_t g(u,t) = - \Lambda(u,t,g(u,t))
\end{equation}
is a gradient flow if
\begin{equation}
\text{mat}( \text{Tr} (\frac{\partial \varphi}{\partial \Lambda}(\Lambda(u,t,g),g) 
\frac{\partial \Lambda^T}{\partial g_{ij} })) + \frac{\partial \varphi}{\partial g}(\Lambda(u,t,g),g) = \Lambda(u,t,g) 
\end{equation}
with respect to the functional
\begin{equation}
\mathcal{F}[g] = \int_{\mathcal{U}} \varphi(\Lambda(u,t,g(u,t)), g(u,t)) du
\end{equation}
for some $\varphi : \mathbb{M}^{d\times d} \times \mathbb{M}^{d \times d} \rightarrow \mathbb{R}^+$, and $\text{mat}$ denotes a matrix concatenation with respect to $ij$.

\vspace{2mm}

\textit{Proof.} See Appendix \ref{app:grad_flow_disc}.

\vspace{2mm}

\textbf{Remark.} Generally, it is nontrivial to deduce a solution to the above analytically (if it exists), and it is very hard to enforce the above condition computationally since $\Lambda, g$ would be learned with neural networks, and $\varphi$ would also need to be deduced; however, the linear flow we constructed is a known gradient flow. As a further note, the physics-informed learning task of our generalized nonlinear flow is the $L^1$ integral loss made discrete,
\begin{equation}
\Big| \Big| \ \Big| \Big| \partial_t g_{\theta_g}(u,t) +  \Lambda_{\theta_{\Lambda}} (u, t, g_{\theta_g} (u, t) )   \Big| \Big|_F^2 \ \Big| \Big|_{L^1(\mathcal{U} \times [0,\tau])} .
\end{equation}

\section{Training}

We utilize neural network $u_{\theta_u}$ to map to a learned parameterization domain $\mathcal{U} \subseteq \mathbb{R}^{d-1}$. This neural network outputs a mean $\mu_{\theta_u}$ and standard deviation $\sigma_{\theta_u}$ belonging to a local region in this domain. Then, we sample from such a local density using \cite{kingma2022autoencodingvariationalbayes}
\begin{equation}
u = \mu_{\theta_u} (x_0) + \sigma_{\theta_u} (x_0) \odot \epsilon
\end{equation}
for standard Gaussian noise $\epsilon$. Now, we map this onto the manifold with encoder $\mathcal{E}$. Our density is given by the change of variables of equation \ref{eqn:change_of_variables}. Hence, denote $z = \mathcal{E}(u,t)$ a point on the manifold, which also takes a time parameter corresponding to a state in the ambient data as well as the manifold's evolution. The ELBO type loss will be used in our loss using such a $z$, but it remains to enforce the manifold. We match the Riemannian metric with the manifold tangent inner products using $z$ with loss of the form
\begin{align}
 \EX_{u \sim \tilde{q}(\cdot|x_0)} [ \sum_{ij} | g_{\theta_g,ij}(u,t) - &
\langle \partial_i \mathcal{E}(u,t), \partial_j \mathcal{E}(u,t) \rangle |^2 ]  = \EX_{u \sim \tilde{q}(\cdot|x_0)} [ || g_{\theta_g}(u,t) - J^T J ||_F^2 ]  .
\end{align}
We define the primary objective in which we work. We define this as the \textcolor{orange}{\textbf{unweighted}} objective
\begin{align}
 &  \alpha \EX_{t\sim U[0,T], u \sim \tilde{q}(\cdot|x_0)} [ - \log p(x_t|z_t)] +  \beta D_{KL} ( \tilde{q}(u|x_0) || p(u) ) 
\\
 &  + \EX_{t \sim U[0,T], u \sim \tilde{q}(\cdot|x_0)} [ \gamma_{\text{geo flow}}  || (\partial_t +  \Lambda_{\theta_{\Lambda}}(u,t) ) g_{\theta_g}(u,t) + \alpha ( g_{\theta_g}(u,t) - \Sigma(u) )  ||_F^2 ]
 \\
 &  + \EX_{t \sim U[0,T], u \sim \tilde{q}(\cdot|x_0)} [ 
   \gamma_{\text{metric}} || g_{\theta_g}(u,t) -
J^T J  ||_F^2  ] .
\end{align}
We also consider an objective the weighted KL divergence (see Appendix \ref{App:modified_KL_div}). The unweighted is computationally simpler, more efficient, and does not require irregularity in scaling coefficients. Also, the parameterization will match the Gaussian prior more closely, and is better for generative-type scenarios. We find the two objectives yield similar results, with the unweighted slightly outperforming. The weighted objective can outperform (see Tables \ref{tab:burgers}, \ref{tab:PM}), which is why it is of consideration. We note the weighted objective is specific to our method and not generalized VAEs since our method is characterized by two latent stages, one of which encodes the corresponding geometric information to be penalized.

\vspace{2mm}

\textbf{Lemma 2.} Denote $\mathcal{J} = \sqrt{\text{det}(J^T J)}$. Denote $M>0$ such that $\mathcal{J} \geq M$, and assume that $\mathcal{J}$ has low variance in the sense that $(\EX[ \mathcal{J}^2 ])^{1/2} \approx \EX[ \mathcal{J} ]$. The KL divergence term of the weighted objective can be approximated as
\begin{align}
-  \EX_{t \sim U[0,T], u \sim \tilde{q}(\cdot|x_0)} & [ \mathcal{J} \log( \frac{ p(u) }{ \tilde{q(u|x_0)}} ) ] \approx  \frac{1}{2 N} \sum_{j=1}^N (\mathcal{J}(u_j,t_j)  ( - \sum_i   \log (  \sigma_{\tilde{q},j,i}^2 ) ) - dM 
\\
& + \mathcal{J}(u_j,t_j) (2 \sum_i \sigma_{\tilde{q},j,i}^4 + 4 \sum_i \mu_{\tilde{q},j,i}^2 \sigma_{\tilde{q},j,i}^2  + ( \sum_i \sigma_{\tilde{q},j,i}^2 + \mu_{\tilde{q},j,i}^2 )^2 )^{1/2} ) 
\end{align}
where $d$ is intrinsic dimension, index $j$ runs over the training batch, and $i$ runs over intrinsic dimension.

\vspace{2mm}

\textit{Proof.} See Appendix \ref{App:modified_KL_div}.

\vspace{2mm}

\textbf{Remark.} While this computation does indeed regularize the latent parameterization, it will not match a Gaussian prior exactly. For instance, if we take $\mu = 0, \sigma=1$, we find the above computation is positive and not zero. As a consequence, this weighted objective is not as effective for generative modeling.

\begin{algorithm*}[t]
\caption{General training with \textcolor{orange}{unweighted} objective}\label{alg:cap}
\textbf{Input:} batch size $N$, data samples $(x_0,t,x_t) $, networks $u_{\theta_u}, g_{\theta_g}, \Lambda_{\theta_{\Lambda}}, \mathcal{E}_{\theta_{\mathcal{E}}},\mathcal{D}_{\theta_{\mathcal{D}}}$ (subscripts omitted as follows for simplicity)
\begin{algorithmic}[1]
\While{$\mathcal{L}_{\text{total}}$ has not converged}
\State Evaluate $u_i = (\mu_{\tilde{q}} + \sigma_{\tilde{q}} \odot \epsilon)(x_{0,i}), \epsilon \sim N(0,I)$
\State Evaluate $\mathcal{E}(u_i,t_i)$
\State Evaluate $ \tilde{x}_t = (\mathcal{D}\circ \mathcal{E})(u_i,t_i)$
\State Evaluate $g(u_i,t_i), \Lambda(u_i, t_i)$
\State For upper triangular entries $ij$, compute $\partial_t g_{ij}$ with automatic differentiation, take $\partial_t g_{ji} = \partial_t g_{ij}$
\State Evaluate $J^T(u_i,t_i) J(u_i, t_i)  $ using $\mathcal{E}$
\State Evaluate objective \begin{align} \mathcal{L}_{\text{total}} = \frac{1}{N}  \sum_i & ( \alpha || \tilde{x}_{t,i} - x_{t,i} ||_2^2 + \beta ( - \sum_j ( \log \sigma_{\tilde{q},j,i}^2 ) - d + \sum_j \sigma_{\tilde{q},j,i}^2 + \sum_j \mu_{\tilde{q},j,i}^2 )
\\ & + \gamma_{\text{geo flow}} || (\partial_t + \Lambda(u_i,t_i)) g(u_i,t_i) + \alpha ( - g(u_i,t_i) + \Sigma(u_i) ) ||_F^2 
\\ & + \gamma_{\text{metric}} || g(u_i,t_i) - J^T (u_i,t_i) J(u_i, t_i) ||_F^2 )
\end{align}
\State (Optional) Add penalty $\frac{1}{N} \sum_i \sum_j \text{relu}(- \lambda_{ij} )$ to loss, where $\lambda_{ij}$ is the $j$-th eigenvalue of $g - \Sigma$ over sample $i$
\EndWhile
\end{algorithmic}
\end{algorithm*}

\subsection{Log-likelihood evaluation}

The log-likelihood is proportional to MSE,
\begin{equation}
- \log p(x_t | u, t) \propto  || x_t - \mu_{\theta_{\mathcal{D}}}(z_t) ||_2^2  = ||  x_t - \mathcal{D}_{\theta_{\mathcal{D}}} \circ \mathcal{E}_{\theta_{\mathcal{E}}} \circ (\mu_{\theta_u} + \sigma_{\theta_u} \odot \epsilon)(x_0,t) ||_2^2 ,
\end{equation}
where we operate under fixed variance here. The ambient dimension of the data, specifically the dimension of the domain $\Omega \subseteq \mathcal{X}$, is not necessarily in $\mathbb{R}$, but for our purposes, we will work with both (spatial) domain dimension and the image dimension of the ambient data in $\mathbb{R}$.

\section{Experiments}
\label{sec:experiments}

We deploy our method on a series of datasets for dynamics in varying complexity, notably:
\begin{enumerate}
  \item Burger's equation
  \item Allen-Cahn equation
  \item Modified porous medium equation
  \item Kuramoto-Sivashinsky equation
\end{enumerate}
We apply our methods for both deterministic mappings onto data in an autoencoder-type fashion but in the variational setting, as well as use the random sampling abilities in the prior to generatively model dynamics (see Appendix \ref{app:gen_modeling}). We emphasize empirically that our method is most suited for steady-state PDEs, or PDEs that maintain gradual, minimal variation as time progresses. This characterizes PDEs (1)-(3). This category of PDEs is paired well with our steady-state geometric flow. Our method can also be applied to non-steady PDEs, but not with guarantee of success in minimization of out-of-distribution error to any further scale that is naturally achieved with VAEs.

\vspace{2mm}

Datasets were constructed by converting the functions and its derivatives into the Fourier domain, calculating the derivatives of the form $u^{(j)}(x) = (i)^j k^j  \hat{u}$, where $k$ is the wave number discretization and $i$ is the imaginary number, and by using an \texttt{odeint} solver.

\vspace{2mm}

\subsection{Architectures}

We have found great empirical success with the modified multilayer perceptron (MLP) architecture of \cite{wang2021learningsolutionoperatorparametric} \cite{wang2020understandingmitigatinggradientpathologies}. We illustrate the architecture here: we embed data in high-dimensional spaces $U,V$, and the architecture proceeds iteratively as
\begin{align}
& U = \psi(X W^{U,1} + b^{U,1}), V = \psi( X W^{U,2} + b^{U,2} )
\\
& H^{(1)} = \psi ( X W^{H,1} + b^{H,1} ), Z^{(i)} = \psi (H^{(i)} W^{Z,(i)} + b^{Z,(i)} ), H^{(i+1)} = (1 - Z^{(i)})\odot U + Z^{(i)} \odot V
\\
& f_{\theta_f} = H^{(\text{final})} W + b .
\end{align}
$W,b$ denotes weights and biases. $\psi$ is a (possibly continuously differentiable) activation function. This architecture is primarily designed for physics-informed frameworks, most suitable for our Riemannian metric networks $g_{\theta_g}$, but we have generally found at least equivalent or superior results for the remaining networks, and so we choose this design for networks $u_{\theta_u}, \mathcal{E}_{\theta_{\mathcal{E}}}, \Lambda_{\theta_{\Lambda}}, \mathcal{D}_{\theta_{\mathcal{D}}}$ as well. We also found value in the choice of activation. For all experiments, we take gelu activation for $u_{\theta_u}, g_{\theta_g}, \Lambda_{\theta_{\Lambda}}$, which we found unimportant. We take tanh activation for $\mathcal{E}_{\theta_{\mathcal{E}}}, \mathcal{D}_{\theta_{\mathcal{D}}}$, which we did find notable: we find our methods are particularly strong with tanh activation in these two networks.

\vspace{2mm}

In all experiments, we make the restriction
\begin{equation}
g_{\theta_g} = A^T A, 
 \ \ \ \ \ \Lambda_{\theta_{\Lambda}} = B^T B,
\end{equation}
where $A,B \in \mathbb{M}^{(d-1)\times(d-1)}$ denote the outputs of the last layers of neural networks $g_{\theta_g}, \Lambda_{\theta_{\Lambda}}$. This ensures the Riemannian metrics and Einstein matrices are positive semi-definite.

\vspace{2mm}

We list training times and the relevant dimensions in Appendices \ref{app:training_time} and \ref{app:dimensions}. All networks were created with width $400$ and depth $3$ corresponding to the number of $H$ layers. We choose identical learning rates among the comparisons, typically beginning with $2\mathrm{e}{-4}$ and lowering it to $5\mathrm{e}{-5}$ in late training. We choose identical coefficients for $\alpha, \beta$ among all comparisons; however, our method has additional loss terms over baseline methods, and so we choose $\gamma_{\text{geo flow}}, \gamma_{\text{metric}}$ freely (usually $=1$). In many of our experiments, we choose a larger batch size $(>1,000)$, as we found this helps with more consistent training (see Figure \ref{fig:consistent_training}). For our error tables, we try to consider diversity of the OOD scenarios across the datasets in order to get a better understanding of how our methods behave on various examples of OOD data. We use the eigenvalue condition on all experiments except that of the porous medium equation, which we found had slightly better results here.

\vspace{2mm}

Networks are trained until approximate log-likelihood equivalence in training loss among the methods at early stages of convergence: the level of training has some significance in OOD error, and so we put in deliberate effort to train the networks as equally as we can. This is also equivalent to saying we train until in-distribution error are the same in value among the methods, up to the level of generalization in the networks. We also remark we typically do not significantly undertrain (early stopping) nor overtrain (grokking) the networks; the exception is the modified porous medium equation, which we slightly early stop. The level of regularization on the prior impacts the capabilities in how low the log-likelihood loss can get, and so we take this into consideration when training.

\subsection{Burger's equation}

\begin{figure}
  \vspace{0mm}
  \centering
  \includegraphics[scale=0.55]{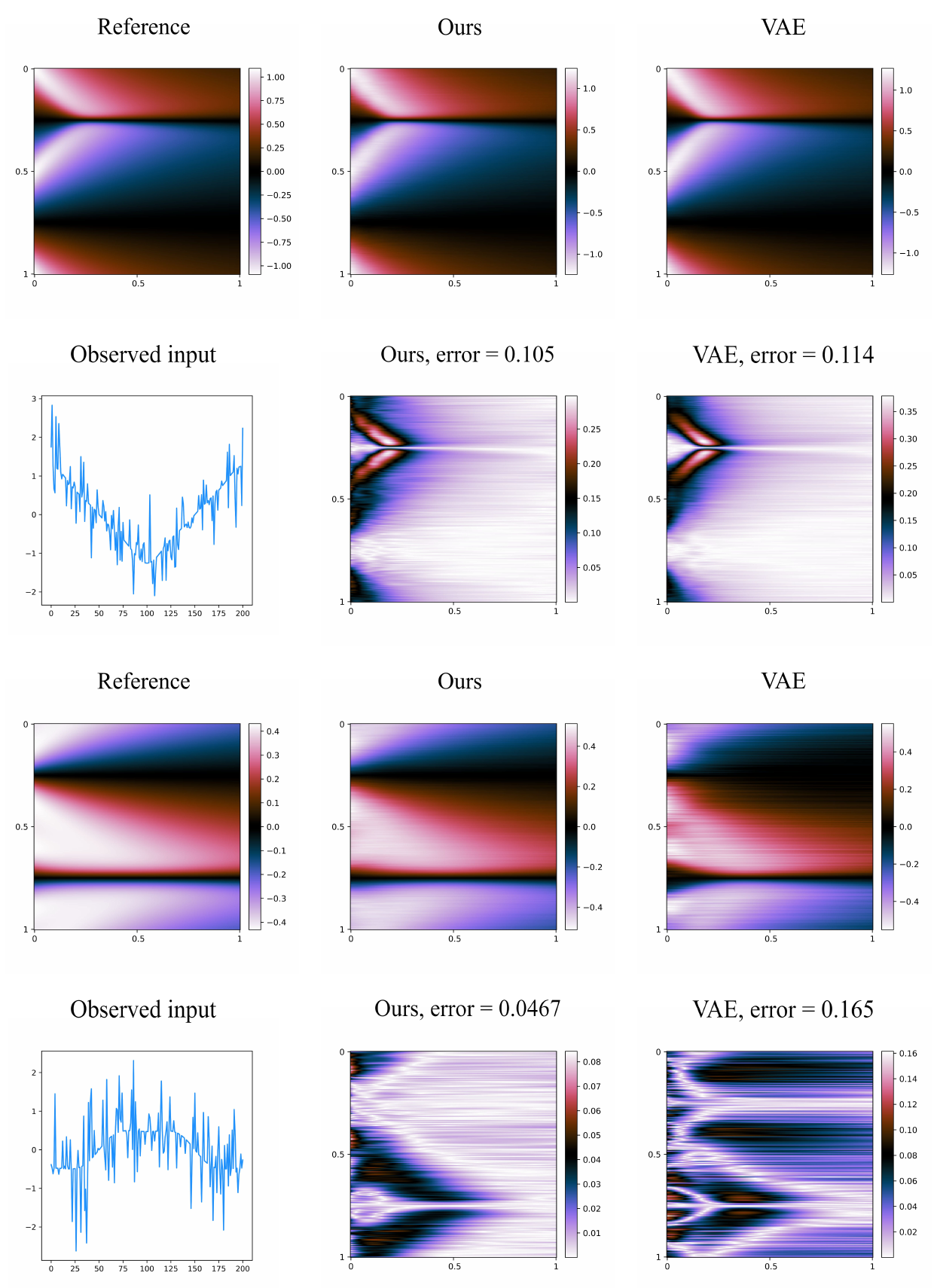}
  \caption{We examine OOD results on our method compared with a vanilla VAE on Burger's equation. The two settings are: \textit{(top)} an in-distribution initial condition injected with noise with $\sigma=0.65$ in 101 locations and scaled by $1.15$; \textit{(bottom)} an in-distribution inital condition injected with noise with $\sigma=0.95$ in 101 locations scaled by $1.15$. We assess our error using relative $L^1$ double integral error with respect to both space and time over $(x,t) \in [0,1]\times[0,1]$ (note the error graphical portion of the figure displays absolute value error).}
  \label{fig:AllenCahn_OOD_eq}
\end{figure}

Our primary experimental benchmark is done with the viscous Burger's equation
\begin{equation}
\begin{cases}
\partial_t \phi(x,t) = \nu \partial_{xx} \phi(x,t) - \phi(x,t) \partial_x \phi(x,t), (x,t) \in [0,1] \times [0,1] ,
\\
\phi(x,0) = \alpha_1 \cos( 2 \pi x) + \alpha_2 \cos^3(2 \pi x) + \alpha_3 \cos^5(2 \pi x) , \alpha_i \in U([-1,1]),
\end{cases}
\end{equation} 
where we take $\nu = 0.01$, and $U$ denotes the uniform distribution. Solutions are discretized  into a $201 \times 201$ mesh. We consider 5,000 training samples with initial conditions sampled from above.

\subsection{Allen-Cahn equation}

We consider the Allen-Cahn equation with initial data
\begin{equation}
\begin{cases}
\partial_t \phi(x,t) = 0.0001 \partial_{xx} \phi (x,t) - 5 \phi^3 (x,t) + 5 \phi (x,t), (x,t) \in [-1,1] , \times [0,1]
\\
\phi(x,0) = \alpha_1 x^2 \cos(  \pi x) + \alpha_2 x^2 \cos^3(\pi x)  , \alpha_i \sim U([-1,1])
\end{cases}
\end{equation} 
discretized into a $201 \times 201$ mesh. We generate 4,000 samples for training.

\subsection{Modified porous medium equation}

We consider a modified version of the porous medium (PM) equation with a diffusion term
\begin{equation}
\begin{cases}
\partial_t \phi(x,t) = D \partial_x (\phi^2(x,t) \partial_x \phi(x,t) ) - \gamma \partial_{xx} \phi(x,t) , (x,t) \in [-0.8,0.8] \times [0,1]
\\
\phi(x,0) = \alpha_1 |x| \cos( \beta_1 x) + \alpha_2 |x| \sin( \beta_2 x), \alpha_i \sim U([-1,1]), \beta_i \sim U([\pi/2, 2 \pi])
\end{cases}
\end{equation} 
We take $D=0.5, \gamma=4.9\mathrm{e}{-4}$. We evaluate over a $201 \times 201$ mesh. We generate 1,000 training samples.

\vspace{2mm}

In this experiment, we found high KL regularization $(\beta = 0.01)$ and a high intrinsic dimension of $8$ had performance gains with our method. In particular, our method underperformed with low intrinsic dimension of $2$, but this is highly restrictive of the latent representation anyway.

\begin{table}[htbp]
\centering

\scriptsize

\begin{tabular}{wl{3.0cm} | P{2.2cm} P{2.2cm} P{2.2cm} P{2.2cm} P{2.2cm} } 
\toprule
\belowrulesepcolor{light-gray} 
\rowcolor{light-gray} \multicolumn{6}{l} {Burger's equation} 
\\ 
\aboverulesepcolor{light-gray} 
\midrule
{VAE}  & {$t=0$} & {$t=0.25$} & {$t=0.5$} & {$t=0.75$} & {$t=1$} 
\\ 
\midrule
{In-distribution}  & {$3.90\mathrm{e}{-2} \pm 2.26\mathrm{e}{-2}$} & {$1.56\mathrm{e}{-2} \pm 1.50\mathrm{e}{-2}$} & {$1.59\mathrm{e}{-2} \pm 1.74\mathrm{e}{-2}$} & {$1.80\mathrm{e}{-2} \pm 2.00\mathrm{e}{-2}$} & {$3.10\mathrm{e}{-2} \pm 3.49\mathrm{e}{-2}$}
\\
{Out-of-distribution 1}  & {$2.61\mathrm{e}{-1} \pm 1.73\mathrm{e}{-1}$} & {$2.81\mathrm{e}{-1} \pm 3.31\mathrm{e}{-1}$} & {$3.32\mathrm{e}{-1} \pm 4.91\mathrm{e}{-1}$} & {$3.42\mathrm{e}{-1} \pm 5.61\mathrm{e}{-1}$} & {$3.35\mathrm{e}{-1} \pm 5.78\mathrm{e}{-1}$}
\\
{Out-of-distribution 2}  & {$2.76\mathrm{e}{-1} \pm 2.22\mathrm{e}{-1}$} & {$2.99\mathrm{e}{-1} \pm 3.73\mathrm{e}{-1}$} & {$3.22\mathrm{e}{-1} \pm 4.86\mathrm{e}{-1}$} & {$3.05\mathrm{e}{-1} \pm 4.50\mathrm{e}{-1}$} & {$2.85\mathrm{e}{-1} \pm 4.00\mathrm{e}{-1}$}
\\
{Out-of-distribution 3}  & {$1.19\mathrm{e}{0} \pm 9.14\mathrm{e}{-1}$} & {$1.20\mathrm{e}{0} \pm 1.49\mathrm{e}{0}$} & {$1.12\mathrm{e}{0} \pm 1.46\mathrm{e}{0}$} & {$1.00\mathrm{e}{0} \pm 1.22\mathrm{e}{0}$} & {$9.07\mathrm{e}{-1} \pm 1.04\mathrm{e}{0}$}
\\
{Out-of-distribution 4}  & {$3.05\mathrm{e}{-1} \pm 3.14\mathrm{e}{-1}$} & {$3.38\mathrm{e}{-1} \pm 6.65\mathrm{e}{-1}$} & {$4.25\mathrm{e}{-1} \pm 1.25\mathrm{e}{-1}$} & {$4.64\mathrm{e}{-1} \pm 1.57\mathrm{e}{-1}$} & {$4.62\mathrm{e}{-1} \pm 1.66\mathrm{e}{0}$}
\\

\midrule
{VAE, extended}  & {$t=0$} & {$t=0.25$} & {$t=0.5$} & {$t=0.75$} & {$t=1$} 
\\ 
\midrule
{In-distribution}  & {$8.05\mathrm{e}{-2} \pm 5.41\mathrm{e}{-2}$} & {$2.83\mathrm{e}{-2} \pm 2.92\mathrm{e}{-2}$} & {$2.31\mathrm{e}{-2} \pm 3.19\mathrm{e}{-2}$} & {$2.55\mathrm{e}{-2} \pm 4.64\mathrm{e}{-2}$} & {$3.65\mathrm{e}{-2} \pm 5.83\mathrm{e}{-2}$}
\\
{Out-of-distribution 1}  & {$2.28\mathrm{e}{-1} \pm 1.87\mathrm{e}{-1}$} & {$2.37\mathrm{e}{-1} \pm 2.76\mathrm{e}{-1}$} & {$2.41\mathrm{e}{-1} \pm 3.14\mathrm{e}{-1}$} & {$2.24\mathrm{e}{-1} \pm 3.02\mathrm{e}{-1}$} & {$2.08\mathrm{e}{-1} \pm 2.88\mathrm{e}{-1}$}
\\
{Out-of-distribution 2}  & {$2.03\mathrm{e}{-1} \pm 1.65\mathrm{e}{-1}$} & {$2.13\mathrm{e}{-1} \pm 2.39\mathrm{e}{-1}$} & {$2.18\mathrm{e}{-1} \pm 2.92\mathrm{e}{-1}$} & {$2.03\mathrm{e}{-1} \pm 2.77\mathrm{e}{-1}$} & {$1.89\mathrm{e}{-1} \pm 2.52\mathrm{e}{-1}$}
\\
{Out-of-distribution 3}  & {$7.92\mathrm{e}{-1} \pm 7.32\mathrm{e}{-1}$} & {$8.84\mathrm{e}{-1} \pm 1.06\mathrm{e}{0}$} & {$8.43\mathrm{e}{-1} \pm 1.14\mathrm{e}{0}$} & {$7.40\mathrm{e}{-1} \pm 1.00\mathrm{e}{0}$} & {$6.53\mathrm{e}{-1} \pm 8.59\mathrm{e}{-1}$}
\\
{Out-of-distribution 4}  & {$2.89\mathrm{e}{-1} \pm 3.15\mathrm{e}{-1}$} & {$2.99\mathrm{e}{-1} \pm 6.74\mathrm{e}{-1}$} & {$3.62\mathrm{e}{-1} \pm 1.23\mathrm{e}{0}$} & {$3.94\mathrm{e}{-1} \pm 1.52\mathrm{e}{0}$} & {$4.03\mathrm{e}{-1} \pm 1.61\mathrm{e}{0}$}
\\

\midrule
{VAE-DLM unweighted (ours)}  & {$t=0$} & {$t=0.25$} & {$t=0.5$} & {$t=0.75$} & {$t=1$} 
\\ 
\midrule
{In-distribution}  & {$8.62\mathrm{e}{-2} \pm 5.71\mathrm{e}{-2}$} & {$3.00\mathrm{e}{-2} \pm 2.97\mathrm{e}{-2}$} & {$3.04\mathrm{e}{-2} \pm 5.11\mathrm{e}{-2}$} & {$2.94\mathrm{e}{-2} \pm 5.40\mathrm{e}{-2}$} & {$4.63\mathrm{e}{-2} \pm 8.23\mathrm{e}{-2}$}
\\
{Out-of-distribution 1}  & {$\textBF{2.19\mathrm{e}{-1} \pm 1.95\mathrm{e}{-1}}$} & \textcolor{blue}{$\textBF{1.77\mathrm{e}{-1} \pm 2.08\mathrm{e}{-1}}$} & \textcolor{blue}{$\textBF{1.71\mathrm{e}{-1} \pm 2.39\mathrm{e}{-1}}$} & \textcolor{blue}{$\textBF{1.65\mathrm{e}{-1} \pm 2.48\mathrm{e}{-1}}$} & {$\textBF{1.68\mathrm{e}{-1} \pm 2.59\mathrm{e}{-1}}$} 
\\
{Out-of-distribution 2}  & {$\textBF{1.98\mathrm{e}{-1} \pm 1.49\mathrm{e}{-1}}$} & {$\textBF{1.76\mathrm{e}{-1} \pm 1.99\mathrm{e}{-1}}$} & \textcolor{blue}{$\textBF{1.63\mathrm{e}{-1} \pm 2.42\mathrm{e}{-1}}$} & \textcolor{blue}{$\textBF{1.48\mathrm{e}{-1} \pm 2.32\mathrm{e}{-1}}$}
 & {$\textBF{1.42\mathrm{e}{-1} \pm 2.24\mathrm{e}{-1}}$}
\\
{Out-of-distribution 3}  & {$7.06\mathrm{e}{-1} \pm 5.54\mathrm{e}{-1}$} & \textcolor{blue}{$\textBF{6.51\mathrm{e}{-1} \pm 7.37\mathrm{e}{-1}}$} & \textcolor{blue}{$\textBF{5.38\mathrm{e}{-1} \pm 6.95\mathrm{e}{-1}}$} & \textcolor{blue}{$\textBF{4.44\mathrm{e}{-1} \pm 5.47\mathrm{e}{-1}}$} & \textcolor{blue}{$\textBF{4.23\mathrm{e}{-1} \pm 4.03\mathrm{e}{-1}}$}
\\
{Out-of-distribution 4}  & {$2.54\mathrm{e}{-1} \pm 2.59\mathrm{e}{-1}$} & {$2.40\mathrm{e}{-1} \pm 5.36\mathrm{e}{-1}$} & {$2.68\mathrm{e}{-1} \pm 9.36\mathrm{e}{-1}$} & {$2.91\mathrm{e}{-1} \pm 1.14\mathrm{e}{0}$} & {$3.10\mathrm{e}{-1} \pm 1.25\mathrm{e}{0}$}
\\

\midrule
{VAE-DLM weighted (ours)}  & {$t=0$} & {$t=0.25$} & {$t=0.5$} & {$t=0.75$} & {$t=1$} 
\\ 
\midrule
{In-distribution}  & {$8.74\mathrm{e}{-2} \pm 5.77\mathrm{e}{-2}$} & {$2.92\mathrm{e}{-2} \pm 2.54\mathrm{e}{-2}$} & {$3.13\mathrm{e}{-2} \pm 4.80\mathrm{e}{-2}$} & {$2.82\mathrm{e}{-2} \pm 3.95\mathrm{e}{-2}$} & {$5.16\mathrm{e}{-2} \pm 1.08\mathrm{e}{-1}$}
\\
{Out-of-distribution 1}  & {$2.25\mathrm{e}{-1} \pm 1.53\mathrm{e}{-1}$} & {$1.91\mathrm{e}{-1} \pm 2.31\mathrm{e}{-1}$} & {$1.83\mathrm{e}{-1} \pm 2.58\mathrm{e}{-1}$} & {$1.84\mathrm{e}{-1} \pm 2.70\mathrm{e}{-1}$} & {$1.97\mathrm{e}{-1} \pm 2.88\mathrm{e}{-1}$}
\\
{Out-of-distribution 2}  & {$2.01\mathrm{e}{-1} \pm 1.35\mathrm{e}{-1}$} & {$1.81\mathrm{e}{-1} \pm 1.92\mathrm{e}{-1}$} & {$1.78\mathrm{e}{-1} \pm 2.40\mathrm{e}{-1}$} & {$1.71\mathrm{e}{-1} \pm 2.31\mathrm{e}{-1}$} & {$1.70\mathrm{e}{-1} \pm 2.20\mathrm{e}{-1}$}
\\
{Out-of-distribution 3}  & {$6.42\mathrm{e}{-1} \pm 5.60\mathrm{e}{-1}$} & {$6.62\mathrm{e}{-1} \pm 7.59\mathrm{e}{-1}$} & {$5.66\mathrm{e}{-1} \pm 6.67\mathrm{e}{-1}$} & {$5.07\mathrm{e}{-1} \pm 5.81\mathrm{e}{-1}$} & {$5.00\mathrm{e}{-1} \pm 5.59\mathrm{e}{-1}$}
\\
{Out-of-distribution 4}  & {$\textBF{2.52\mathrm{e}{-1} \pm 2.38\mathrm{e}{-1}}$} & {$\textBF{2.31\mathrm{e}{-1} \pm 4.69\mathrm{e}{-1}}$} & {\textcolor{blue}{$\textBF{2.58\mathrm{e}{-1} \pm 8.29\mathrm{e}{-1}}$}} & \textcolor{blue}{$\textBF{2.72\mathrm{e}{-1} \pm 9.98\mathrm{e}{-1}}$} & 
\textcolor{blue}{$\textBF{2.86\mathrm{e}{-1} \pm 1.07\mathrm{e}{0}}$}
\\

\bottomrule

\end{tabular}

\caption{We list relative $L^1$ errors of our method along with a VAE baseline on the Burger's equation on 30 test samples. The seed is set to ensure identicality among both noise and initial data. For the VAE methods, we include both a vanilla VAE and a VAE with extended architecture, which is an architecture identical to ours but without the latent geometric flow (so $u_{\theta_u}, \mathcal{E}_{\theta_{\mathcal{E}}}, \mathcal{D}_{\theta_{\mathcal{D}}}$ are all included). For our methods, we include both the unweighted and weighted formulation. We train all scenarios until approximate convergence (no early stopping or overtraining). All hyperparameters in both our method and the traditional VAE are the same when applicable. We choose scaling coefficients $(\alpha, \beta, \gamma_{\text{geo flow}}, \gamma_{\text{metric}}) = (100,1\mathrm{e}{-4},1,1)$. Our OOD scenarios are: (1) no scaling of an in-distribution initial condition with noise with $\sigma=0.75$ injected in 101 locations; (2) noise with $\sigma=1.0$ injected in 51 locations; (3) noise with $\sigma=4.0$ injected in 51 locations; (4) scaling of 1.15 and noise with $\sigma=0.45$ injected in all 201 locations. \textbf{Bolded} numbers indicate best OOD results with a margin under 25\%. $\textcolor{blue}{\textBF{\text{Blue}}}$ denotes the error improvement is by a margin of over 25\% over the next best alternative that is not ours.}

\label{tab:burgers}

\end{table}

\begin{table}[htbp]
\centering

\scriptsize

\begin{tabular}{wl{3.0cm} | P{2.2cm} P{2.2cm} P{2.2cm} P{2.2cm} P{2.2cm} } 
\toprule
\belowrulesepcolor{light-gray} 
\rowcolor{light-gray} \multicolumn{6}{l} {Allen-Cahn equation} 
\\ 
\aboverulesepcolor{light-gray} 
\midrule
{VAE}  & {$t=0$} & {$t=0.25$} & {$t=0.5$} & {$t=0.75$} & {$t=1$} 
\\ 
\midrule
{In-distribution}  & {$1.08\mathrm{e}{-1} \pm 2.32\mathrm{e}{-1}$} & {$1.91\mathrm{e}{-2} \pm 3.59\mathrm{e}{-2}$} & {$9.84\mathrm{e}{-3} \pm 1.44\mathrm{e}{-2}$} & {$7.19\mathrm{e}{-3} \pm 7.03\mathrm{e}{-3}$} & {$1.40\mathrm{e}{-2} \pm 1.36\mathrm{e}{-2}$}
\\
{Out-of-distribution 1}  & {$1.44\mathrm{e}{0} \pm 1.60\mathrm{e}{0}$} & {$9.76\mathrm{e}{-1} \pm 9.73\mathrm{e}{-1}$} & {$8.13\mathrm{e}{-1} \pm 7.98\mathrm{e}{-1}$} & {$7.48\mathrm{e}{-1} \pm 7.13\mathrm{e}{-1}$} & {$7.13\mathrm{e}{-1} \pm 6.59\mathrm{e}{-1}$}
\\
{Out-of-distribution 2}  & {$1.94\mathrm{e}{0} \pm 1.77\mathrm{e}{0}$} & {$1.36\mathrm{e}{0} \pm 1.26\mathrm{e}{0}$} & {$1.13\mathrm{e}{0} \pm 9.23\mathrm{e}{-1}$} & {$9.96\mathrm{e}{-1} \pm 6.67\mathrm{e}{-1}$} & {$9.40\mathrm{e}{-1} \pm 5.48\mathrm{e}{-1}$}
\\
{Out-of-distribution 3}  & {$1.95\mathrm{e}{0} \pm 2.33\mathrm{e}{0}$} & {$1.17\mathrm{e}{0} \pm 1.20\mathrm{e}{0}$} & {$8.51\mathrm{e}{-1} \pm 6.89\mathrm{e}{-1}$} & {$7.42\mathrm{e}{-1} \pm 5.25\mathrm{e}{-1}$} & {$6.91\mathrm{e}{-1} \pm 5.36\mathrm{e}{-1}$}
\\
{Out-of-distribution 4}  & {$1.15\mathrm{e}{0} \pm 1.37\mathrm{e}{0}$} & {$6.62\mathrm{e}{-1} \pm 8.40\mathrm{e}{-1}$} & {$4.72\mathrm{e}{-1} \pm 5.99\mathrm{e}{-1}$} & {$4.07\mathrm{e}{-1} \pm 5.06\mathrm{e}{-1}$} & {$3.60\mathrm{e}{-1} \pm 4.77\mathrm{e}{-1}$}
\\

\midrule
{VAE, extended}  & {$t=0$} & {$t=0.25$} & {$t=0.5$} & {$t=0.75$} & {$t=1$} 
\\ 
\midrule
{In-distribution}  & {$6.36\mathrm{e}{-2} \pm 5.37\mathrm{e}{-2}$} & {$1.29\mathrm{e}{-2} \pm 8.23\mathrm{e}{-3}$} & {$9.11\mathrm{e}{-3} \pm 6.10\mathrm{e}{-3}$} & {$8.78\mathrm{e}{-3} \pm 5.59\mathrm{e}{-3}$} & {$1.96\mathrm{e}{-2} \pm 1.52\mathrm{e}{-2}$}
\\
{Out-of-distribution 1}  & {$1.00\mathrm{e}{0} \pm 8.82\mathrm{e}{-1}$} & {$6.98\mathrm{e}{-1} \pm 5.87\mathrm{e}{-1}$} & {$5.60\mathrm{e}{-1} \pm 4.55\mathrm{e}{-1}$} & {$4.97\mathrm{e}{-1} \pm 4.56\mathrm{e}{-1}$} & {$4.52\mathrm{e}{-1} \pm 4.60\mathrm{e}{-1}$}
\\
{Out-of-distribution 2}  & {$1.01\mathrm{e}{0} \pm 1.03\mathrm{e}{0}$} & {$6.32\mathrm{e}{-1} \pm 5.38\mathrm{e}{-1}$} & {$5.02\mathrm{e}{-1} \pm 3.71\mathrm{e}{-1}$} & {$4.55\mathrm{e}{-1} \pm 4.00\mathrm{e}{-1}$} & {$4.19\mathrm{e}{-1} \pm 4.37\mathrm{e}{-1}$}
\\
{Out-of-distribution 3}  & {$1.14\mathrm{e}{0} \pm 4.39\mathrm{e}{-1}$} & {$8.39\mathrm{e}{-1} \pm 7.47\mathrm{e}{-1}$} & {$6.93\mathrm{e}{-1} \pm 6.13\mathrm{e}{-1}$} & {$6.00\mathrm{e}{-1} \pm 5.81\mathrm{e}{-1}$} & {$5.19\mathrm{e}{-1} \pm 5.82\mathrm{e}{-1}$}
\\
{Out-of-distribution 4}  & {$7.48\mathrm{e}{-1} \pm 6.16\mathrm{e}{-1}$} & {$5.17\mathrm{e}{-1} \pm 5.26\mathrm{e}{-1}$} & {$4.00\mathrm{e}{-1} \pm 3.97\mathrm{e}{-1}$} & {$3.24\mathrm{e}{-1} \pm 3.61\mathrm{e}{-1}$} & {$2.64\mathrm{e}{-1} \pm 3.57\mathrm{e}{-1}$}
\\

\midrule
{VAE-DLM unweighted (ours)}  & {$t=0$} & {$t=0.25$} & {$t=0.5$} & {$t=0.75$} & {$t=1$} 
\\ 
\midrule
{In-distribution}  & {$1.12\mathrm{e}{-1} \pm 6.87\mathrm{e}{-2}$} & {$4.37\mathrm{e}{-2} \pm 1.78\mathrm{e}{-2}$} & {$2.92\mathrm{e}{-2} \pm 1.13\mathrm{e}{-2}$} & {$2.33\mathrm{e}{-2} \pm 1.14\mathrm{e}{-2}$} & {$2.41\mathrm{e}{-2} \pm 1.55\mathrm{e}{-2}$}
\\
{Out-of-distribution 1}  & \textcolor{blue}{$\textBF{7.11\mathrm{e}{-1} \pm 4.94\mathrm{e}{-1}}$} & 
\textcolor{blue}{$\textBF{4.48\mathrm{e}{-1} \pm 3.92\mathrm{e}{-1}}$} & \textcolor{blue}{$\textBF{3.78\mathrm{e}{-1} \pm 3.35\mathrm{e}{-1}}$} & \textcolor{blue}{$\textBF{3.56\mathrm{e}{-1} \pm 3.50\mathrm{e}{-1}}$} & \textcolor{blue}{$\textBF{3.25\mathrm{e}{-1} \pm 3.52\mathrm{e}{-1}}$}
\\
{Out-of-distribution 2}  & {$\textBF{8.27\mathrm{e}{-1} \pm 8.69\mathrm{e}{-1}}$} & {$\textBF{5.39\mathrm{e}{-1} \pm 4.69\mathrm{e}{-1}}$} & {$\textBF{4.40\mathrm{e}{-1} \pm 3.66\mathrm{e}{-1}}$} & {$\textBF{4.01\mathrm{e}{-1} \pm 4.09\mathrm{e}{-1}}$} & {$\textBF{3.67\mathrm{e}{-1} \pm 4.39\mathrm{e}{-1}}$}
\\
{Out-of-distribution 3}  & \textcolor{blue}{$\textBF{7.10\mathrm{e}{-1} \pm 5.16\mathrm{e}{-1}}$} & \textcolor{blue}{$\textBF{5.93\mathrm{e}{-1} \pm 5.07\mathrm{e}{-1}}$} & {$\textBF{5.76\mathrm{e}{-1} \pm 5.15\mathrm{e}{-1}}$} & {$\textBF{5.29\mathrm{e}{-1} \pm 5.15\mathrm{e}{-1}}$} & {$\textBF{4.62\mathrm{e}{-1} \pm 5.15\mathrm{e}{-1}}$}
\\
{Out-of-distribution 4} & \textcolor{blue}{$\textBF{4.98\mathrm{e}{-1} \pm 3.51\mathrm{e}{-1}}$} & \textcolor{blue}{$\textBF{3.71\mathrm{e}{-1} \pm 3.63\mathrm{e}{-1}}$} & {$\textBF{3.20\mathrm{e}{-1} \pm 3.24\mathrm{e}{-1}}$} & {$\textBF{2.71\mathrm{e}{-1} \pm 3.15\mathrm{e}{-1}}$} & {$\textBF{2.25\mathrm{e}{-1} \pm 3.19\mathrm{e}{-1}}$}
\\

\bottomrule

\end{tabular}

\caption{We list relative $L^1$ errors of our method along with a VAE baseline on the Allen-Cahn equation on 30 identical (both noise and initial data) test samples. We choose scaling coefficients $(\alpha, \beta, \gamma_{\text{geo flow}}, \gamma_{\text{metric}}) = (100,1\mathrm{e}{-5},1,1)$. All hyperparameters in both our method and the traditional VAE are the same when applicable. Our OOD scenarios are: (1) 1.5 scaling of an in-distribution initial condition and noise with $\sigma = 3.0$ injected in 11 locations; (2) 1.3 scaling and noise with $\sigma=2.0$ injected in 51 locations; (3) absolute value noise with $\sigma=1.0$ injected in 151 locations; (4) 1.3 scaling with absolute value noise with $\sigma=0.5$ injected in all 201 locations.}

\end{table}

\begin{table}[htbp]
\centering

\scriptsize

\begin{tabular}{wl{3.0cm} | P{2.2cm} P{2.2cm} P{2.2cm} P{2.2cm} P{2.2cm} } 
\toprule
\belowrulesepcolor{light-gray} 
\rowcolor{light-gray} \multicolumn{6}{l} {Modified porous medium equation} 
\\ 
\aboverulesepcolor{light-gray} 
\midrule
{VAE}  & {$t=0$} & {$t=0.25$} & {$t=0.5$} & {$t=0.75$} & {$t=1$} 
\\ 
\midrule
{In-distribution}  & {$2.43\mathrm{e}{-1} \pm 1.13\mathrm{e}{-1}$} & {$8.47\mathrm{e}{-2} \pm 7.43\mathrm{e}{-2}$} & {$7.06\mathrm{e}{-2} \pm 6.67\mathrm{e}{-2}$} & {$6.52\mathrm{e}{-2} \pm 6.35\mathrm{e}{-2}$} & {$6.94\mathrm{e}{-2} \pm 6.50\mathrm{e}{-2}$}
\\
{Out-of-distribution 1}  & {$4.20\mathrm{e}{-1} \pm 3.12\mathrm{e}{-1}$} & {$2.89\mathrm{e}{-1} \pm 3.42\mathrm{e}{-1}$} & {$2.79\mathrm{e}{-1} \pm 3.50\mathrm{e}{-1}$} & {$2.76\mathrm{e}{-1} \pm 3.58\mathrm{e}{-1}$} & {$2.78\mathrm{e}{-1} \pm 3.67\mathrm{e}{-1}$}
\\
{Out-of-distribution 2}  & {$3.57\mathrm{e}{-1} \pm 1.79\mathrm{e}{-1}$} & {$2.16\mathrm{e}{-1} \pm 1.67\mathrm{e}{-1}$} & {$2.07\mathrm{e}{-1} \pm 1.68\mathrm{e}{-1}$} & {$2.08\mathrm{e}{-1} \pm 1.72\mathrm{e}{-1}$} & {$2.11\mathrm{e}{-1} \pm 1.79\mathrm{e}{-1}$}
\\
{Out-of-distribution 3}  & {$\textBF{4.40\mathrm{e}{-1} \pm 2.76\mathrm{e}{-1}}$} & {$\textBF{3.02\mathrm{e}{-1} \pm 2.69\mathrm{e}{-1}}$} & {$\textBF{2.95\mathrm{e}{-1} \pm 2.70\mathrm{e}{-1}}$} & {$\textBF{2.99\mathrm{e}{-1} \pm 2.74\mathrm{e}{-1}}$} & {$3.05\mathrm{e}{-1} \pm 2.82\mathrm{e}{-1}$}
\\
{Out-of-distribution 4}  & {$4.19\mathrm{e}{-1} \pm 1.83\mathrm{e}{-1}$} & {$2.62\mathrm{e}{-1} \pm 1.30\mathrm{e}{-1}$} & {$2.50\mathrm{e}{-1} \pm 1.21\mathrm{e}{-1}$} & {$2.45\mathrm{e}{-1} \pm 1.18\mathrm{e}{-1}$} & {$2.45\mathrm{e}{-1} \pm 1.16\mathrm{e}{-1}$}
\\

\midrule
{VAE, extended}  & {$t=0$} & {$t=0.25$} & {$t=0.5$} & {$t=0.75$} & {$t=1$} 
\\ 
\midrule
{In-distribution}  & {$2.22\mathrm{e}{-1} \pm 9.39\mathrm{e}{-2}$} & {$7.62\mathrm{e}{-2} \pm 6.02\mathrm{e}{-2}$} & {$5.60\mathrm{e}{-2} \pm 4.74\mathrm{e}{-2}$} & {$4.99\mathrm{e}{-2} \pm 4.55\mathrm{e}{-2}$} & {$5.08\mathrm{e}{-2} \pm 6.10\mathrm{e}{-2}$}
\\
{Out-of-distribution 1}  & {$4.00\mathrm{e}{-1} \pm 2.86\mathrm{e}{-1}$} & {$3.14\mathrm{e}{-1} \pm 3.34\mathrm{e}{-1}$} & {$3.17\mathrm{e}{-1} \pm 3.45\mathrm{e}{-1}$} & {$3.22\mathrm{e}{-1} \pm 3.56\mathrm{e}{-1}$} & {$3.29\mathrm{e}{-1} \pm 3.66\mathrm{e}{-1}$}
\\
{Out-of-distribution 2}  & {$\textBF{3.34\mathrm{e}{-1} \pm 1.88\mathrm{e}{-1}}$} & {$2.12\mathrm{e}{-1} \pm 2.19\mathrm{e}{-1}$} & {$2.05\mathrm{e}{-1} \pm 2.33\mathrm{e}{-1}$} & {$2.05\mathrm{e}{-1} \pm 2.45\mathrm{e}{-1}$} & {$2.08\mathrm{e}{-1} \pm 2.56\mathrm{e}{-1}$}
\\
{Out-of-distribution 3}  & {$4.85\mathrm{e}{-1} \pm 3.67\mathrm{e}{-1}$} & {$3.63\mathrm{e}{-1} \pm 3.99\mathrm{e}{-1}$} & {$3.59\mathrm{e}{-1} \pm 4.17\mathrm{e}{-1}$} & {$3.63\mathrm{e}{-1} \pm 4.40\mathrm{e}{-1}$} & {$3.69\mathrm{e}{-1} \pm 4.63\mathrm{e}{-1}$}
\\
{Out-of-distribution 4}  & {$4.07\mathrm{e}{-1} \pm 2.12\mathrm{e}{-1}$} & {$2.86\mathrm{e}{-1} \pm 1.92\mathrm{e}{-1}$} & {$2.76\mathrm{e}{-1} \pm 1.94\mathrm{e}{-1}$} & {$2.74\mathrm{e}{-1} \pm 1.95\mathrm{e}{-1}$} & {$2.72\mathrm{e}{-1} \pm 2.00\mathrm{e}{-1}$}
\\

\midrule
{VAE-DLM unweighted (ours)}  & {$t=0$} & {$t=0.25$} & {$t=0.5$} & {$t=0.75$} & {$t=1$} 
\\ 
\midrule
{In-distribution}  & {$2.59\mathrm{e}{-1} \pm 1.04\mathrm{e}{-1}$} & {$9.59\mathrm{e}{-2} \pm 6.73\mathrm{e}{-2}$} & {$8.18\mathrm{e}{-2} \pm 6.66\mathrm{e}{-2}$} & {$7.71\mathrm{e}{-2} \pm 6.55\mathrm{e}{-2}$} & {$7.95\mathrm{e}{-2} \pm 6.71\mathrm{e}{-2}$}
\\
{Out-of-distribution 1}  & {$4.01\mathrm{e}{-1} \pm 2.17\mathrm{e}{-1}$} & {$2.53\mathrm{e}{-1} \pm 2.51\mathrm{e}{-1}$} & {$2.43\mathrm{e}{-1} \pm 2.67\mathrm{e}{-1}$} & {$2.42\mathrm{e}{-1} \pm 2.84\mathrm{e}{-1}$} & {$2.43\mathrm{e}{-1} \pm 2.96\mathrm{e}{-1}$}
\\
{Out-of-distribution 2}  & {$3.48\mathrm{e}{-1} \pm 1.51\mathrm{e}{-1}$} & {$\textBF{1.86\mathrm{e}{-1} \pm 1.60\mathrm{e}{-1}}$} & {$\textBF{1.75\mathrm{e}{-1} \pm 1.75\mathrm{e}{-1}}$} & {$\textBF{1.73\mathrm{e}{-1} \pm 1.91\mathrm{e}{-1}}$} & {$\textBF{1.73\mathrm{e}{-1} \pm 2.01\mathrm{e}{-1}}$}
\\
{Out-of-distribution 3}  & {$4.52\mathrm{e}{-1} \pm 2.86\mathrm{e}{-1}$} & {$3.05\mathrm{e}{-1} \pm 3.07\mathrm{e}{-1}$} & {$3.00\mathrm{e}{-1} \pm 3.34\mathrm{e}{-1}$} & {$3.01\mathrm{e}{-1} \pm 3.57\mathrm{e}{-1}$} & {$\textBF{3.03\mathrm{e}{-1} \pm 3.71\mathrm{e}{-1}}$}
\\
{Out-of-distribution 4}  & {$\textBF{3.92\mathrm{e}{-1} \pm 1.53\mathrm{e}{-1}}$} & {$\textBF{2.22\mathrm{e}{-1} \pm 1.36\mathrm{e}{-1}}$} & {$\textBF{2.02\mathrm{e}{-1} \pm 1.41\mathrm{e}{-1}}$} & {$\textBF{1.97\mathrm{e}{-1} \pm 1.43\mathrm{e}{-1}}$} & {$\textBF{1.98\mathrm{e}{-1} \pm 1.43\mathrm{e}{-1}}$}
\\

\midrule
{VAE-DLM weighted (ours)}  & {$t=0$} & {$t=0.25$} & {$t=0.5$} & {$t=0.75$} & {$t=1$} 
\\ 
\midrule
{In-distribution}  & {$2.63\mathrm{e}{-1} \pm 1.12\mathrm{e}{-1}$} & {$9.88\mathrm{e}{-2} \pm 6.35\mathrm{e}{-2}$} & {$8.19\mathrm{e}{-2} \pm 6.45\mathrm{e}{-2}$} & {$7.47\mathrm{e}{-2} \pm 6.28\mathrm{e}{-2}$} & {$7.60\mathrm{e}{-2} \pm 5.84\mathrm{e}{-2}$}
\\
{Out-of-distribution 1}  & {$\textBF{3.92\mathrm{e}{-1} \pm 2.20\mathrm{e}{-1}}$} & {$\textBF{2.40\mathrm{e}{-1} \pm 2.48\mathrm{e}{-1}}$} & {$\textBF{2.34\mathrm{e}{-1} \pm 2.63\mathrm{e}{-1}}$} & {$\textBF{2.35\mathrm{e}{-1} \pm 2.76\mathrm{e}{-1}}$} & {$\textBF{2.33\mathrm{e}{-1} \pm 2.88\mathrm{e}{-1}}$}
\\
{Out-of-distribution 2}  & {$3.65\mathrm{e}{-1} \pm 1.60\mathrm{e}{-1}$} & {$2.11\mathrm{e}{-1} \pm 1.75\mathrm{e}{-1}$} & {$2.00\mathrm{e}{-1} \pm 1.88\mathrm{e}{-1}$} & {$1.98\mathrm{e}{-1} \pm 2.03\mathrm{e}{-1}$} & {$1.98\mathrm{e}{-1} \pm 2.16\mathrm{e}{-1}$}
\\
{Out-of-distribution 3}  & {$4.68\mathrm{e}{-1} \pm 2.87\mathrm{e}{-1}$} & {$3.22\mathrm{e}{-1} \pm 3.03\mathrm{e}{-1}$} & {$3.11\mathrm{e}{-1} \pm 3.24\mathrm{e}{-1}$} & {$3.08\mathrm{e}{-1} \pm 3.41\mathrm{e}{-1}$} & {$3.04\mathrm{e}{-1} \pm 3.51\mathrm{e}{-1}$}
\\
{Out-of-distribution 4}  & {$4.09\mathrm{e}{-1} \pm 1.92\mathrm{e}{-1}$} & {$2.41\mathrm{e}{-1} \pm 1.54\mathrm{e}{-1}$} & {$2.26\mathrm{e}{-1} \pm 1.64\mathrm{e}{-1}$} & {$2.18\mathrm{e}{-1} \pm 1.62\mathrm{e}{-1}$} & {$2.09\mathrm{e}{-1} \pm 1.57\mathrm{e}{-1}$}
\\

\bottomrule

\end{tabular}

\caption{We list relative $L^1$ errors of our method along with a VAE baseline on the PM equation on 30 identical (both noise and initial data) test samples. We choose scaling coefficients $(\alpha, \beta, \gamma_{\text{geo flow}}, \gamma_{\text{metric}}) = (100,0.01,1,1)$. All hyperparameters in both our method and the traditional VAE are the same when applicable. Our OOD scenarios are: (1) no scaling of an-in-distribution initial condition and noise with $\sigma=0.45$ injected in all 201 locations; (2) noise with $\sigma=0.45$ injected in 101 locations; (3) noise with $\sigma = 0.95$ injected in 51 locations; (4) 1.2 scaling and noise with $\sigma=0.95$ injected in 21 locations. We remark we did find higher variation among training iterations in this experiment across both our method and baselines, even though we used a large batch size of 1,000. For example, we found VAE-extended error was as high as $3.85\mathrm{e}{-1}$ for the OOD 1, $t=1$ case. In this sense, we remark our errors here tend to characterize the lower end of errors across training iterations. We also remark we did not use an eigenvalue penalization here, which improved performance slightly.}

\label{tab:PM}

\end{table}

\subsection{Kuramoto-Sivashinsky equation}
\label{sec:KS_eq}

In this section, we consider the Kuramoto-Sivashinsky equation, given by 
\begin{equation}
\begin{cases}
\partial_t \phi + \nu_1 \phi \partial_x \phi + \nu_2 \partial_{xx} \phi + \nu_3 \partial_{xxxx} \phi = 0 , (x,t) \in [-6,6] \times [0,1] ,
\\
\phi(x,0) = \cos(x) ( 1 + \alpha_0 \sin(x ) ), \alpha_0 \sim U([-1,1]) .
\end{cases}
\end{equation}
We choose coefficients $(\nu_1, \nu_2, \nu_3) = (5,0.5,0.005)$, which is similar to the experiments of \cite{wang2022respectingcausalityneedtraining}. We train on 800 solutions.

\vspace{2mm}

Data for this experiment is oscillatory and difficult to learn. This is resolved with a Fourier feature \cite{tancik2020fourierfeaturesletnetworks} embedding into the encoder network, where we take the mapping
\begin{align}
& t \rightarrow ( \cos( 2 \pi B t), \sin( 2 \pi  B t) ), \ \ \ \ \ [B]_{ij} \sim N(0,\sigma^2)
\\
& \mathcal{E}_{\theta_{\mathcal{E}}} : ( u, \cos( 2 \pi B t), \sin( 2 \pi  B t) ) \rightarrow \mathbb{R}^d ,
\end{align}
where $B \in \mathbb{M}^{q \times 1}$ for some dimension $q$, which achieves order of magnitude lower training loss in a significant reduction in training time. We take $\sigma = 4$.

\section{Conclusions and limitations}

We advance the variational autoencoder paradigm for time-dependent, non-stochastic data dynamics by introducing latent dynamics with a geometric flow. Our methods are computationally efficient in the offline stage by choice of a simple geometric flow. Our method scales for moderately high latent dimension and allows more versatile latent dynamics and helps accompany more diverse initial conditions. Our method advances strategies towards robustness in the learning of dynamics; our experiments have emphasized steady dynamics and more extreme cases of robustness. We find our methods on our chosen datasets perform at least as well the vanilla VAE and the extended VAE with high frequency on most datasets. In scenarios of dynamics with lesser variation as time increases, we find our methods outperform baselines consistently. On select other datasets, our methods do not necessarily prevail, and can perform equivalently to baselines. We find our methods rarely underperform a baseline.

\vspace{2mm}

While our method can accommodate considerable diversity in initial conditions, alternative methods including Transformer architectures \cite{wang2024bridgingoperatorlearningconditioned}, state-space models \cite{hu2024statespacemodelsaccurateefficient}, and neural operators tend to generalize more effectively for input data. Also, our method is mesh-dependent. A potential solution is this is by introducing a spatial input into the VAE system. We remark alternative paradigms have particular emphasis on low in-distribution error, and mitigating failures through robustness is an area of current study and deserve further attention.

\vspace{2mm}

Solving high-dimensional physics informed neural networks (PINNs) \cite{Hu_2024} is a hard task. Some research is poured into this direction but for more generalized PDEs. Our choice of geometric flow is highly specialized, and because of this, the complexity of the learning task is greatly reduced. It may be of interest to consider more generalized geometric flows, in which the area of solving high-dimensional PINNs should be emphasized.

\vspace{2mm}

The primary point of inquiry in this research is whether or not disparate geometric properties in latent spaces in encoder-decoder type methods have different outcomes. We have empirically demonstrated certain PDEs perform well with certain latent characteristics; however, the notion that the role of intrinsic properties in the PDE data are intrinsically associated with the geometries of the latent space is an immediate question of interest.

\bibliographystyle{plainnat}
\bibliography{bibliography}

\appendix

\newpage

\section{Additional PDEs for experiments}

As a supplemental experiment, we also consider the Korteweg-De Vries (KdV) equation in this experiment, where we consider
\begin{equation}
\begin{cases}
\partial_t \phi(x,t) = - \phi(x,t) \partial_x \phi(x,t) - \delta^2 \partial_{xxx} \phi(x,t) , (x,t) \in [0,2] \times [0,1],
\\
\phi(x,0) = \alpha_1 \cos(\pi x) + \alpha_2 \cos^3(\pi x), \alpha_i \sim U([-1,1]),
\end{cases}
\end{equation}
where $\delta \in \mathbb{R}$ is a small constant (we choose $\delta = 0.022$), and $(x,t) \in [0,2] \times[0,1]$ discretized into a $201 \times 201$ mesh. We use the Fourier feature embedding for this experiment with $\sigma=2$. We find our manifold method had negligible impact on this experiment, and results were effectively equivalent to baselines.

\section{Riemannian metric calculation}
\label{app:metric_vs_changeofvar}

Let $d-1$ denote the intrinsic dimension, and $m$ the extrinsic. Let $\partial_i$ denote the partial derivative with respect to $i$ and $\mathcal{E}_j$ the $j$-th component of $\mathcal{E} : \mathbb{R}^{d-1} \rightarrow \mathbb{R}^m$. $\mathcal{E}$ is typically computed with neural network $\mathcal{E}_{\theta_{\mathcal{E}}}$.

\vspace{2mm}

The Jacobian is the quantity
\begin{equation}
\frac{\partial \mathcal{E}}{\partial u}(u,t) = J =  \begin{pmatrix}
\partial_1 \mathcal{E}_1(u,t) & \partial_2 \mathcal{E}_1(u,t)  &  \hdots & \partial_{d-1} \mathcal{E}_1(u,t) \\
\partial_1 \mathcal{E}_2(u,t) & \ddots & & \vdots
\\
\vdots & & \\
\partial_{1} \mathcal{E}_{m}(u,t) & \hdots & & \partial_{d-1} \mathcal{E}_m(u,t)
\end{pmatrix}
\in \mathbb{M}^{m \times d-1} .
\end{equation}

The Riemannian metric is computed as
\begin{align}
J^T J & = \begin{pmatrix}
\partial_1 \mathcal{E}_1 & \partial_1 \mathcal{E}_2  &  \hdots & \partial_{1} \mathcal{E}_m \\
\partial_2 \mathcal{E}_1 & \ddots & & \vdots
\\
\vdots & & \\
\partial_{d-1} \mathcal{E}_{1} & \hdots & & \partial_{d-1} \mathcal{E}_m
\end{pmatrix}
\begin{pmatrix}
\partial_1 \mathcal{E}_1 & \partial_2 \mathcal{E}_1  &  \hdots & \partial_{d-1} \mathcal{E}_1 \\
\partial_1 \mathcal{E}_2 & \ddots & & \vdots
\\
\vdots & &
\\
\partial_{1} \mathcal{E}_m & \hdots & & \partial_{d-1} \mathcal{E}_m
\end{pmatrix}
\\
& = \begin{pmatrix}
||\partial_1 \mathcal{E} ||_2^2 & \langle \partial_1 \mathcal{E}, \partial_2 \mathcal{E} \rangle  &  \hdots & \langle \partial_1 \mathcal{E}, \partial_{d-1} \mathcal{E} \rangle
\\
\langle \partial_2 \mathcal{E}, \partial_1 \mathcal{E} \rangle & || \partial_2 \mathcal{E} ||_2^2 & & \vdots
\\
\vdots & & \ddots &
\\
\langle \partial_m \mathcal{E}, \partial_1 \mathcal{E} \rangle & \hdots & & || \partial_{d-1} \mathcal{E} ||_2^2
\end{pmatrix}
\approx g \in \mathbb{R}^{(d-1) \times (d-1)} .
\end{align}

\section{Weighted objective and derivation of weighted Kullback-Leibler divergence}
\label{App:modified_KL_div}

\subsection{The weighted objective}

We also work with the \textcolor{blue}{\textbf{weighted}} objective, which is as follows:
\begin{align}
 \EX_{t \sim U[0,T], u \sim \tilde{q}(\cdot|x_0)} [   &  \underbrace{ -  ( \alpha  \log p(x_t|z_t) +  \beta  \mathcal{J}(u,t) \log ( \frac{ p(u) }{ \tilde{q}(u|x_0)} ) )  }_{\text{ELBO-type loss}}
\\ & + \underbrace{ \gamma_{\text{geo flow}} || (\partial_t +  \Lambda_{\theta_{\Lambda}}(u,t) ) g_{\theta_g}(u,t) + \alpha ( g_{\theta_g}(u,t) - \Sigma(u) )  ||_F^2 }_{\text{physics-informed geometric flow}}
 + 
  \underbrace{ \gamma_{\text{metric}} || g_{\theta_g}(u,t) -
J^T J  ||_F^2 }_{\text{matching metric with manifold}} ] .
\end{align}
As mentioned, we have found empirical success in taking $\mathcal{J}(u,t) = \sqrt{\text{det}(J^T(u,t) J(u,t))}$ for the weighted loss. We find this scaled loss can outperform the unweighted method in certain settings (see Tables \ref{tab:burgers}, \ref{tab:PM}), but not particularly significantly. We do not scale log-likelihood, but we penalize the KL divergence term by the geometric distortion induced by the manifold mapping. Regions of greater geometric distortion are regularized more, and regions of lower contribution to distortion are penalized less and are allowed greater freedom in what is learned. In particular, more geometrically distorted regions are more clustered. Note that this regularization primarily affects the learned parameterization. We consider this weighting for Burger's equation ($\beta$ is taken low) and the modified porous medium equation ($\beta$ is taken high).

\vspace{2mm}

We can normalize the weighting terms, as this ensures a more balanced scaling of the loss. We do this by taking
\begin{equation} \beta_{\text{scaled}} = \frac{N \beta}{\sum_{i=1}^N  \mathcal{J}(u_i, t_i)} =  \frac{N \beta}{\sum_{i=1}^N \sqrt{ \text{det}(J^T(u_i, t_i) J(u_i, t_i) )}},
\end{equation}
where the summation runs over the batch. With this scaling, we normalize the coefficients with the means of the variable tranformation. We take $ \beta_{\text{scaled}}$ as our coefficient instead of $\beta$.

\vspace{2mm}

The primary mathematical motivation for the weighted KL divergence is the introduction of a Jacobian term under a change of variables. Using the map $(u,t) \rightarrow z_t$ with square Jacobian, we have
\begin{equation}
D_{KL}( p(z_t) || q(z_t)) = \int p(z_t) \log( \frac{ p(z_t) }{ q(z_t) } ) dz_t = \int p(u) | \text{det}(J(u,t)) |^{-1} \log ( \frac{ p(u) }{q(u)} ) | \text{det}(J(u,t)) | du .
\end{equation}
Thus, the two Jacobian terms cancel out. The weighted KL divergence is of consideration to incorporate this loss without the inclusion $|\text{det}(J(u,t))|^{-1}$, thus there is mathematical significance.

\subsection{Weighted KL divergence derivation}

We derive the modified KL divergence loss as in our loss function. Recall we want to minimize 
\begin{equation}
\EX_{t \sim U[0,T]} [ D_{KL}^{\text{weighted}} ( \tilde{q}(u | x_0) || p(u) ) ] = - \EX_{t \sim U[0,T]} \EX_{u \sim \tilde{q}(\cdot|x_0)} [ \mathcal{J}   \log ( \frac{ p(u) }{ \tilde{q}(u|x_0)} )  ] .
\end{equation}
We choose prior $p(u) \sim N(0, I)$. First, we let
\begin{equation}
\mathcal{J}  \geq M ,
\end{equation}
where the lower bound $M$ is by guaranteed since $\mathcal{J} \geq 0$ (i.e. we can always take $M=0$ at the least). This is necessary since an empirical computation relying on the mean of the posterior is intractable. Since both distributions are Gaussians, in particular since $p \sim N(0,I)$, we have
\begin{align}
& \EX_t \EX_{\tilde{q}} [ \mathcal{J} \log(\tilde{q}) - \mathcal{J} \log (p) ]
\\
& = \underbrace{\EX_t \EX_{\tilde{q}} [ \frac{\mathcal{J}}{2} \log ( \frac{ |\Sigma_{p}|   }{| \Sigma_{\tilde{q}}|} )  ]}_{(1)}  - \underbrace{\EX_t \EX_{\tilde{q}}[ \frac{\mathcal{J}}{2}(u - \mu_{\tilde{q}}) \Sigma_{\tilde{q}}^{-1} (u - \mu_{\tilde{q}}) ] }_{(2)} + \underbrace{\EX_t \EX_{\tilde{q}} [ \frac{\mathcal{J}}{2}(u - \mu_p)^T \Sigma_p^{-1} (u - \mu_p) ]}_{(3)} .
\end{align}
We examine the terms individually. We have
\begin{equation}
(1) = \EX_t \EX_{\tilde{q}} [ \frac{\mathcal{J}}{2} \log ( \frac{ |\Sigma_{p}|   }{| \Sigma_{\tilde{q}}|} )  ] = \frac{1}{2} \log ( \frac{ |\Sigma_{p}|   }{| \Sigma_{\tilde{q}}|} )   \EX_t \EX_{\tilde{q}} [ \mathcal{J} ] = - \frac{1}{2} \log( | \Sigma_{\tilde{q}} | ) \EX_t \EX_{\tilde{q}} [ \mathcal{J}] ,
\end{equation}
where we use $\Sigma_p = I$. We have
\begin{align}
\label{eqn:weighted_kl_second_term}
(2) = \EX_t \EX_{\tilde{q}}[ \frac{\mathcal{J}}{2}(u - \mu_{\tilde{q}}) \Sigma_{\tilde{q}}^{-1} (u - \mu_{\tilde{q}}) ] & = \frac{1}{2} \EX_t \EX_{\tilde{q}} [ \mathcal{J} \ \text{Tr}( u - \mu_{\tilde{q}})^T \Sigma_{\tilde{q}}^{-1} ( u - \mu_{\tilde{q}})] .
\end{align}
Now, note
\begin{equation}
(u - \mu_{\tilde{q}} ) \Sigma_{\tilde{q}}^{-1} ( u - \mu_{\tilde{q}}) \geq 0, \ \ \ \ \ \Sigma_{\tilde{q}}^{-1} \succeq 0 ,
\end{equation}
since $\Sigma_{\tilde{q}}$ is diagonal with positive entries, thus its inverse is positive semi-definite. Hence, since $\mathcal{J} \geq 0$, we must have
\begin{align}
\ref{eqn:weighted_kl_second_term} & \geq \frac{1}{2} \text{Tr} ( \EX_t \EX_{\tilde{q}} [ M \ ( u - \mu_{\tilde{q}})  ( u - \mu_{\tilde{q}})^T ] \Sigma_{\tilde{q}}^{-1} ) = \frac{M}{2} \text{Tr} ( \Sigma_{\tilde{q}} \Sigma_{\tilde{q}}^{-1} ) = \frac{dM}{2} .
\end{align}
The last term bound follows using Hölder's inequality for expectations. Also, we assume low variance on $\mathcal{J}$. Since $(\mu_p, \Sigma_p) = (0,I)$, we have
\begin{align}
(3) =  \EX_t \EX_{\tilde{q}} [ \frac{\mathcal{J}}{2}(u - \mu_p)^T \Sigma_p^{-1} (u - \mu_p) ] & = \frac{1}{2} \EX_t \EX_{\tilde{q}} [ \mathcal{J} u^T u ]
\\
& \leq \frac{1}{2} ( \EX_t \EX_{\tilde{q}} [ \mathcal{J}^2 ] )^{1/2} ( \EX_t \EX_{\tilde{q}} [ ||u||^4] )^{1/2}
\\
& = \frac{1}{2} ( \EX_t \EX_{\tilde{q}} [ \mathcal{J}^2 ] )^{1/2} ( 2 \text{Tr}(\Sigma_{\tilde{q}}^2) + 4 \mu_{\tilde{q}}^T \Sigma_{\tilde{q}} \mu_{\tilde{q}}  + ( \text{Tr}(\Sigma_{\tilde{q}}) + \mu_{\tilde{q}}^T \mu_{\tilde{q}} )^2 )^{1/2}
\\
& \approx \frac{1}{2}  \EX_t \EX_{\tilde{q}} [ \mathcal{J} ]  ( 2 \text{Tr}(\Sigma_{\tilde{q}}^2) + 4 \mu_{\tilde{q}}^T \Sigma_{\tilde{q}} \mu_{\tilde{q}}  + ( \text{Tr}(\Sigma_{\tilde{q}}) + \mu_{\tilde{q}}^T \mu_{\tilde{q}} )^2 )^{1/2} ,
\end{align}
where we use low variance in the last line. The final term is simplified using the formula, when $y \sim N(\mu, \Sigma)$,
\begin{equation}
\EX [ y^T y y^T y ] = \EX [ || y||^4] = 2 \text{Tr}(\Sigma^2) + 4 \mu^T \Sigma \mu  + ( \text{Tr}(\Sigma) + \mu^T \mu )^2 ,
\end{equation}
which is found in \cite{petersen2012matrix}.

Now, we apply numerical properties to solve the above. $M$ can be approximated by taking the minimum of $\mathcal{J}$ over the batch.  Also, recall $\Sigma_{\tilde{q}}$ is diagonal. We have
\begin{align}
& \text{loss} \approx 
\frac{1}{2 N} \sum_{j=1}^N  ( \mathcal{J}(u_j,t_j) ( - \log( \prod_i  
 \sigma_{\tilde{q},i}^2 )  ) - dM + \mathcal{J}(u_j,t_j) ( 2 \text{Tr}(\Sigma_{\tilde{q}}^2) + 4 \mu_{\tilde{q}}^T \Sigma_{\tilde{q}} \mu_{\tilde{q}}  + ( \text{Tr}(\Sigma_{\tilde{q}}) + \mu_{\tilde{q}}^T \mu_{\tilde{q}} )^2  )^{1/2} ) 
\\
 & = \frac{1}{2 N} \sum_{j=1}^N (\mathcal{J}(u_j,t_j)  ( - \sum_i   \log (  \sigma_{\tilde{q},j,i}^2 ) ) - dM + \mathcal{J}(u_j,t_j) (2 \sum_i \sigma_{\tilde{q},j,i}^4 + 4 \sum_i \mu_{\tilde{q},j,i}^2 \sigma_{\tilde{q},j,i}^2  + ( \sum_i \sigma_{\tilde{q},j,i}^2 + \mu_{\tilde{q},j,i}^2 )^2 )^{1/2} ) 
\end{align}
where $j$ runs over the batch and $i$ runs over intrinsic dimension. In order to achieve this bound, accuracy is lost by $\mathcal{J} \geq M$, by applying the Hölder's inequality, and by assuming the variance of $\mathcal{J}$ is low. Since scaling coefficient $\beta$ is (often) taken quite small, this loss of accuracy is not important.

\section{Gradient flow discussion}
\label{app:grad_flow_disc}

\textbf{Definition.} Let $g \in \mathbb{M}^{d \times d}$. We say the functional derivative is the function $\delta \mathcal{F}(x) / \delta g$ such that the first variation can $\delta \mathcal{F}[g, \phi]$ be expressed as
\begin{equation}
\delta \mathcal{F}(g, \phi) = \sum_{ij} \int_{\mathcal{U}} [\frac{ \delta \mathcal{F}}{\delta g}]_{ij}(u) \phi_{ij} (u) du = \int_{\mathcal{U}}  \sum_{ij} [\frac{ \delta \mathcal{F}}{\delta g}]_{ij}(u) \phi_{ij} (u) du  = \int_{\mathcal{U}} \langle \frac{\delta \mathcal{F}}{\delta g}, \phi \rangle_F du .
\end{equation}

\vspace{2mm}

Generally, we operate under conditions so that the exchange of integration and summation is justified.

\vspace{2mm}

\textbf{Remark.} Often, the above is expressed as a typical product between real-valued functions. Since $g, \phi$ are matrices, the above is a natural extension of such a definition but for matrix inputs. Using the formulation with an inner product is discussed in \cite{kamalinejad2012optimal}.

\vspace{2mm}

\textbf{Definition.} We say $g \in \mathbb{M}^{d \times d}$ follows a gradient flow if it satisfies the equation
\begin{equation}
\partial_t g = - \frac{ \delta \mathcal{F}}{\delta g} 
\end{equation}
with respect to the Euclidean metric and the functional $\mathcal{F}$.

\vspace{2mm}

\textbf{Remark.}  The definition we constructed for a gradient flow is a natural definition with respect to functionals. A well known example of the above gradient flow formulation is that the Laplacian is the functional derivative of the Dirichlet energy, which gives the gradient flow equation of the heat equation \cite{liu_gradient_flow} \cite{yu2020repulsivecurves}.

\vspace{2mm}

\textbf{Theorem 1.} The geometric flow of equation \ref{eqn:lin_flow} is a gradient flow with respect to the functional
\begin{equation}
\mathcal{F}(g) = \int_{\mathcal{U}} \frac{1}{2} \langle  (\Lambda + \alpha I) g , g\rangle - \alpha \langle \Sigma, g \rangle du .
\end{equation}

\vspace{2mm}

\textit{Proof.} Let $\phi$ be a perturbation metric (so it is symmetric, as perturbations preserve properties of the original functions). Observe the gradient flow takes the form
\begin{equation}
\partial_t g = - \frac{ \delta \mathcal{F}}{\delta g} ,  \ \ \ \  \  \frac{d}{d \epsilon} \mathcal{F}(g + \epsilon \phi) |_{\epsilon = 0} = \int_{\mathcal{U}}  \langle \frac{\delta \mathcal{F}}{\delta g}, \phi \rangle_F du .
\end{equation}
Now, given our functional,
\begin{align}
\mathcal{F}[g+\epsilon \phi] & =  \int_{\mathcal{U}}  \frac{1}{2} \langle (\Lambda + \alpha I) ( g + \epsilon \phi ) , g + \epsilon \phi \rangle - \alpha \langle \Sigma, g + \epsilon \phi \rangle du \\
& =   \int_{\mathcal{U}}  \frac{1}{2} 
\epsilon \langle (\Lambda + \alpha I)g, \phi \rangle + \frac{1}{2} 
\epsilon \langle (\Lambda + \alpha I) \phi, g \rangle  - \alpha \epsilon \langle \Sigma, \phi \rangle + f_{\epsilon^2} + f_{\epsilon^{\perp}} du  ,
\end{align}
where $f_{\epsilon^2}$ is a second-order function of $\epsilon$, and $f_{\epsilon^{\perp}}$ is independent of $\epsilon$. Differentiating and evaluating at $\epsilon = 0$,
\begin{equation}
\label{eqn:first_variation}
\frac{d}{d \epsilon} \mathcal{F}[g + \epsilon \phi] |_{\epsilon = 0} = \int_{\mathcal{U}}  \frac{1}{2}  \langle (\Lambda + \alpha I)g , \phi \rangle + \frac{1}{2}  \langle (\Lambda + \alpha I) \phi, g \rangle  - \alpha \langle \Sigma, \phi \rangle du  \stackrel{(1)}{=} \int_{\mathcal{U}}  \langle  ( \Lambda + \alpha I ) g - \alpha \Sigma, \phi \rangle du, 
\end{equation}
and we conclude
\begin{equation}
\frac{ \delta \mathcal{F}}{\delta g} =  \Lambda g + \alpha ( g - \Sigma )
\end{equation}
as desired. The equality (1) in \ref{eqn:first_variation} is because
\begin{align}
\langle (\Lambda + \alpha I) \phi, g  \rangle & = \sum_i \sum_j [(\Lambda + \alpha I) \phi]_{ij} g_{ij} 
= \sum_i \sum_j ( \sum_k [  \Lambda + \alpha I]_{ik} \phi_{kj} ) g_{ij} 
\\
& =  \sum_j \sum_k (\sum_i [ \Lambda + \alpha I]_{ik} g_{ij} ) \phi_{kj} 
= \sum_j \sum_k [ (\Lambda + \alpha I) g ]_{kj} \phi_{kj} = \langle ( \Lambda + \alpha I) g, \phi \rangle , 
\end{align}
where we also use the matrices are symmetric.

$ \square $

\vspace{2mm}

\textbf{Theorem 2.} The functional defined by \ref{eqn:functional} corresponding to the linear geometric flow of equation \ref{eqn:lin_flow} is nonincreasing on $t > 0$ if 
\begin{equation}
\frac{1}{2} \langle \Lambda_t g, g \rangle_F \leq || ( \Lambda + \alpha I)g - \alpha \Sigma ||_F^2 .
\end{equation}

\vspace{2mm}

\textit{Proof.} It suffices to differentiate the functional in time. Assume continuity and continuous differentiability on $\Lambda, g, \Sigma$ over $\mathcal{U}$. First, note
\begin{equation}
\int_{\mathcal{U}} \frac{1}{2}  \langle (\Lambda  + \alpha I )g,g\rangle - \alpha \langle \Sigma, g \rangle du < + \infty ,
\end{equation}
and the hypotheses of the Leibniz integral rule are satisfied. Observe
\allowdisplaybreaks
\begin{align}
\frac{d}{dt} \mathcal{F}[g]
& = \frac{d}{dt}  \int_{\mathcal{U}} \frac{1}{2}  \langle (\Lambda  + \alpha I )g,g\rangle - \alpha \langle \Sigma, g \rangle du
\\
& \stackrel{(1)}{=}  \int_{\mathcal{U}} \frac{1}{2} 
\frac{d}{dt} \langle (\Lambda  + \alpha I )g,g\rangle - \alpha \frac{d}{dt} \langle \Sigma, g \rangle du
\\
& =  \int_{\mathcal{U}} \frac{1}{2} 
\langle \Lambda_t g, g \rangle + \frac{1}{2} \langle (\Lambda + \alpha I) g_t, g \rangle + \frac{1}{2}  \langle (\Lambda + \alpha I ) g, g_t \rangle - \alpha \langle \Sigma, g_t \rangle du
\\
& \stackrel{(2)}{=}  \int_{\mathcal{U}} \frac{1}{2} 
\langle \Lambda_t g, g \rangle +  \langle (\Lambda + \alpha I ) g, -(\Lambda + \alpha I) g + \alpha \Sigma \rangle  - \alpha \langle \Sigma, -(\Lambda + \alpha I) g + \alpha \Sigma \rangle  du
\\
& = \int_{\mathcal{U}} \frac{1}{2} \langle \Lambda_t g, g \rangle - \langle  (\Lambda + \alpha I ) g - \alpha \Sigma, (\Lambda + \alpha I ) g - \alpha \Sigma \rangle du
\\
& =  \int_{\mathcal{U}} \frac{1}{2} \langle \Lambda_t g, g \rangle - || (\Lambda + \alpha I ) g - \alpha \Sigma ||_F^2 du .
\end{align}
(1) is by the Leibniz integral rule, and (2) is by using the PDE. If this integrand is negative over its entire domain, the integral is negative, and we have our result.

$\square$

\vspace{2mm}

\textbf{Theorem 3.} The nonlinear geometric flow of equation \ref{eqn:nonlinear_geo_flow} is a gradient flow if
\begin{equation}
\text{mat}( \text{Tr} (\frac{\partial \varphi}{\partial \Lambda}(\Lambda(u,t,g),g) 
\frac{\partial \Lambda^T}{\partial g_{ij} })) + \frac{\partial \varphi}{\partial g}(\Lambda(u,t,g),g) = \Lambda(u,t,g) .
\end{equation}

\vspace{2mm}

\textit{Proof.} Suppose that the geometric flow
\begin{equation}
\partial_t g(u,t) = -\Lambda(u,t,g(u,t)) 
\end{equation}
is a gradient flow. Then it can be expressed as
\begin{equation}
\int_{\mathcal{U}} \langle \frac{ \partial \mathcal{F}}{\partial g}, h \rangle du = \int_{\mathcal{U}} \langle \Lambda, h \rangle du 
\end{equation}
for some functional $\mathcal{F}$ such that
\begin{equation}
\mathcal{F}[g] = \int_{\mathcal{U}} \varphi(\Lambda(u,t,g(u,t)), g(u,t)) du
\end{equation}
for some real-valued, nonnegative function $\varphi : \mathbb{M}^{d \times d} \times \mathbb{M}^{d \times d} \rightarrow \mathbb{R}^+$. Now, consider a perturbation
\begin{equation}
\mathcal{F}[g + \epsilon \phi] =  \int_{\mathcal{U}} \varphi(\Lambda(u,t,g + \epsilon \phi), g + \epsilon \phi) du  .
\end{equation}
Now, we differentiate and evaluate at $\epsilon = 0$. Using the chain rule, we have
\begin{equation}
\frac{d}{d \epsilon} \mathcal{F}[g + \epsilon \phi] |_{\epsilon = 0 } = \int_{\mathcal{U}}  \langle \text{mat}( \text{Tr} (\frac{\partial \varphi}{\partial \Lambda}(\Lambda(u,t,g),g) 
\frac{\partial \Lambda^T}{\partial g_{ij} })), \phi \rangle
+  \langle \frac{\partial \varphi}{\partial g}(\Lambda(u,t,g),g) , \phi \rangle du ,
\end{equation}
where the first term is done with the rules found in \cite{nel1980matrix} (see page 150). We use notation $\text{mat}(\cdot)$ as the concatenation of the $ij$ elements into a matrix. As a side note, observe the transpose is redundant since $\Lambda, g$ are symmetric. Since $\phi$ is an arbitrary test perturbation function, we must have
\begin{equation}
\text{mat}( \text{Tr} (\frac{\partial \varphi}{\partial \Lambda}(\Lambda(u,t,g),g) 
\frac{\partial \Lambda^T}{\partial g_{ij} })) + \frac{\partial \varphi}{\partial g}(\Lambda(u,t,g),g) = \Lambda(u,t,g) ,
\end{equation}
and we are done.

$\square$

\section{High in-distribution performance figures} 

We provide figures of our methods on our various datasets. By high in-distribution performance, we enforce a low KL regularization ($\beta \leq 1\mathrm{e}{-4})$.

\begin{figure}
  \vspace{0mm}
  \centering
  \includegraphics[scale=0.55]{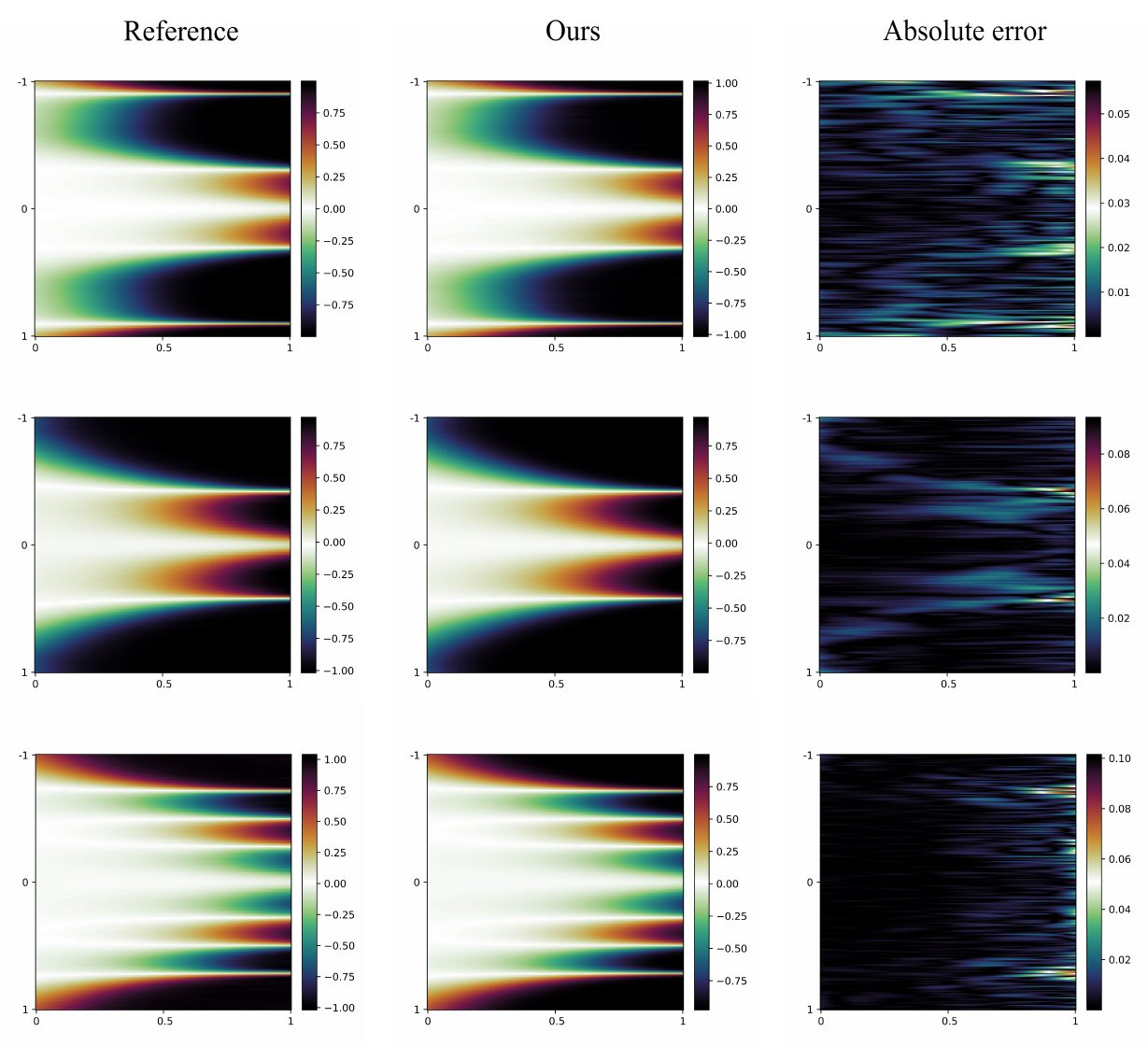}
  \caption{We examine three examples of our predicted versus reference solutions of the Allen-Cahn equation on the high in-distribution data performance setting.}
  \label{fig:AllenCahn_eq}
\end{figure}

\begin{figure}
  \vspace{0mm}
  \centering
  \includegraphics[scale=0.55]{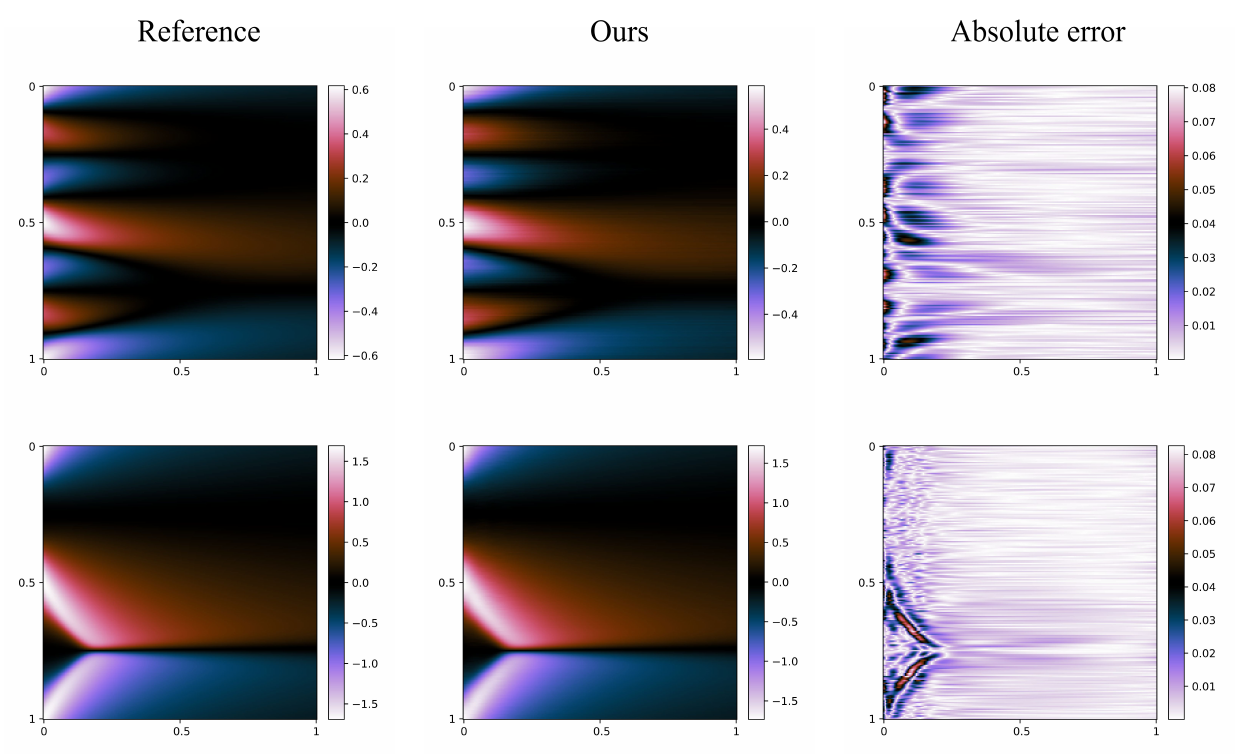}
  \caption{We examine two examples of our predicted versus reference solutions of the Burger's equation on the high in-distribution data performance setting.}
  \label{fig:AllenCahn_eq}
\end{figure}

\begin{figure}
  \vspace{0mm}
  \centering
  \includegraphics[scale=0.55]{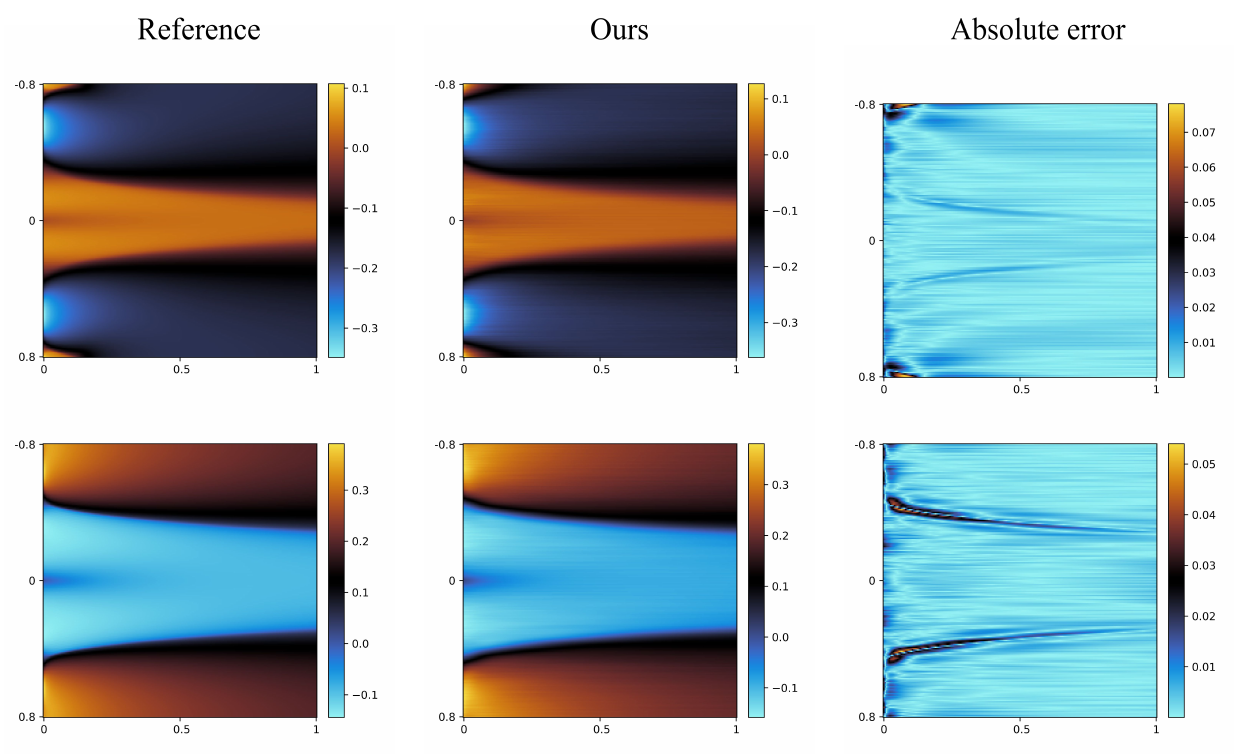}
  \caption{We examine two examples of our predicted versus reference solutions of the modified porous medium equation on the high in-distribution data performance setting.}
  \label{fig:AllenCahn_eq}
\end{figure}

\begin{figure}
  \vspace{0mm}
  \centering
  \includegraphics[scale=0.55]{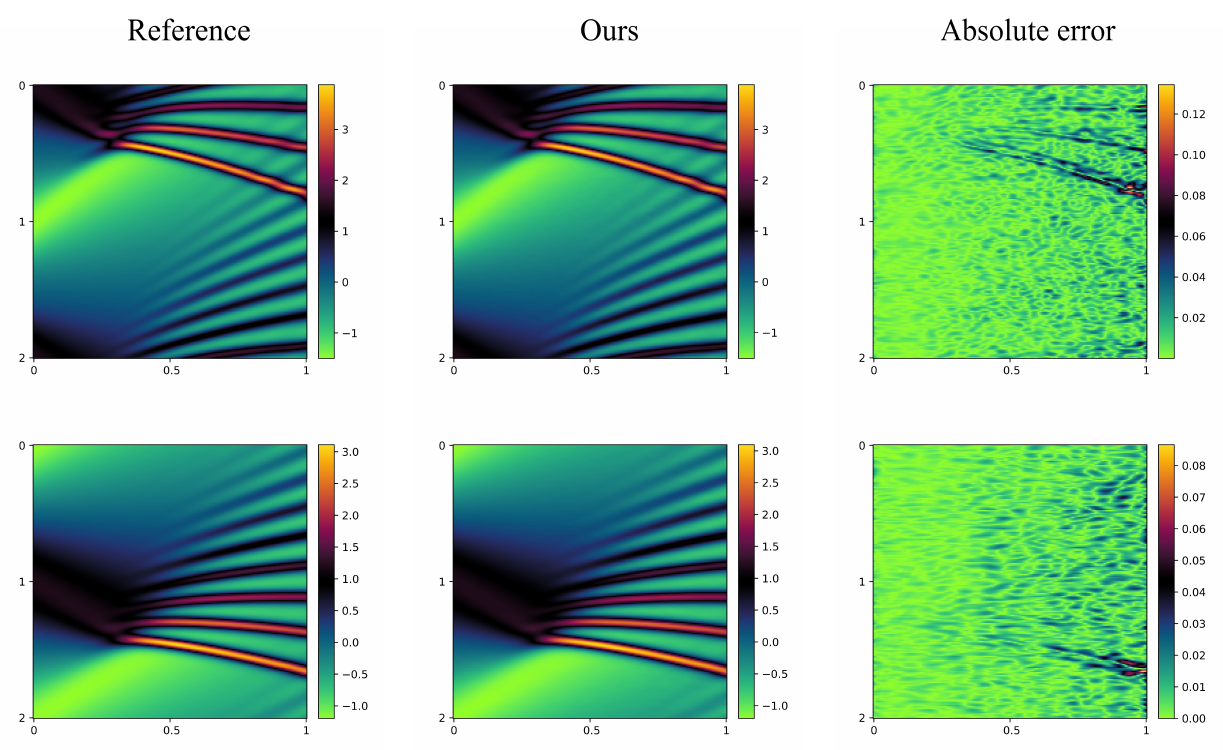}
  \caption{We examine two examples of our predicted versus reference solutions of the Korteweg-De Vries equation on the high in-distribution data performance setting.}
  \label{fig:KS_eq}
\end{figure}

\begin{figure}[t]
  \vspace{0mm}
  \centering
  \includegraphics[scale=0.55]{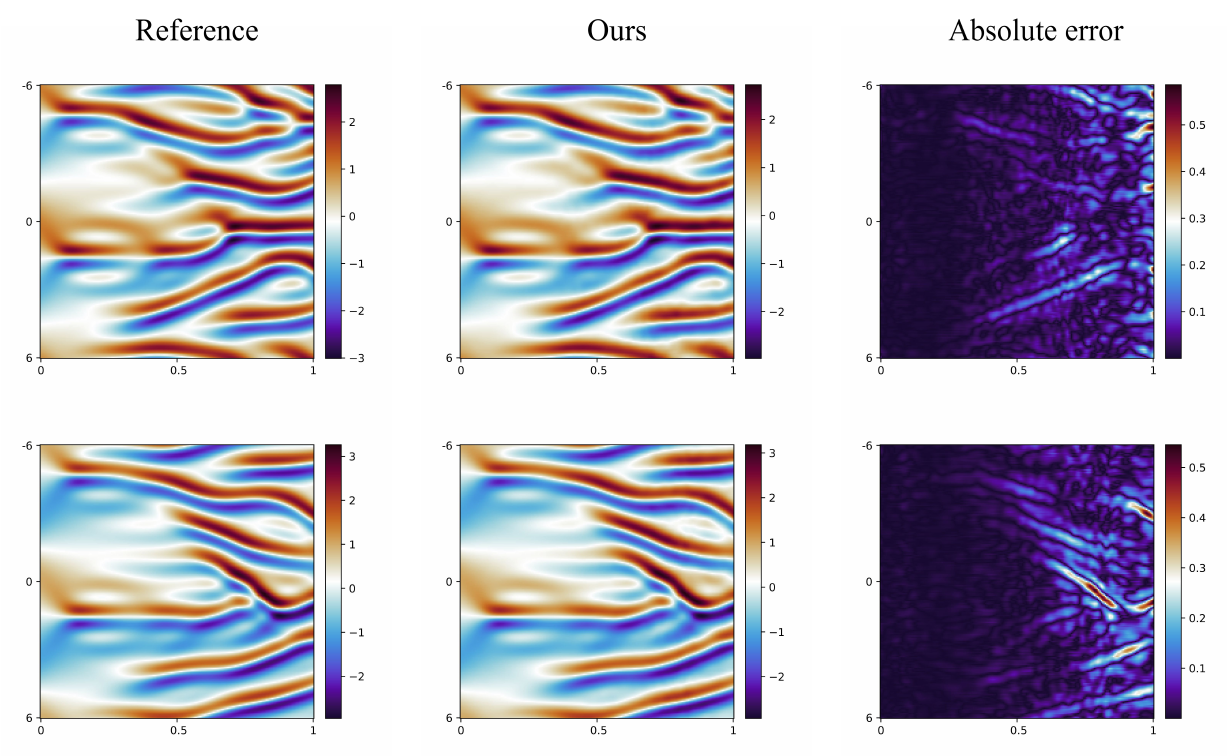}
  \caption{We examine two examples of our predicted versus reference solutions of the Kuramoto-Sivashinsky equation on the high in-distribution data performance setting.}
  \label{fig:KS_eq}
\end{figure}

\vfill

\clearpage
\newpage

\section{Additional robustness figures}
\label{app:robust_figs}

\begin{figure}[H]
  \vspace{0mm}
  \centering
  \includegraphics[scale=0.55]{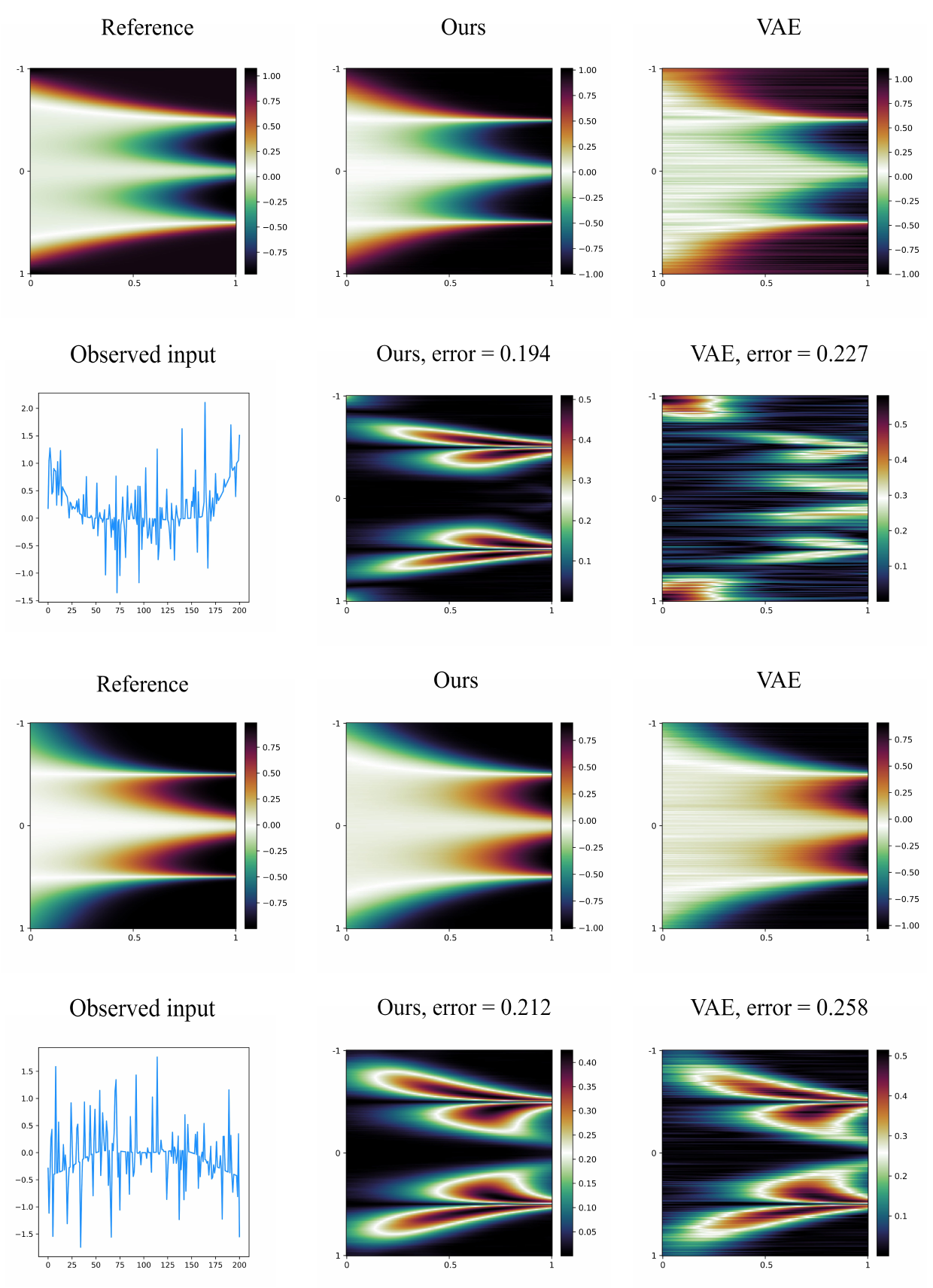}
  \caption{We examine OOD results on our method compared with a vanilla VAE on the Allen-Cahn equation. The two settings are: \textit{(top)} an in-distribution initial condition injected with noise with $\sigma=0.5$ in 101 locations; \textit{(bottom)} an in-distribution inital condition injected with noise with $\sigma=0.75$ in 101 locations. We assess our error using relative $L^1$ double integral error with respect to both space and time over $(x,t) \in [-1,1]\times[0,1]$ (note the error graphical portion of the figure displays absolute value error). Please see the error table for results over numerous data.}
  \label{fig:AllenCahn_OOD_eq}
\end{figure}

\begin{figure}[htbp]
  \vspace{0mm}
  \centering
  \includegraphics[scale=0.55]{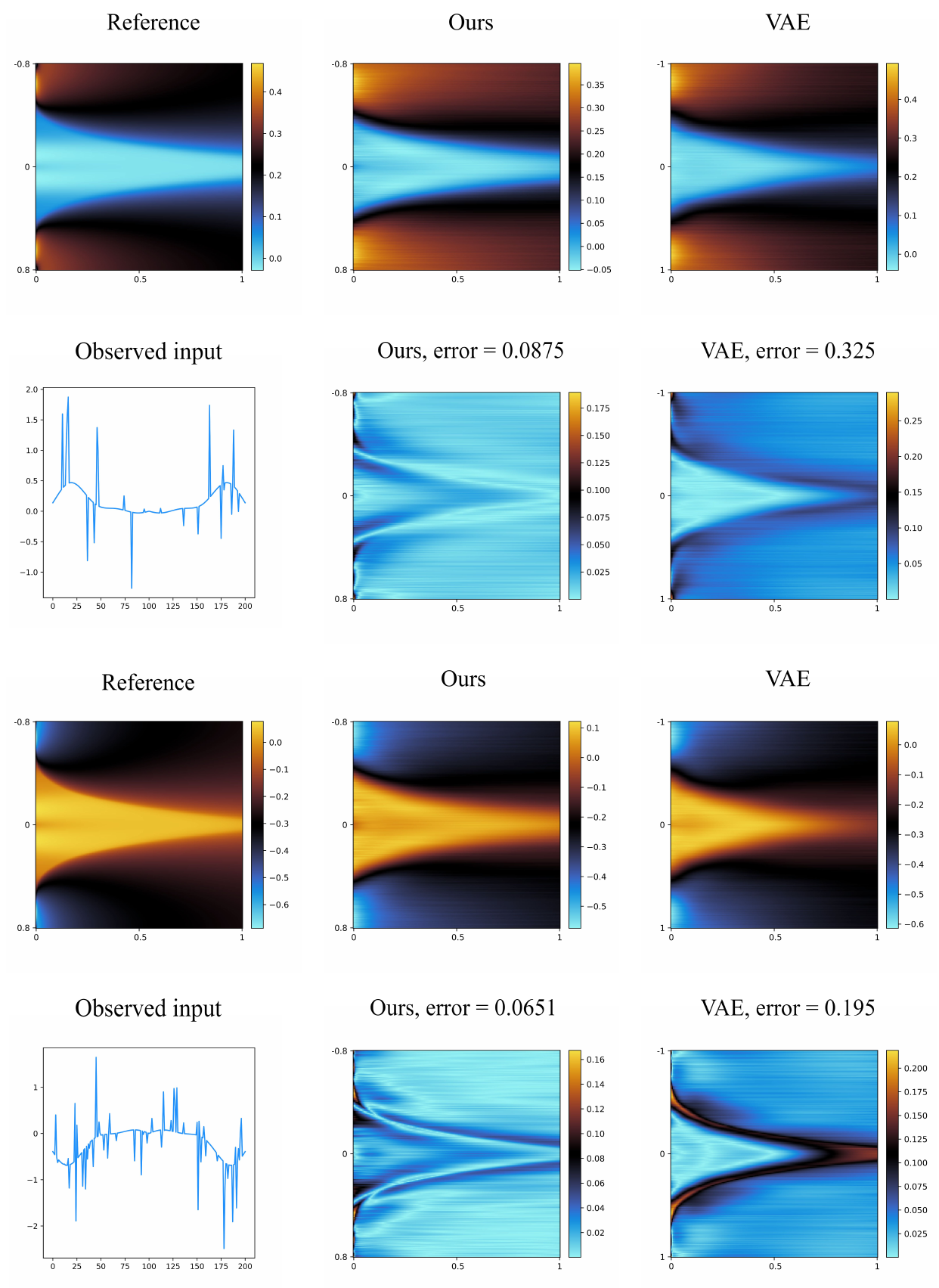}
  \caption{We examine OOD results on our method compared with a vanilla VAE on the porous medium equation. The two settings are: \textit{(top)} an in-distribution initial condition with noise with $\sigma=0.95$ injected in 21 locations; \textit{(bottom)} an in-distribution inital condition injected with noise with $\sigma=0.65$ in all 51 locations. We assess our error using relative $L^1$ double integral error with respect to both space and time over $(x,t) \in [-0.8,0.8] \times[0,1]$ (note the error graphical portion of the figure displays absolute value error). We emphasize, for this figure especially, these are some of our best results, as the error reduction is by a large percentage. Please see the error table for results over numerous data.}
  \label{fig:AllenCahn_OOD_eq}
\end{figure}

\begin{figure}[htbp]
  \vspace{0mm}
  \centering
  \includegraphics[scale=0.55]{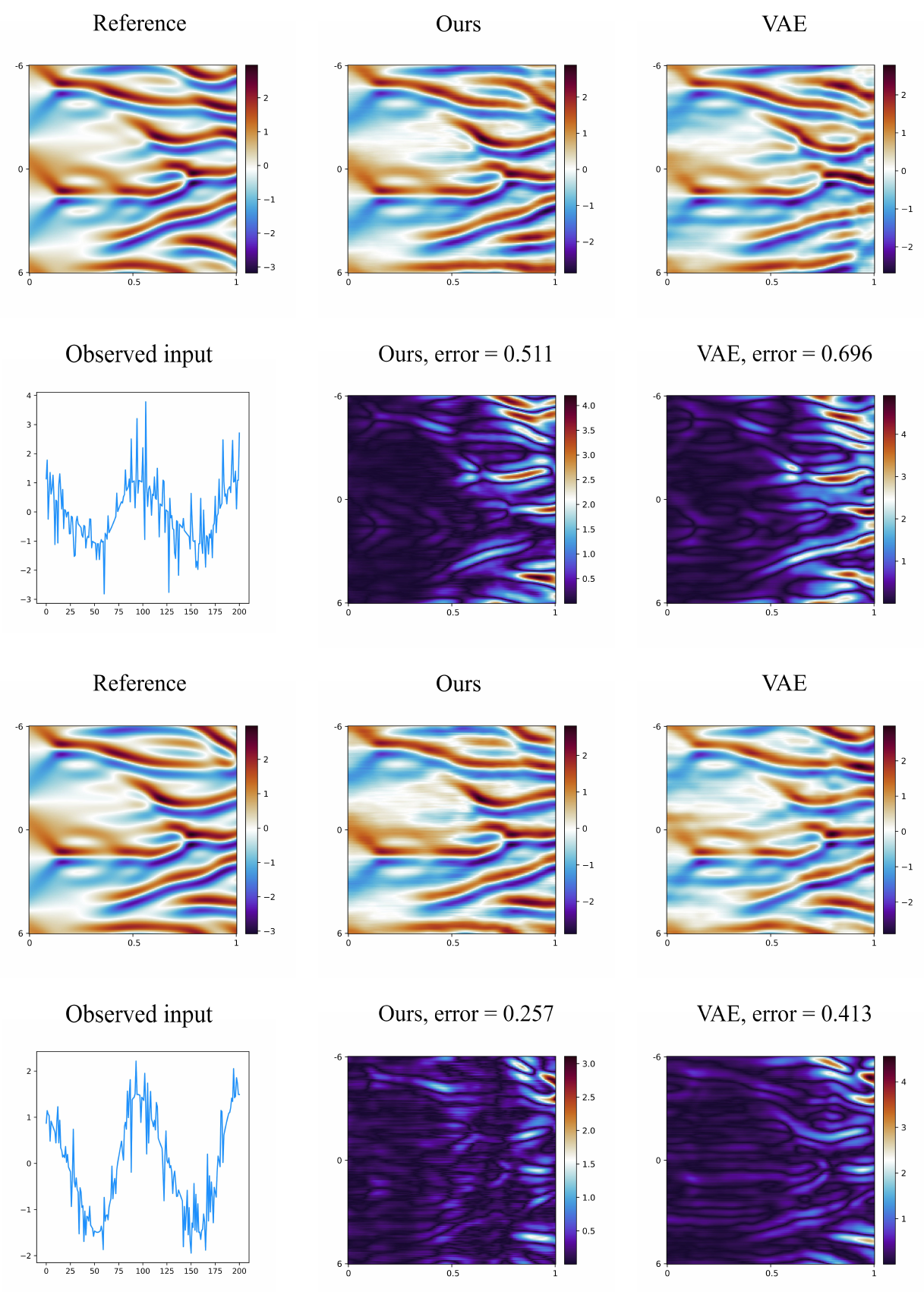}
  \caption{We examine OOD results on our method compared with a vanilla VAE on the KS equation. The two settings are: \textit{(top)} an in-distribution initial condition injected with noise with $\sigma=0.95$ in 101 locations; \textit{(bottom)} an in-distribution inital condition scaled by 1.4 and noise with $\sigma=0.45$ injected in 101 locations. We assess our error using relative $L^1$ double integral error with respect to both space and time over $(x,t) \in [-6,6]\times[0,1]$ (note the error graphical portion of the figure displays absolute value error). Please see the error table for results over numerous data.}
  \label{fig:AllenCahn_OOD_eq}
\end{figure}

\clearpage

\newpage

\section{Additional figures}
\label{app:additional_figs}

\subsection{KdV equation generative modeling}
\label{app:gen_modeling}

\begin{figure}[H]
  \vspace{0mm}
  \centering
  \includegraphics[scale=0.6]{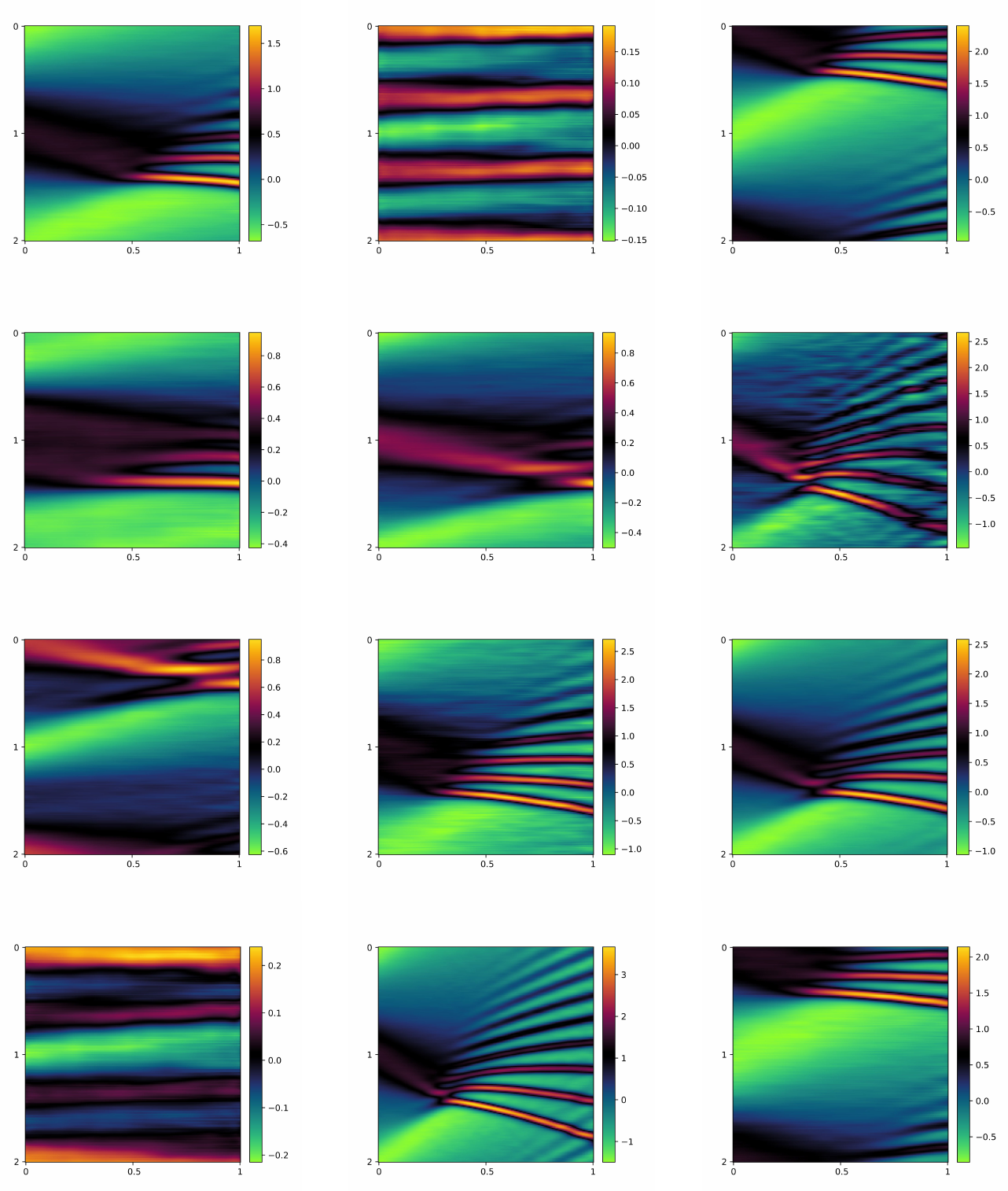}
  \caption{We provide uncurated samples of results on the KdV equation via generative modeling. The generation is done by sampling the parameterization prior with a standard normal Gaussian. We found heavy KL divergence regularization to be necessary to achieve higher quality samples, and so we take $(\alpha, \beta, \gamma_{\text{geo flow}}, \gamma_{\text{metric}}) = (1000,1,1,1)$ in this experiment. We also emphasize the manifold is only loosely enforced with these coefficients, and enforcing the manifold even more comes at a cost of sample quality.}
  \label{fig:KdV_eq_generative}
\end{figure}

\newpage

\subsection{KS equation generative modeling}

\begin{figure}[H]
  \vspace{0mm}
  \centering
  \includegraphics[scale=0.6]{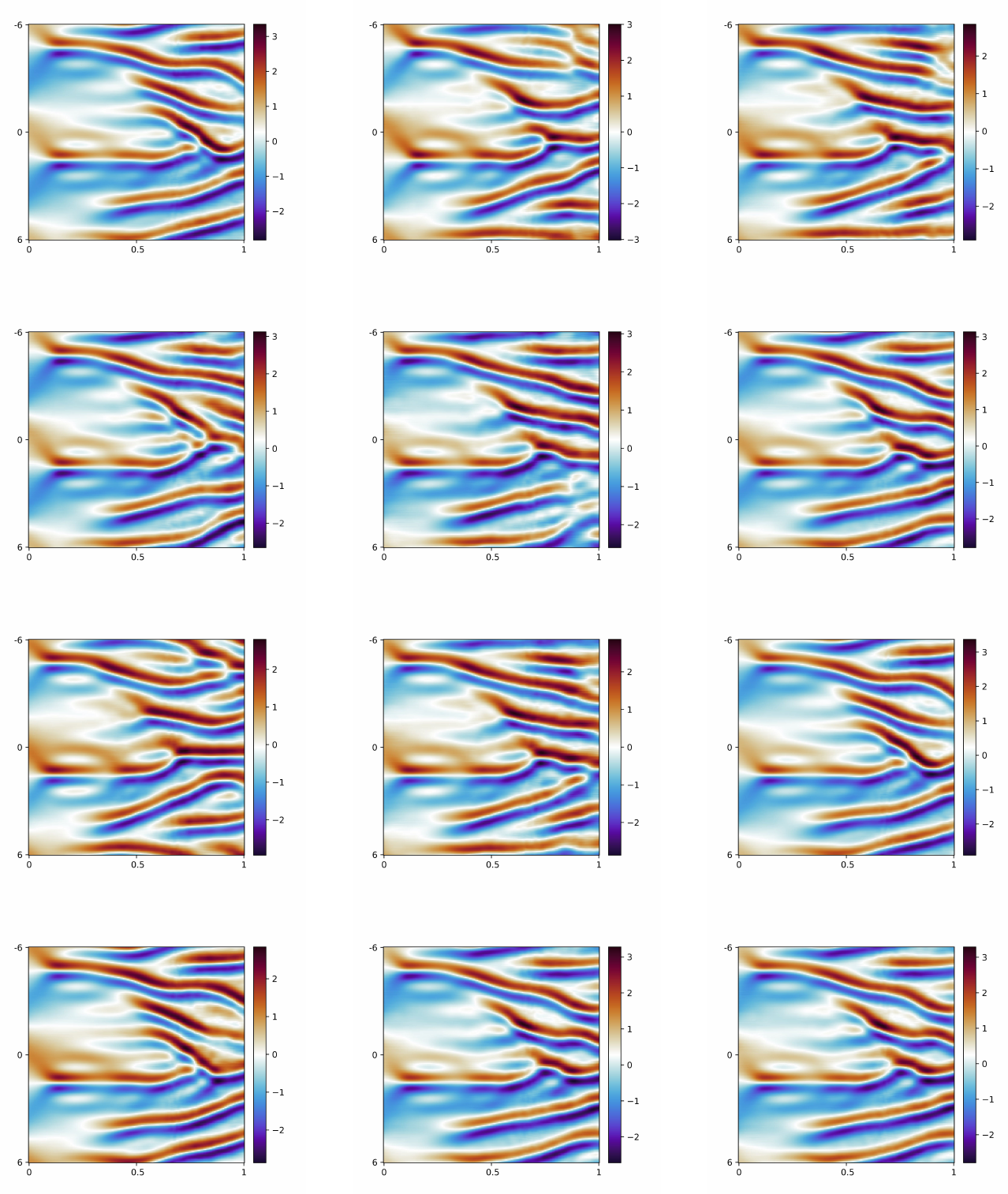}
  \caption{We provide uncurated samples of results on the KS equation via generative modeling. The generation is done by sampling the parameterization prior with a standard normal Gaussian. We take $(\alpha, \beta, \gamma_{\text{geo flow}}, \gamma_{\text{metric}}) = (1000,1,1,1)$ in this experiment. We also emphasize the manifold is only loosely enforced with these coefficients, and enforcing the manifold even more comes at a cost of sample quality.}
  \label{fig:KS_eq_generative}
\end{figure}

\newpage

\subsection{Manifold figure}

\begin{figure}[!htb]
  \vspace{0mm}
  \centering
  \includegraphics[scale=0.6]{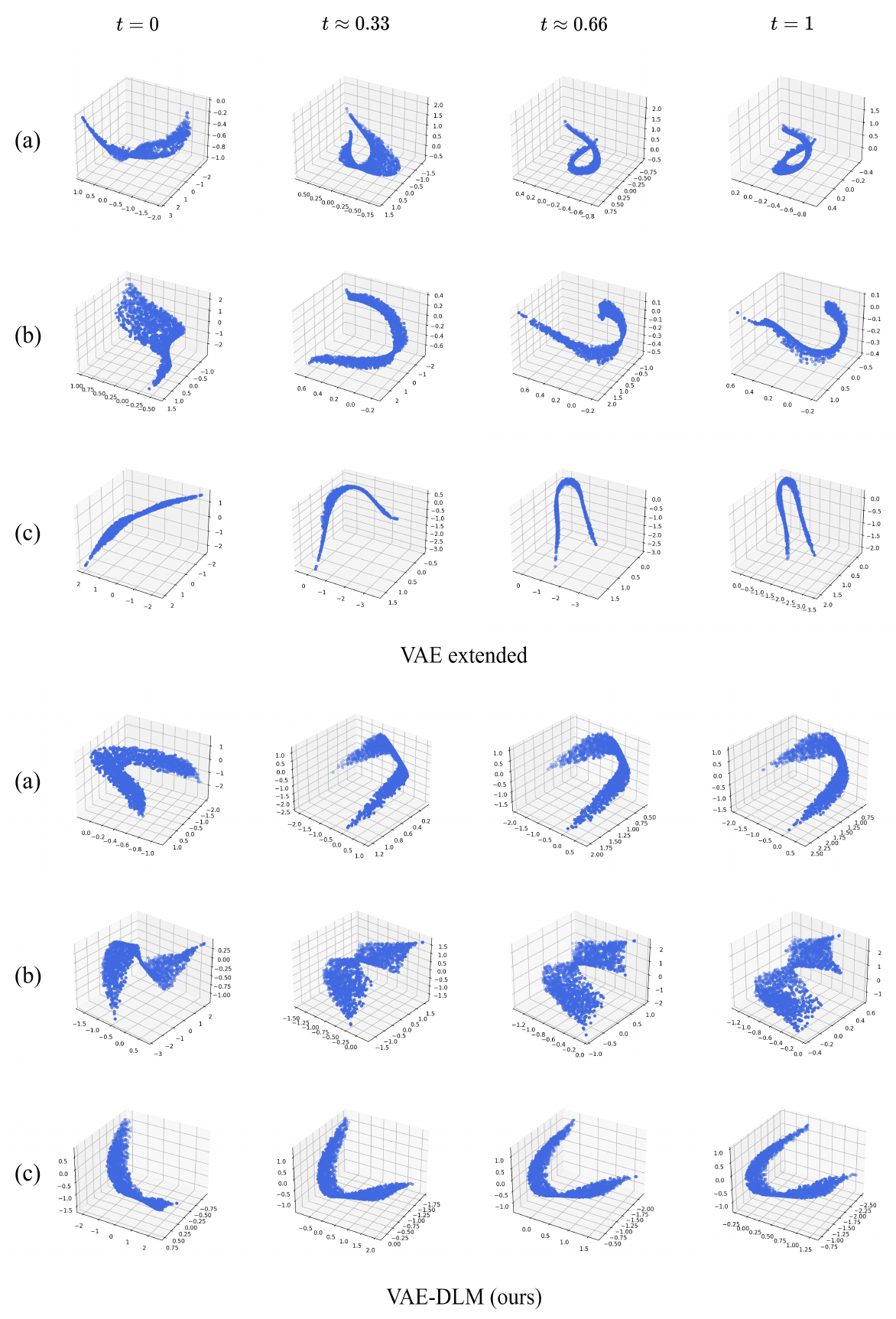}
  \caption{We illustrate the learned latent spaces on three instances of training on the Burger's equation with the VAE extended and our method. We examine the latent space in the $\mathcal{E}$ stage. We choose intrinsic dimension $2$ for visualization purposes. As we can see, both latent spaces can effectively be categorized as manifolds despite no restrictions being placed on the latent space in the VAE extended setting; however, our latent space organizes into a more canonical geometry, and in turn helps robustness and additional properties in the ambient data.}
  \label{fig:manifold_comparison}
\end{figure}

\newpage

\FloatBarrier
\subsection{Examining the learned metric}

\begin{figure}[!htb]
  \vspace{0mm}
  \centering
  \includegraphics[scale=0.5]{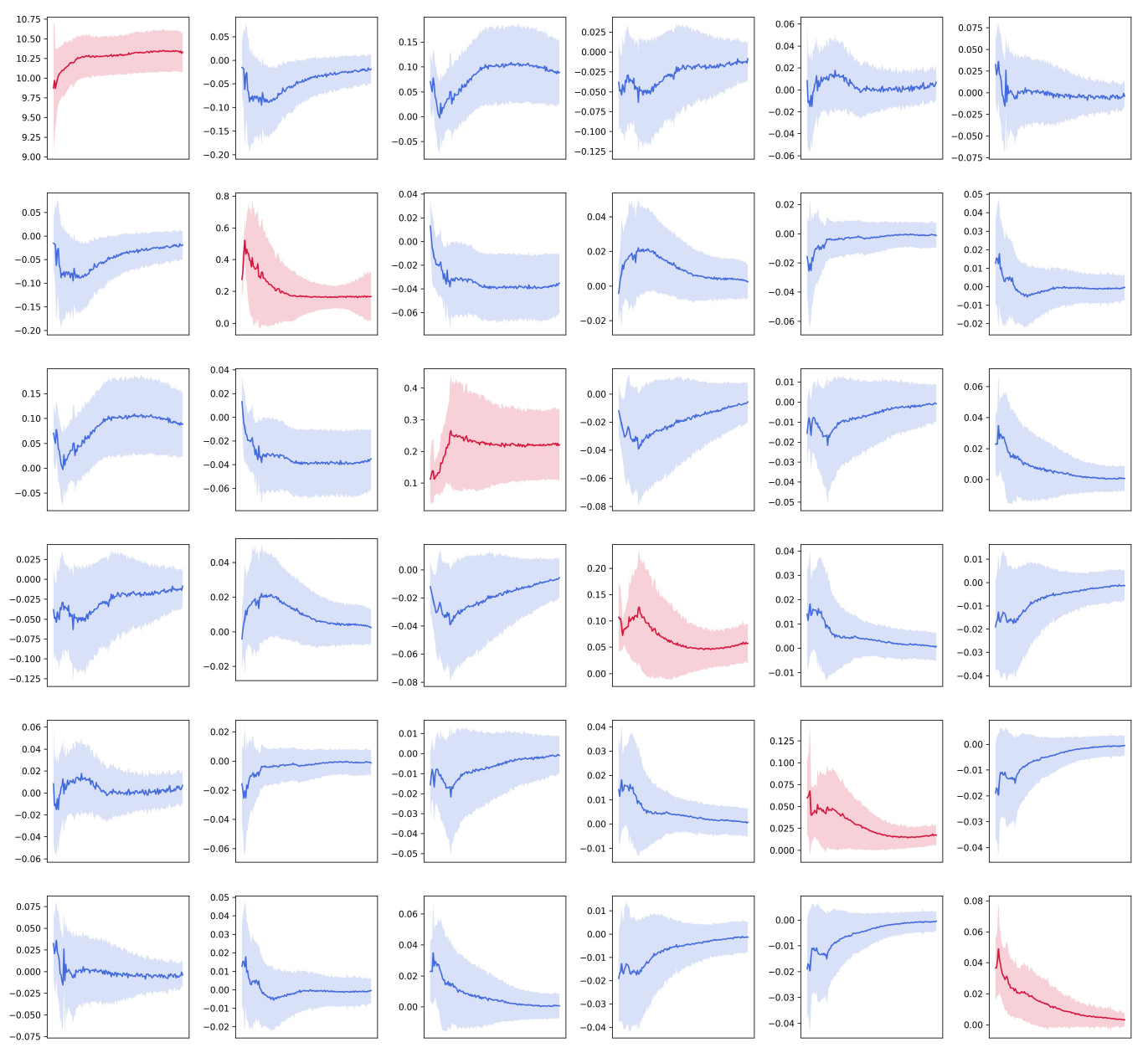}
  \caption{We plot the matrix of the learned Riemannian metric $g_{\theta_g}$ with respect to training iteration on our VAE-DLM proposed method. We plot the means of the metric over the batch with the standard deviation of each element. This experiment was upon the Allen-Cahn equation. We use a steady state with $\alpha=2$ and the spherical metric with $r^2 = 10$. We include the eigenvalue penalization. Also, note we favor KL divergence regularization with the prior in this experiment, as this yields slightly more diversity in the learned parameterization domain. Observe the matrix is $6 \times 6$, which corresponds to $(\text{intrinsic dimension} \times \text{intrinsic dimension})$, so we have $6 \times 6$ plots. We distinguish the diagonal elements of the metric (red) from the off-diagonal (blue). Also note the matrix is symmetric. Clearly from the plot, our metric somewhat resembles that of a perturbed sphere, which is a canonical geometry. This is most observable in the $(1,1)$-element, as the metric of the sphere here is exactly $10$ in this element.}
  \label{fig:AllenCahn_metric_no_reg}
\end{figure}

\clearpage
\begin{figure}[!htb]
  \vspace{0mm}
  \centering
  \includegraphics[scale=0.5]{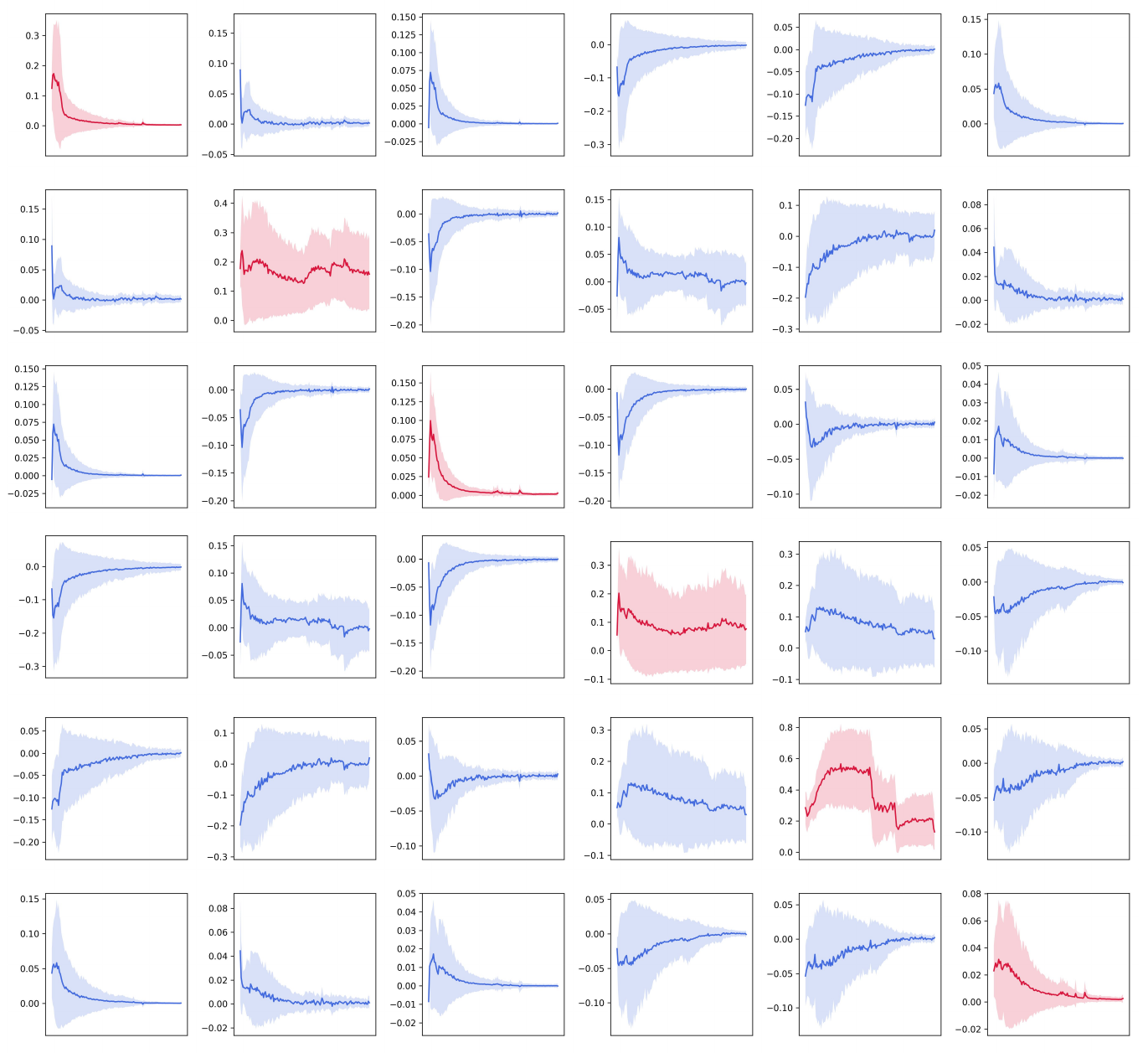}
  \caption{We plot the matrix of the learned Riemannian metric $g_{\theta_g}$ with respect to training iteration using a non-steady state geometric flow of the form $\partial_t g = - \Lambda g$. We plot the means of the metric over the batch with the standard deviation of each element. This experiment was upon the Allen-Cahn equation. There is no eigenvalue penalization. Also, note we favor KL divergence regularization with the prior in this experiment, as this yields slightly more diversity in the learned parameterization domain. Observe the matrix is $6 \times 6$, which corresponds to $(\text{intrinsic dimension} \times \text{intrinsic dimension})$, so we have $6 \times 6$ plots. We distinguish the diagonal elements of the metric (red) from the off-diagonal (blue). We emphasize most elements of this learned Riemannian metric tend to $0$ or small values, and so a considerably degenerate manifold is learned: this is a primary reason we include the steady state term with eigenvalue penalization. }
  \label{fig:AllenCahn_metric_no_reg}
\end{figure}

\clearpage

\newpage

\FloatBarrier
\section{Additional tables for experimental results}

 \vfill

\begin{table}[!htbp]
\centering

\scriptsize

\begin{tabular}{wl{3.0cm} | P{2.2cm} P{2.2cm} P{2.2cm} P{2.2cm} P{2.2cm} } 
\toprule
\belowrulesepcolor{light-gray} 
\rowcolor{light-gray} \multicolumn{6}{l} {Kuramoto-Sivashinsky equation} 
\\ 
\aboverulesepcolor{light-gray} 
\midrule
{VAE}  & {$t=0$} & {$t=0.25$} & {$t=0.5$} & {$t=0.75$} & {$t=1$} 
\\ 
\midrule
{In-distribution}  & {$4.17\mathrm{e}{-2} \pm 1.19\mathrm{e}{-2}$} & {$2.54\mathrm{e}{-2} \pm 9.69\mathrm{e}{-3}$} & {$2.05\mathrm{e}{-2} \pm 7.24\mathrm{e}{-3}$} & {$5.13\mathrm{e}{-2} \pm 7.92\mathrm{e}{-2}$} & {$2.04\mathrm{e}{-1} \pm 1.70\mathrm{e}{-1}$}
\\
{Out-of-distribution 1}  & {$1.65\mathrm{e}{-1} \pm 3.70\mathrm{e}{-1}$} & {$1.97\mathrm{e}{-1} \pm 3.43\mathrm{e}{-1}$} & {$3.76\mathrm{e}{-1} \pm 2.81\mathrm{e}{-1}$} & {$8.03\mathrm{e}{-1} \pm 4.29\mathrm{e}{-1}$} & {$1.07\mathrm{e}{0} \pm 4.03\mathrm{e}{-1}$}
\\
{Out-of-distribution 2}  & {$4.11\mathrm{e}{-1} \pm 4.25\mathrm{e}{-1}$} & {$4.42\mathrm{e}{-1} \pm 4.06\mathrm{e}{-1}$} & {$4.60\mathrm{e}{-1} \pm 2.26\mathrm{e}{-1}$} & {$7.75\mathrm{e}{-1} \pm 3.04\mathrm{e}{-1}$} & {$1.14\mathrm{e}{0} \pm 3.33\mathrm{e}{-1}$}
\\
{Out-of-distribution 3}  & {$2.47\mathrm{e}{-1} \pm 6.11\mathrm{e}{-2}$} & {$\textBF{2.56\mathrm{e}{-1} \pm 7.88\mathrm{e}{-2}}$} & {$4.72\mathrm{e}{-1} \pm 2.50\mathrm{e}{-1}$} & {$\textBF{7.57\mathrm{e}{-1} \pm 4.01\mathrm{e}{-1}}$} & {$\textBF{9.90\mathrm{e}{-1} \pm 3.94\mathrm{e}{-1}}$}
\\
{Out-of-distribution 4}  & {$2.97\mathrm{e}{-1} \pm 2.06\mathrm{e}{-1}$} & {$\textBF{3.21\mathrm{e}{-1} \pm 1.98\mathrm{e}{-1}}$} & {$5.49\mathrm{e}{-1} \pm 2.88\mathrm{e}{-1}$} & {$8.58\mathrm{e}{-1} \pm 3.25\mathrm{e}{-1}$} & {$1.15\mathrm{e}{0} \pm 3.73\mathrm{e}{-1}$}
\\

\midrule
{VAE, extended}  & {$t=0$} & {$t=0.25$} & {$t=0.5$} & {$t=0.75$} & {$t=1$} 
\\ 
\midrule
{In-distribution}  & {$5.38\mathrm{e}{-2} \pm 1.00\mathrm{e}{-2}$} & {$3.85\mathrm{e}{-2} \pm 4.99\mathrm{e}{-3}$} & {$4.35\mathrm{e}{-2} \pm 1.46\mathrm{e}{-2}$} & {$8.91\mathrm{e}{-2} \pm 8.86\mathrm{e}{-2}$} & {$3.13\mathrm{e}{-1} \pm 1.98\mathrm{e}{-1}$}
\\
{Out-of-distribution 1}  & {$1.25\mathrm{e}{-1} \pm 2.01\mathrm{e}{-1}$} & {$\textBF{1.28\mathrm{e}{-1} \pm 9.73\mathrm{e}{-2}}$} & {$3.46\mathrm{e}{-1} \pm 2.46\mathrm{e}{-1}$} & {$7.54\mathrm{e}{-1} \pm 3.95\mathrm{e}{-1}$} & {$1.05\mathrm{e}{0} \pm 3.61\mathrm{e}{-1}$}
\\
{Out-of-distribution 2}  & {$4.65\mathrm{e}{-1} \pm 3.37\mathrm{e}{-1}$} & {$\textBF{4.16\mathrm{e}{-1} \pm 2.01\mathrm{e}{-1}}$} & {$5.07\mathrm{e}{-1} \pm 2.33\mathrm{e}{-1}$} & {$8.30\mathrm{e}{-1} \pm 3.48\mathrm{e}{-1}$} & {$1.09\mathrm{e}{0} \pm 3.67\mathrm{e}{-1}$}
\\
{Out-of-distribution 3}  & {$2.44\mathrm{e}{-1} \pm 7.33\mathrm{e}{-2}$} & {$3.32\mathrm{e}{-1} \pm 4.63\mathrm{e}{-2}$} & {$5.87\mathrm{e}{-1} \pm 2.11\mathrm{e}{-1}$} & {$8.42\mathrm{e}{-1} \pm 3.02\mathrm{e}{-1}$} & {$1.17\mathrm{e}{0} \pm 3.31\mathrm{e}{-1}$}
\\
{Out-of-distribution 4}  & {$2.52\mathrm{e}{-1} \pm 9.18\mathrm{e}{-2}$} & {$3.28\mathrm{e}{-1} \pm 9.50\mathrm{e}{-2}$} & {$5.78\mathrm{e}{-1} \pm 3.07\mathrm{e}{-1}$} & {$\textBF{7.62\mathrm{e}{-1} \pm 3.88\mathrm{e}{-1}}$} & {$\textBF{1.02\mathrm{e}{0} \pm 4.06\mathrm{e}{-1}}$}
\\

\midrule
{VAE-DLM (ours)}  & {$t=0$} & {$t=0.25$} & {$t=0.5$} & {$t=0.75$} & {$t=1$} 
\\ 
\midrule
{In-distribution}  & {$4.89\mathrm{e}{-2} \pm 9.15\mathrm{e}{-3}$} & {$2.72\mathrm{e}{-2} \pm 5.08\mathrm{e}{-2}$} & {$3.43\mathrm{e}{-2} \pm 1.44\mathrm{e}{-2}$} & {$1.00\mathrm{e}{-1} \pm 1.16\mathrm{e}{-1}$} & {$3.11\mathrm{e}{-1} \pm 1.69\mathrm{e}{-1}$}
\\
{Out-of-distribution 1}  & {$\textBF{1.12\mathrm{e}{-1} \pm 1.09\mathrm{e}{-1}}$} & {$1.48\mathrm{e}{-1} \pm 1.70\mathrm{e}{-1}$} & {$\textBF{2.98\mathrm{e}{-1} \pm 2.24\mathrm{e}{-1}}$} & {$\textBF{6.23\mathrm{e}{-1} \pm 3.86\mathrm{e}{-1}}$} & {$\textBF{8.67\mathrm{e}{-1} \pm 4.12\mathrm{e}{-1}}$}
\\
{Out-of-distribution 2}  & {$\textBF{3.60\mathrm{e}{-1} \pm 2.25\mathrm{e}{-1}}$} & {$4.72\mathrm{e}{-1} \pm 2.59\mathrm{e}{-1}$} & {$\textBF{4.20\mathrm{e}{-1} \pm 1.76\mathrm{e}{-1}}$} & {$\textBF{7.47\mathrm{e}{-1} \pm 3.59\mathrm{e}{-1}}$} & {$1.09\mathrm{e}{0} \pm 4.07\mathrm{e}{-1}$}
\\
{Out-of-distribution 3}  & {$\textBF{1.84\mathrm{e}{-1} \pm 4.85\mathrm{e}{-2}}$} & {$3.16\mathrm{e}{-1} \pm 1.50\mathrm{e}{-1}$} & {$4.72\mathrm{e}{-1} \pm 3.20\mathrm{e}{-1}$} & {$7.85\mathrm{e}{-1} \pm 3.85\mathrm{e}{-1}$} & {$1.01\mathrm{e}{0} \pm 3.94\mathrm{e}{-1}$}
\\
{Out-of-distribution 4}  & {$\textBF{2.12\mathrm{e}{-1} \pm 7.31\mathrm{e}{-2}}$} & {$3.35\mathrm{e}{-1} \pm 1.50\mathrm{e}{-1}$} & {$\textBF{5.05\mathrm{e}{-1} \pm 2.97\mathrm{e}{-1}}$} & {$8.52\mathrm{e}{-1} \pm 3.15\mathrm{e}{-1}$} & {$1.13\mathrm{e}{0} \pm 3.19\mathrm{e}{-1}$}
\\

\bottomrule

\end{tabular}

\caption{ We list relative $L^1$ errors of our method along with a VAE baseline on the KS equation on 30 identical (both noise and initial data) test samples. We choose scaling coefficients $(\alpha, \beta, \gamma_{\text{geo flow}}, \gamma_{\text{metric}}) = (100,1\mathrm{e}{-5},1,10)$.  Our OOD scenarios are: (1) no scaling of an in-distribution initial condition and noise with $\sigma=0.5$ injected in all 201 locations; (2) 1.3 scaling and noise with $\sigma=0.65$ injected in 51 locations; (3) choosing the initial condition to be 0 in 50 locations (solution is sparsely observed in 151 locations); (4) choosing the initial condition to be 0 in 50 locations and noise with $\sigma=0.65$ injected in the 151 other locations. We choose the same OOD conditions as Table \ref{tab:KS_highKLreg}}

\label{tab:KS}

\end{table}

\vfill

\vfill

\clearpage
\newpage

\begin{table}[t]
\centering

\scriptsize

\begin{tabular}{wl{3.0cm} | P{2.2cm} P{2.2cm} P{2.2cm} P{2.2cm} P{2.2cm} } 
\toprule
\belowrulesepcolor{light-gray} 
\rowcolor{light-gray} \multicolumn{6}{l} {Kuramoto-Sivashinsky equation} 
\\ 
\aboverulesepcolor{light-gray} 
\midrule
{VAE}  & {$t=0$} & {$t=0.25$} & {$t=0.5$} & {$t=0.75$} & {$t=1$} 
\\ 
\midrule
{In-distribution}  & {$4.29\mathrm{e}{-2} \pm 1.64\mathrm{e}{-2}$} & {$2.73\mathrm{e}{-2} \pm 1.15\mathrm{e}{-2}$} & {$2.22\mathrm{e}{-2} \pm 1.10\mathrm{e}{-2}$} & {$5.85\mathrm{e}{-2} \pm 8.47\mathrm{e}{-2}$} & {$2.16\mathrm{e}{-1} \pm 1.74\mathrm{e}{-1}$}
\\
{Out-of-distribution 1}  & {$1.56\mathrm{e}{-1} \pm 3.12\mathrm{e}{-1}$} & {$1.90\mathrm{e}{-1} \pm 2.91\mathrm{e}{-1}$} & {$4.18\mathrm{e}{-1} \pm 3.14\mathrm{e}{-1}$} & {$8.08\mathrm{e}{-1} \pm 3.48\mathrm{e}{-1}$} & {$1.06\mathrm{e}{0} \pm 3.44\mathrm{e}{-1}$}
\\
{Out-of-distribution 2}  & {$2.11\mathrm{e}{-1} \pm 2.57\mathrm{e}{-1}$} & {$2.37\mathrm{e}{-1} \pm 2.63\mathrm{e}{-1}$} & {$3.49\mathrm{e}{-1} \pm 3.31\mathrm{e}{-1}$} & {$\textBF{6.39\mathrm{e}{-1} \pm 3.61\mathrm{e}{-1}}$} & {$9.42\mathrm{e}{-1} \pm 4.21\mathrm{e}{-1}$}
\\
{Out-of-distribution 3}  & {$9.75\mathrm{e}{-2} \pm 4.77\mathrm{e}{-2}$} & {$1.40\mathrm{e}{-1} \pm 9.76\mathrm{e}{-2}$} & {$\textBF{3.48\mathrm{e}{-1} \pm 3.03\mathrm{e}{-1}}$} & {$\textBF{6.32\mathrm{e}{-1} \pm 2.71\mathrm{e}{-1}}$} & {$\textBF{8.92\mathrm{e}{-1} \pm 3.36\mathrm{e}{-1}}$}
\\
\midrule
{Out-of-distribution 4}  & {$1.45\mathrm{e}{-1} \pm 1.83\mathrm{e}{-1}$} & {$1.91\mathrm{e}{-1} \pm 1.51\mathrm{e}{-1}$} & {$4.53\mathrm{e}{-1} \pm 3.13\mathrm{e}{-1}$} & {$\textBF{7.62\mathrm{e}{-1} \pm 2.89\mathrm{e}{-1}}$} & {$1.07\mathrm{e}{0} \pm 3.63\mathrm{e}{-1}$}
\\
{Out-of-distribution 5}  & {$9.54\mathrm{e}{-2} \pm 3.23\mathrm{e}{-2}$} & {$1.16\mathrm{e}{-1} \pm 7.15\mathrm{e}{-2}$} & {$1.98\mathrm{e}{-1} \pm 1.30\mathrm{e}{-1}$} & {$4.58\mathrm{e}{-1} \pm 3.04\mathrm{e}{-1}$} & {$7.54\mathrm{e}{-1} \pm 3.73\mathrm{e}{-1}$}
\\
{Out-of-distribution 6}  & {$1.35\mathrm{e}{-1} \pm 1.76\mathrm{e}{-1}$} & {$1.62\mathrm{e}{-1} \pm 1.63\mathrm{e}{-1}$} & {$2.91\mathrm{e}{-1} \pm 2.51\mathrm{e}{-1}$} & {$5.69\mathrm{e}{-1} \pm 3.65\mathrm{e}{-1}$} & {$8.72\mathrm{e}{-1} \pm 4.30\mathrm{e}{-1}$}
\\

\midrule
{VAE, extended}  & {$t=0$} & {$t=0.25$} & {$t=0.5$} & {$t=0.75$} & {$t=1$} 
\\ 
\midrule
{In-distribution}  & {$3.19\mathrm{e}{-2} \pm 1.01\mathrm{e}{-2}$} & {$2.51\mathrm{e}{-2} \pm 4.24\mathrm{e}{-3}$} & {$3.94\mathrm{e}{-2} \pm 1.28\mathrm{e}{-2}$} & {$1.17\mathrm{e}{-1} \pm 1.09\mathrm{e}{-1}$} & {$3.53\mathrm{e}{-1} \pm 2.14\mathrm{e}{-1}$}
\\
{Out-of-distribution 1}  & {$\textBF{1.04\mathrm{e}{-1} \pm 1.99\mathrm{e}{-1}}$} & {$1.52\mathrm{e}{-1} \pm 2.53\mathrm{e}{-1}$} & {$3.42\mathrm{e}{-1} \pm 2.54\mathrm{e}{-1}$} & {$7.58\mathrm{e}{-1} \pm 4.19\mathrm{e}{-1}$} & {$9.81\mathrm{e}{-1} \pm 3.90\mathrm{e}{-1}$}
\\
{Out-of-distribution 2}  & {$\textBF{1.16\mathrm{e}{-1} \pm 1.40\mathrm{e}{-1}}$} & {$\textBF{1.59\mathrm{e}{-1} \pm 1.23\mathrm{e}{-1}}$} & {$3.55\mathrm{e}{-1} \pm 1.91\mathrm{e}{-1}$} & {$7.77\mathrm{e}{-1} \pm 3.51\mathrm{e}{-1}$} & {$1.10\mathrm{e}{0} \pm 3.95\mathrm{e}{-1}$}
\\
{Out-of-distribution 3}  & {$\textBF{6.56\mathrm{e}{-2} \pm 3.41\mathrm{e}{-2}}$} & {$\textBF{1.13\mathrm{e}{-1} \pm 7.24\mathrm{e}{-2}}$} & {$3.79\mathrm{e}{-1} \pm 3.13\mathrm{e}{-1}$} & {$7.46\mathrm{e}{-1} \pm 3.87\mathrm{e}{-1}$} & {$9.87\mathrm{e}{-1} \pm 3.93\mathrm{e}{-1}$}
\\
\midrule
{Out-of-distribution 4}  & {$1.20\mathrm{e}{-1} \pm 1.23\mathrm{e}{-1}$} & {$1.73\mathrm{e}{-1} \pm 1.22\mathrm{e}{-1}$} & {$4.74\mathrm{e}{-1} \pm 3.33\mathrm{e}{-1}$} & {$8.23\mathrm{e}{-1} \pm 3.62\mathrm{e}{-1}$} & {$1.13\mathrm{e}{0} \pm 3.47\mathrm{e}{-1}$}
\\
{Out-of-distribution 5}  & {$6.56\mathrm{e}{-2} \pm 2.84\mathrm{e}{-2}$} & {$1.05\mathrm{e}{-1} \pm 6.43\mathrm{e}{-2}$} & {$2.72\mathrm{e}{-1} \pm 1.63\mathrm{e}{-1}$} & {$6.84\mathrm{e}{-1} \pm 3.29\mathrm{e}{-1}$} & {$9.93\mathrm{e}{-1} \pm 4.20\mathrm{e}{-1}$}
\\
{Out-of-distribution 6}  & {$8.25\mathrm{e}{-2} \pm 5.13\mathrm{e}{-2}$} & {$1.22\mathrm{e}{-1} \pm 8.20\mathrm{e}{-2}$} & {$3.28\mathrm{e}{-1} \pm 1.96\mathrm{e}{-1}$} & {$7.63\mathrm{e}{-1} \pm 3.58\mathrm{e}{-1}$} & {$1.03\mathrm{e}{0} \pm 3.93\mathrm{e}{-1}$}
\\

\midrule
{VAE-DLM (ours)}  & {$t=0$} & {$t=0.25$} & {$t=0.5$} & {$t=0.75$} & {$t=1$} 
\\ 
\midrule
{In-distribution}  & {$3.98\mathrm{e}{-2} \pm 1.15\mathrm{e}{-2}$} & {$3.05\mathrm{e}{-2} \pm 3.52\mathrm{e}{-3}$} & {$4.44\mathrm{e}{-2} \pm 2.18\mathrm{e}{-2}$} & {$1.34\mathrm{e}{-1} \pm 1.29\mathrm{e}{-1}$} & {$4.10\mathrm{e}{-1} \pm 2.16\mathrm{e}{-1}$} 
\\
{Out-of-distribution 1}  & {$1.22\mathrm{e}{-1} \pm 2.92\mathrm{e}{-1}$} & {$\textBF{1.47\mathrm{e}{-1} \pm 2.83\mathrm{e}{-1}}$} & {$\textBF{3.09\mathrm{e}{-1} \pm 2.51\mathrm{e}{-1}}$} & {$\textBF{6.74\mathrm{e}{-1} \pm 4.11\mathrm{e}{-1}}$} & {$\textBF{9.30\mathrm{e}{-1} \pm 4.35\mathrm{e}{-1}}$}
\\
{Out-of-distribution 2}  & {$1.34\mathrm{e}{-1} \pm 2.65\mathrm{e}{-1}$} & {$1.70\mathrm{e}{-1} \pm 2.44\mathrm{e}{-1}$} & {$\textBF{2.97\mathrm{e}{-1} \pm 1.97\mathrm{e}{-1}}$} & {$6.55\mathrm{e}{-1} \pm 3.85\mathrm{e}{-1}$} & {$\textBF{9.27\mathrm{e}{-1} \pm 4.08\mathrm{e}{-1}}$}
\\
{Out-of-distribution 3}  & {$7.71\mathrm{e}{-2} \pm 4.61\mathrm{e}{-2}$} & {$1.16\mathrm{e}{-1} \pm 7.26\mathrm{e}{-2}$} & {$3.63\mathrm{e}{-1} \pm 2.76\mathrm{e}{-1}$} & {$7.30\mathrm{e}{-1} \pm 4.47\mathrm{e}{-1}$} & {$9.83\mathrm{e}{-1} \pm 4.46\mathrm{e}{-1}$}
\\
\midrule
{Out-of-distribution 4}  & {$\textBF{1.13\mathrm{e}{-1} \pm 1.06\mathrm{e}{-1}}$} & {$\textBF{1.59\mathrm{e}{-1} \pm 1.23\mathrm{e}{-1}}$} & {$\textBF{4.04\mathrm{e}{-1} \pm 2.82\mathrm{e}{-1}}$} & {$7.99\mathrm{e}{-1} \pm 3.76\mathrm{e}{-1}$} & {$\textBF{1.02\mathrm{e}{0} \pm 3.75\mathrm{e}{-1}}$}
\\
{Out-of-distribution 5}  & {$\textBF{5.15\mathrm{e}{-2} \pm 1.91\mathrm{e}{-2}}$} & \textcolor{blue}{$\textBF{6.69\mathrm{e}{-2} \pm 4.27\mathrm{e}{-2}}$} & \textcolor{blue}{$\textBF{1.45\mathrm{e}{-1} \pm 9.34\mathrm{e}{-2}}$} & {$\textBF{3.67\mathrm{e}{-1} \pm 2.33\mathrm{e}{-1}}$} & {$\textBF{6.76\mathrm{e}{-1} \pm 3.57\mathrm{e}{-1}}$}
\\
{Out-of-distribution 6}  & {$\textBF{6.28\mathrm{e}{-2} \pm 5.93\mathrm{e}{-2}}$} & \textcolor{blue}{$\textBF{8.32\mathrm{e}{-2} \pm 8.02\mathrm{e}{-2}}$} & \textcolor{blue}{$\textBF{1.77\mathrm{e}{-1} \pm 1.40\mathrm{e}{-1}}$} & {$\textBF{4.52\mathrm{e}{-1} \pm 3.47\mathrm{e}{-1}}$} & {$\textBF{7.41\mathrm{e}{-1} \pm 3.92\mathrm{e}{-1}}$}
\\

\bottomrule

\end{tabular}

\caption{ We list relative $L^1$ errors of our method along with a VAE baseline on the KS equation on 30 identical (both noise and initial data) test samples. We choose scaling coefficients $(\alpha, \beta, \gamma_{\text{geo flow}}, \gamma_{\text{metric}}) = (100,1,1,10)$: this is an experiment with high KL regularization with the prior. Our OOD scenarios are: (1) no scaling of an in-distribution initial condition and noise with $\sigma=0.5$ injected in all 201 locations; (2) 1.3 scaling and noise with $\sigma=0.65$ injected in 51 locations; (3) choosing the initial condition to be 0 in 50 locations (solution is sparsely observed in 151 locations); (4) choosing the initial condition to be 0 in 50 locations and noise with $\sigma=0.65$ injected in the 151 other locations; (5) 1.5 scaling and inserting 0 in 50 locations; (6) 1.5 scaling, noise with $\sigma=0.25$ injected in 151 locations, and inserting 0 in the 50 locations without noise. We choose the same OOD conditions as Table \ref{tab:KS} for (1)-(4).}

\label{tab:KS_highKLreg}

\end{table}

\clearpage
\newpage

\section{Vanilla multilayer perceptron}

\begin{figure}[H]
  \vspace{0mm}
  \centering
  \includegraphics[scale=0.7]{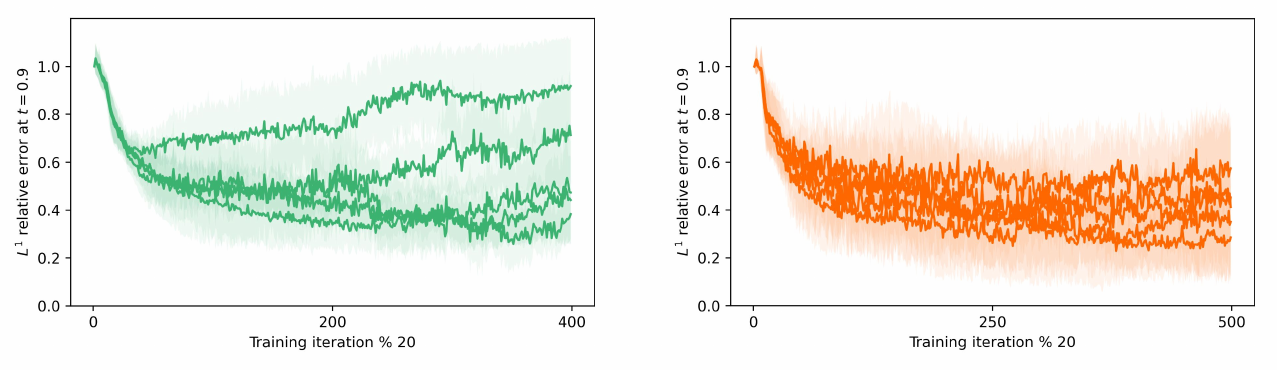}
  \caption{We examine $L^1$ relative error on 5 instances of training for using a vanilla (non-modified) multilayer perceptron (MLP) architecture for \textit{(left)} a vanilla VAE, and \textit{(right)} our method. We emphasize the primary experiments of this paper are done with the modified multilayer perceptron architecture, as presented in in \cite{wang2021learningsolutionoperatorparametric}. This was done on the KS equation, using the same training data of section \ref{sec:KS_eq}. The OOD data was done on 15 samples by injecting noise with $\sigma = 0.75$ in 151 locations, and scaling the initial condition by $1.15$. Batch size is 500. The seed is set for identical OOD data in both cases, and all relevant hyperparameters are otherwise held constant between both cases. For our method, we choose $r^2 = 10, \alpha=0.5$. 
  Note that our method exhibits considerably greater consistency. We also experimented on the Allen-Cahn equation. We found in almost all cases of OOD data on this equation's data, our method performed equivalently to the baseline.}
  \label{fig:vanilla_and_bestlin_tanh_all}
\end{figure}

\begin{figure}[H]
  \vspace{0mm}
  \centering
  \includegraphics[scale=0.7]{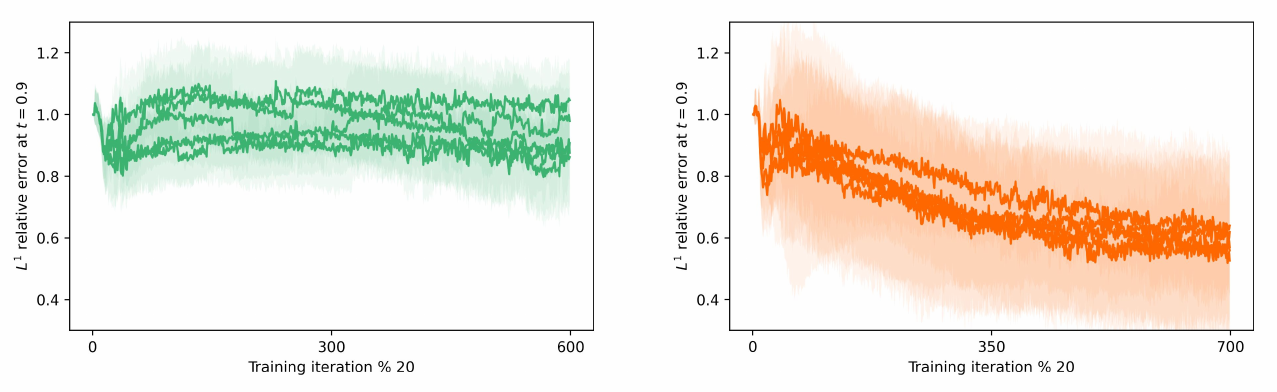}
  \caption{We examine $L^1$ relative error on 5 instances of training for using a vanilla (non-modified) multilayer perceptron (MLP) architecture for \textit{(left)} a vanilla VAE, and \textit{(right)} our method. This figure is also upon the KS equation, with noise injected in 21 locations with $\sigma =0.5$, and initial condition scaled by $1.3$. We highlight we used gelu activation for all networks in this experiment, which is notable since we often we tanh in the encoder $\mathcal{E}$ and decoder $\mathcal{D}$. We used a larger batch size of 1,000 for this experiment, and we found tuning the loss coefficients in the vanilla VAE framework had minimal effect on results. We stress that this figures illustrates approximately our best results using the vanilla MLP, and using tanh in the decoder generally yields less difference in results.}
  \label{fig:vanilla_and_bestlin_tanh_all}
\end{figure}

\newpage

\section{Training time}
\label{app:training_time}

\begin{table}[ht]
\small
\centering
\begin{tabular}{wl{5.0cm} | P{3.0cm} } 
\toprule
\belowrulesepcolor{light-gray} 
\rowcolor{light-gray} \multicolumn{2}{l} {Training time} 
\\ 
\aboverulesepcolor{light-gray} 
\midrule
{} & {Training time} 
\\ 
\midrule
{Burger's equation}  & {$\sim$ 25 minutes}
\\
{Allen-Cahn equation}  & {$\sim$ 30 minutes}
\\
{Porous medium equation}  & {$\sim$ 10 minutes} 
\\
{KS equation ($\beta = 1\mathrm{e}{-4}$)}  & {$\sim$ 10 minutes} 
\\
{KdV equation}  & {$\sim$ 10 minutes} 
\\
\bottomrule

\end{tabular}

\vspace{2mm}

\caption{We illustrate the approximate training time of our VAE-DLM methods of the experiments presented in section \ref{sec:experiments} and the error tables. Note that training time is affected by levels of KL regularization ($\beta$) and training convergence. Our training time for the KdV equation corresponds to that of Figure \ref{fig:KdV_eq_generative}. Training is done on a T4 GPU.}

\end{table}

\section{Dimensions of experiments}
\label{app:dimensions}

\begin{table}[ht]
\small
\centering
\begin{tabular}{wl{5.0cm} | P{3.0cm} | P{3.0cm} } 
\toprule
\belowrulesepcolor{light-gray} 
\rowcolor{light-gray} \multicolumn{3}{l} {Training time} 
\\ 
\aboverulesepcolor{light-gray} 
\midrule
{} & {Intrinsic dimension}  & {Extrinsic dimension (embedding dimension)} 
\\ 
\midrule
{Burger's equation}  & {6} & {7}
\\
{Allen-Cahn equation}  & {6} & {7}
\\
{Porous medium equation}  & {8}  & {9}
\\
{KS equation}  & {6}  & {7}
\\
{KdV equation}  & {8}  & {9}
\\
\bottomrule

\end{tabular}

\vspace{2mm}

\caption{We illustrate the dimensions of our geometric flows. We remark we found these dimensions to be sufficient, and our method is applicable scaling up to $15 +$ latent dimension with modest offline cost, depending on computing power. Also note the offline cost is almost entirely determined by intrinsic dimension and not extrinsic; one could consider higher extrinsic dimension. We empirically found our method often works well with $\text{extrinsic dimension} = \text{intrinsic dimension} + 1$, but higher dimensions could be considered.}

\end{table}

\section{Miscellaneous discussion}

Our work is closely related to the work of ~\cite{lopez2022gdvaesgeometricdynamicvariational}, so we recommend a researcher in this area to also read this work in GD-VAEs. A key difference between our methods and the alternative methods of this paper, aside from the dynamic manifold itself, is that, in the GD-VAE case, the manifold is fixed beforehand with preset datapoints in space. Mapping onto the manifold is done directly onto this fixed space as a way to progress the dynamics. With our methods, the in-distribution data is mapped to the manifold, but it cannot be ascertained that out-of-distribution (OOD) data lies exactly on the manifold. The neural network latent space thus acts as a regularizer, which helps learning, but does not guarantee a geometric structure for such OOD data.

\end{document}